\documentclass{article}
\usepackage[final]{corl_2019}
\usepackage{float} 
\usepackage{mathtools}
\usepackage{hyperref}
\usepackage{amssymb}
\usepackage{bbm}
\usepackage{graphicx}
\usepackage{multicol}
\usepackage{multirow}
\usepackage{caption}
\usepackage{subcaption}
\usepackage{algorithmic}
\usepackage{algorithm}
\usepackage[space]{grffile}

\usepackage{enumitem,kantlipsum}
\usepackage{tablefootnote}
\usepackage{booktabs}
\usepackage{dsfont}
\usepackage{gensymb}
\usepackage{wrapfig}
\let\ACMmaketitle=\maketitle
\renewcommand{\maketitle}{\begingroup\let\footnote=\thanks \ACMmaketitle\endgroup}
\makeatletter
\usepackage{graphicx}
\let\@fnsymbol\@arabic
\makeatother
\title{MAT: Multi-Fingered Adaptive Tactile Grasping via Deep Reinforcement Learning}
\author{
  Bohan Wu\footnotemark[1]\hspace{1.5mm}, Iretiayo Akinola\footnotemark[1]\hspace{1.5mm}, Jacob Varley\footnotemark[2]\hspace{1.5mm}, Peter K. Allen\footnotemark[1]\\
  \footnotemark[1]\hspace{2mm}Columbia University, New York, NY, United States\\
  \footnotemark[2]\hspace{2mm}Robotics at Google, United States\\
  \texttt{bw2505@columbia.edu, iakinola@cs.columbia.edu,}\\
  \texttt{jakevarley@google.com, allen@cs.columbia.edu}
}

\begin{document}
\maketitle

\begin{abstract}
Vision-based grasping systems typically adopt an open-loop execution of a planned grasp. This policy can fail due to many reasons, including ubiquitous calibration error.  Recovery from a failed grasp is further complicated by visual occlusion, as the hand is usually occluding the vision sensor as it attempts another open-loop regrasp. This work presents MAT, a tactile closed-loop method capable of realizing grasps provided by a coarse initial positioning of the hand above an object. Our algorithm is a deep reinforcement learning (RL) policy optimized through the clipped surrogate objective within a maximum entropy RL framework to balance exploitation and exploration. The method utilizes tactile and proprioceptive information to act through both fine finger motions and larger regrasp movements to execute stable grasps. A novel curriculum of action motion magnitude makes learning more tractable and helps turn common failure cases into successes. Careful selection of features that exhibit small sim-to-real gaps enables this tactile grasping policy, trained purely in simulation, to transfer well to real world environments without the need for additional learning. Experimentally, this methodology improves over a vision-only grasp success rate substantially on a multi-fingered robot hand. When this methodology is used to realize grasps from coarse initial positions provided by a vision-only planner, the system is made dramatically more robust to calibration errors in the camera-robot transform. 
\end{abstract}

\keywords{Tactile, Deep Reinforcement Learning, Multi-Fingered Grasping} 


\section{Introduction}
As multi-fingered grasping becomes more tractable thanks to advances in vision and deep RL, improving state-of-the-art methods that achieve $90\%+$ grasp success rates becomes more difficult. Among the few percentages of failed grasps are those caused by grasp slip, low friction, calibration error and adversarial object shapes. A promising direction for finishing the last mile of the race towards high-performance, high-success autonomous grasping is the idea of closed-loop grasping: continuously adjusting the robot's DOFs to improve the quality of the current grasp based on sensory feedback. Closed-loop grasping is attractive because it enables the robot to correct the initial grasp to achieve even higher pick-up success rates, given an approximately correct initial grasp pose. 

Performing high-quality closed-loop grasping requires a sensor modality that is both free of external disturbances from the robot's ongoing actions and accurate in providing information about the state of the current grasp. Vision, RGB or RGB-D, becomes a less favorable candidate in this case due to visual occlusion. As the robot's end-effector approaches the graspable object, camera vision will be blocked by either the end-effector palm or fingers. Therefore, it is difficult to enable vision to provide undisturbed and accurate information about the status of the current grasp.

\begin{figure}[!t]
    \begin{minipage}[t]{0.21\textwidth}
        \vspace{0pt}
        \centering
        \begin{subfigure}{0.75\textwidth}
        \captionsetup{skip=2pt}
            \includegraphics[width=\linewidth]{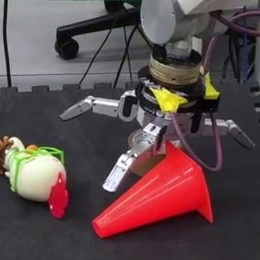}
            \caption{}
        \end{subfigure}
        \begin{subfigure}{0.75\textwidth}
        \captionsetup{skip=2pt}
            \includegraphics[width=\linewidth]{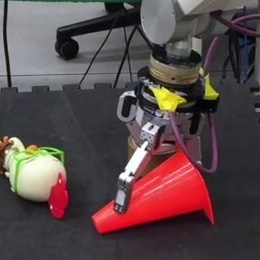}
            \caption{}
        \end{subfigure}
        \caption{\small{\textbf{Open-Loop Grasping.} Open-loop grasping \textbf{(a)} planned from an initial image of the scene fails to form a stable grasp \textbf{(b)}.}}
        \label{fig:openloop1}
    \end{minipage}
    \hspace{3mm}
    \begin{minipage}[t]{.73\textwidth}
        \vspace{0pt}
        \centering
        \begin{subfigure}[t]{0.215\textwidth}
        \captionsetup{skip=2pt}
            \includegraphics[width=\linewidth]{images/closeloop_v2/frame418_cropped.jpg}\caption{}
        \end{subfigure}
        \begin{subfigure}[t]{0.215\textwidth}
        \captionsetup{skip=2pt}
            \includegraphics[width=\linewidth]{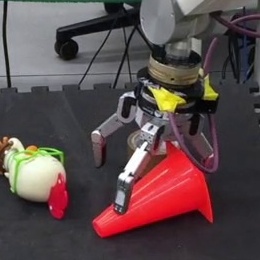}\caption{}
        \end{subfigure}
        \begin{subfigure}[t]{0.215\textwidth}
        \captionsetup{skip=2pt}
            \includegraphics[width=\linewidth]{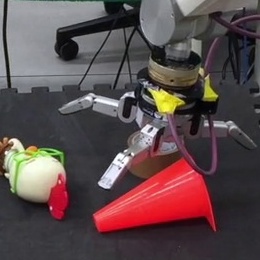}\caption{}
        \end{subfigure}
        \begin{subfigure}[t]{0.215\textwidth}
        \captionsetup{skip=2pt}
            \includegraphics[width=\linewidth]{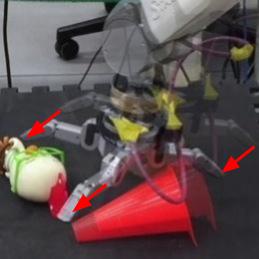}\caption{}
        \end{subfigure}
        
        \begin{subfigure}[t]{0.215\textwidth}
        \captionsetup{skip=2pt}
            \includegraphics[width=\linewidth]{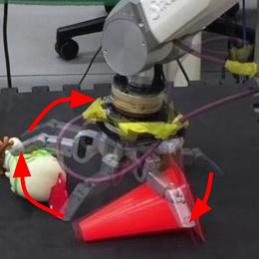}\caption{}
        \end{subfigure}
        \begin{subfigure}[t]{0.215\textwidth}
        \captionsetup{skip=2pt}
            \includegraphics[width=\linewidth]{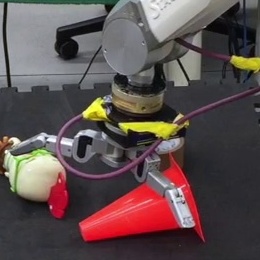}\caption{}
        \end{subfigure}
        \begin{subfigure}[t]{0.215\textwidth}
        \captionsetup{skip=2pt}
            \includegraphics[width=\linewidth]{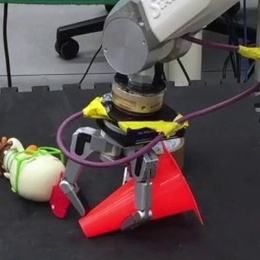}\caption{}
        \end{subfigure}
        \begin{subfigure}[t]{0.215\textwidth}
        \captionsetup{skip=2pt}
            \includegraphics[width=\linewidth]{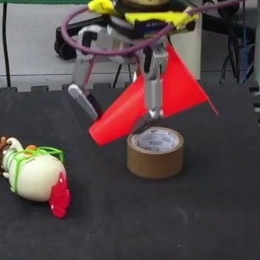}\caption{}
        \end{subfigure}
      \caption{\small{\textbf{Multi-Fingered Adaptive Tactile Grasping.} Behavior of our policy given \textbf{(a)} a coarse initial grasp pose. \textbf{(b)} The robot begins by closing the fingers in small increments to form a grasp adapting to tactile contacts as they occur. \textbf{(c)} Eventually, unsatisfied with the tactile and proprioceptive observations, the policy decides to reopen the hand and \textbf{(d-e)} adjust the end-effector position and orientation. \textbf{(f-g)} The policy closes the fingers incrementally again. \textbf{(h)} Finally, the policy ends the episode with a lift action and successfully picks the object up.}} \label{fig:closeloop}
  \end{minipage}%
  \vspace{-5mm}
\end{figure}

Tactile, in this case, is one of the best candidate sensory modalities for closed-loop grasping. Tactile sensors are both rich in information with many sensor cells on each finger (and palm in some cases) and free of external disturbances that visual systems usually face with different levels of occlusion. 
Figure~\ref{fig:openloop1} shows a common failure case of an open-loop grasping system due to calibration error. Highlighted in Figure~\ref{fig:closeloop}, this paper introduces Multi-Fingered Adaptive Tactile Grasping, or MAT, a high-performance deep RL algorithm that leverages tactile and proprioceptive information for multi-fingered grasping in an adaptive, closed-loop manner, with the ultimate purpose of substantially improving state-of-the-art open-loop grasping systems.
First, MAT allows the robot to learn grasp action primitives in a generative manner via maximum entropy deep RL. These action primitives not only include decisions of granular movements of each of the fingers, lifting the end-effector for pick-up, but also reopening the fingers and adjusting the end-effector position and orientation, thus forming a tight, closed-loop grasping system. Second, since maximum entropy deep RL requires high sample complexity and training experiences that are diverse in terms of object types, quantities, poses, and clutter levels, direct learning or transfer learning in real-world environments becomes a challenge. MAT overcomes this challenge by training in simulation and directly transferring to real without additional learning in a high fidelity way: by choosing observation and action modalities that maintain small sim-to-real gaps to the real world, such as joint angles, binary tactile contacts, tactile contact Cartesian locations, etc. Finally, MAT demonstrates substantially improved pick-up success rates in real-robot experiments over a vision-based, open-loop grasping system. Video of this paper can be found at \url{http://crlab.cs.columbia.edu/MAT/}. In summary, our contributions are:
\begin{enumerate}
\item An adaptive, closed-loop, tactile grasping method that significantly improves single-object and cluttered scene grasp success rates over a strong vision-based open-loop baseline
\item A tactile grasping method that pairs with and improves any open-loop grasping system that generates initial grasp poses. This method \textbf{a)} vastly alleviates the need for a perfectly calibrated robot-camera setup and \textbf{b)} is robust to severe visual occlusion during grasping as its policy, which is active when the hand approaches the object, does not use vision
\item A method of sim-to-real transfer where an end-to-end tactile grasping policy trained in simulation transfers directly to the real world with high fidelity
\item A curriculum learning approach that learns a tightly closed-loop grasping policy from an initial open-loop policy, by gradually increasing the granularity of finger-close movements
\end{enumerate}

\section{Related Work}

\subsection{Vision-Based Closed-Loop Grasping}
Recent works have focused on closed-loop grasping using vision or other non-touch sensor modalities. \cite{kalashnikov2018qt} provides a promising deep RL approach to learn closed-loop grasping using vision.  \cite{morrison2018closing}\cite{viereck2017learning}\cite{levine2018learning} use supervised learning approaches to combine visual learning with closed-loop grasping. In some cases, the closed-loop mechanism in these works becomes less effective as the robot approaches the object due to visual occlusion from the arm or the end-effector.

\subsection{Robotic Grasping with Tactile Only (Blind Grasping without Vision)}
In this setting, tactile readings help design manipulation primitives~\cite{felip2012contact}\cite{felip2013manipulation} or estimate the location and geometry of the objects~\cite{kaboli2017tactile}\cite{kaboli2019tactile}. In \cite{murali2018learning}, the robot makes sweeping motions around the scene to localize the object before attempting grasps roughly at the object centroid. Subsequently, tactile can be used to iteratively adjust the grasp until object pick-up. This approach depends heavily on obtaining a good object localization from touch scanning, which can displace scene objects in ways not detected by the limited tactile coverage, disturbing and negatively affecting the entire grasping process. Consequently, visual-tactile multi-modal methods are becoming more popular.

\subsection{Improving Vision-Based Grasping using Tactile or Other Contact Force Modalities}
Tactile sensors enhance vision-based grasping in a number of ways. Prior to grasping, tactile measurements can help improve geometric knowledge of the grasping scene especially for occluded regions.
During grasping, they help evaluate the success likelihood of a grasp being executed~\cite{li2014learning}\cite{dang2014stable} and decide against lifting the object if the tactile readings indicate a loose grip and an unsuccessful lift.
Further, tactile can be used to predict a re-grasping plan that adjusts the current grasp into a more stable grasp pose~\cite{li2014learning}\cite{dang2014stable}\cite{hogan2018tactile}.
Finally, tactile information can serve as feedback into a closed-loop grasping process that determines how to close robot's fingers given the tactile readings~\cite{merzic2019leveraging}.

\textbf{Leveraging Tactile Information for Shape-Completion Enabled Robotic Grasping.} Recent research has focused on using tactile information to perform more accurate shape modeling of target objects in a scene in order to improve grasp success rate using traditional grasping planners~\cite{wang20183d}\cite{watkins2018multi}\cite{bjorkman2013enhancing}\cite{ilonen2013fusing}. However, these works are still open-loop.

\textbf{Predicting Grasp Success and Stability from Tactile or Proprioceptive Information.}
Learning methods can predict the grasp stability \cite{hyttinen2017estimating}\cite{hyttinen2015learning} or success probability \cite{li2014learning}\cite{dang2014stable}\cite{dang2011blind}\cite{su2015force}\cite{zapata2018non}\cite{cockbum2017grasp}\cite{kwiatkowski2017grasp}\cite{calandra2017feeling}\cite{lu2018planning}\cite{lu2019modeling} given tactile or proprioceptive data received from the robot hand. These readings are recorded after grasping an object, then the object is lifted to obtain a \textit{grasp success or stability} label. The dataset obtained is used to train a grasp critic. In contrast, MAT does not require a critic but learns a unified end-to-end grasping policy.

\textbf{Learning to Regrasp using Tactile Information.} \cite{chebotar2016self} successfully learns regrasping behaviors using a multi-fingered hand as well as a grasp success predictor for predicting grasping outcomes. \cite{calandra2018more} uses Gelsight tactile sensors to learn a state-action value function to predict grasping success and selects relatively good grasping actions from a set of randomly sampled candidates. In contrast to these approaches, our approach does not trigger reopening behaviors using a grasp success predictor but instead learns the task of tactile-enabled closed-loop grasping in a coherent framework, enabling the robot to reopen fingers at any point during grasping. Our algorithm performs the grasping control at a more granular resolution-- moving each finger separately and in small increments.
In addition, while \cite{chebotar2016self}\cite{calandra2018more} require real-world data which can be expensive in time and effort, our tactile grasping policy was trained purely in simulation and transfers directly to the real robot.

\textbf{Reinforcement-Learning to Grasp Using Tactile and Contact Forces.} In \cite{merzic2019leveraging} contact forces are leveraged to improve multi-fingered grasping success and stability. However, their work focuses on simulation results without extending into the real world. \cite{chebotar2016self} presented an RL approach to obtain reopening behaviors on a real robot for picking up a cylindrical object. However, they used a linear function approximator to represent the policy and noted that a more complex policy class can improve robustness especially when dealing with a wider variety of objects. Like \cite{chebotar2016self}, MAT learns to iteratively generate improved grasp poses when needed. In contrast to \cite{chebotar2016self}, MAT is active throughout the entire finger closing process to quickly adapt the current grasp to tactile sensory data as contact occurs, and handle various failure cases during grasping.
Also, MAT was demonstrated in both real-world single-object and cluttered scenes. 

\section{Preliminaries}
\textbf{RL Formulation for MAT.} In MAT's RL formulation, a tactile grasping robotic \textit{agent} interacts with an \textit{environment} to maximize the expected reward~\cite{sutton1998reinforcement}.
The environment is a Partially Observable Markov Decision Process (POMDP), since the agent cannot observe any visual information. Even with vision, this environment is still a POMDP since the agent can encounter visual occlusion and cannot observe the complete 3D geometry of any object or the entire scene.
\begin{wrapfigure}{r}{0.28\textwidth}
\centering
    \begin{minipage}{\linewidth}
        \begin{subfigure}[t]{0.975\textwidth}
            \centering
            \includegraphics[width=3.5cm]{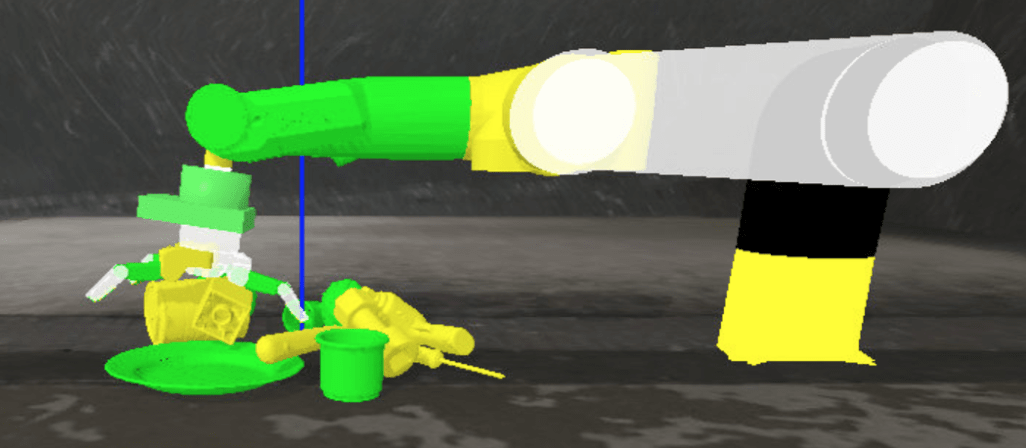}\caption{Simulation Scene (Train)}
            \label{fig:pbscene}
        \end{subfigure}
        \begin{subfigure}[t]{0.975\textwidth}
            \centering
            \includegraphics[width=3.5cm]{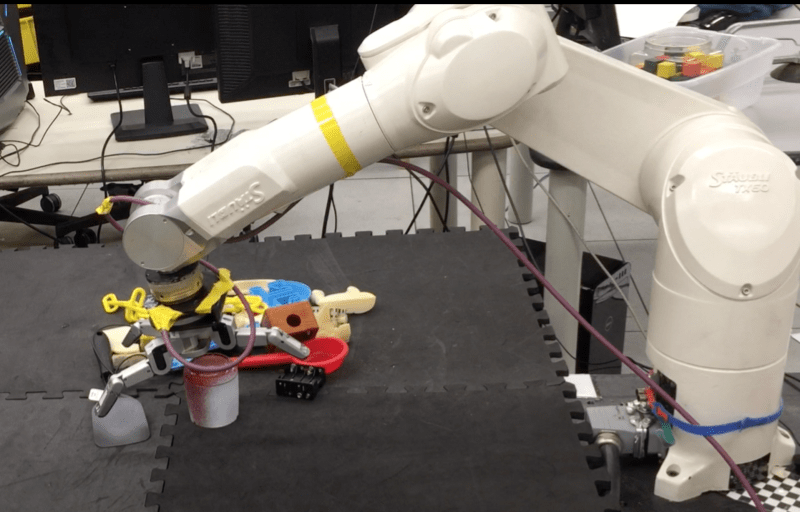}
            \caption{Real Scene (Test)}
            \label{fig:robotscene}
        \end{subfigure}
    \end{minipage}
    \caption{\small Setup. \textbf{(a)} A tactile-enabled Barrett Hand in PyBullet simulation for training. \textbf{(b)} Real robot for evaluation.}
    \label{fig:hardware}
    \vspace{-8mm}
\end{wrapfigure}
To foster good generalization and transfer, MAT models this environment as an MDP defined by $\langle \mathcal{S}, \rho_0, \mathcal{A}, \mathcal{R}, \mathcal{T}, \gamma, H \rangle$ with an observation space $\mathcal{S}$, an initial state distribution $\rho_0 \in \Pi(\mathcal{S})$, an action space $\mathcal{A}$, a reward function $\mathcal{R}: \mathcal{S} \times \mathcal{A} \to \mathbb{R}$, a dynamics model $\mathcal{T}: \mathcal{S} \times \mathcal{A} \to \Pi(\mathcal{S})$, a discount factor $\gamma \in [0, 1)$, and a finite horizon $H$.
$\Pi(\cdot)$ defines a probability distribution over a set. The agent acts according to stationary stochastic policy $\pi: \mathcal{S} \to \Pi(\mathcal{A})$, which specifies action choice probabilities for each observation.
Each policy $\pi$ has a corresponding $Q_\pi: \mathcal{S}\times\mathcal{A} \to \mathbb{R}$ state-action value function that defines the expected discounted cumulative reward for taking an action $a$ from observation $s$ and following $\pi$ from that point onward.

\textbf{Hardware and Simulation Setup.} In both PyBullet (Figure~\ref{fig:pbscene}) \cite{coumans2016PyBullet} simulation and real-world (Figure~\ref{fig:robotscene}), we use the Staubli-TX60 Arm and the Barrett Hand (BH-282), which has 24 capacitive tactile cells on each of the three fingers and the palm, totaling 96 cells. Hereafter, we use $n$ to refer to the number of fingers the robotic hand has, and $t_{final}$ to denote the last timestep of the episode. The finger joint angles of the Barrett Hand range from $0 \text{ rad}$ (open) to $2.44 \text{ rad}$ (close).

\section{Multi-Fingered Adaptive Tactile Grasping}
MAT models the task of closed-loop tactile grasping as a finite-horizon MDP. During each episode, the robot makes a single pick-up attempt on the scene. To begin, we assume that a grasping system of any kind generates an end-effector grasp pose for a cluttered scene, after which visual information is assumed to be unavailable as the arm occludes the scene while realizing the grasp. Next, the robot collects a set of observations, including tactile contacts, finger joint angles, and Cartesian positions for all tactile contacts, as elaborated in Section~\ref{observation}. Given such observations, the robot either 1) adjusts the joint angle of one or more fingers; 2) issues a ``reopen'' maneuver by reopening all fingers and adjusting the end-effector position and orientation; or 3) makes an attempt to lift the object, after which the episode terminates. In the case of 1) or 2), the episode moves forward to the next time step, and a new set of tactile observations are collected. Details of these three types of action primitives are elaborated in Section~\ref{action}. Through a simple reward structure (Section~\ref{reward}), the robot receives a higher reward for successfully picking an object up and a lower reward otherwise. Parameterized by a multi-modal network architecture and optimized through a soft surrogate objective (Section~\ref{softppo}) and curriculum learning (Section~\ref{curriculum}), the robot gradually learns to perform a series of actions that ultimately leads to a higher grasp success rate. Design rationale is elaborated in Appendix~\ref{sec:rationale}.

\subsection{Observation Space}\label{observation}
The robot's observation includes recent history and delta values ($\Delta$) of tactile contacts, finger joint angles, and Cartesian positions for all tactile contacts: $s_t = \{s_t^{contacts\_binary},  s_t^{\Delta contacts\_binary},  s_t^{joint\_angles},  s_t^{\Delta joint\_angles},  s_t^{contacts\_xyz}, s_t^{\Delta contacts\_xyz}\}$.

\textbf{Tactile Contacts.} 
The robot can observe the history of tactile contacts over the last 20 timesteps, which are binary indicators of whether each of the 96 tactile cells is activated or not: $s_t^{contacts\_binary} \in \{0, 1\}^{20 \times 96}$, as well as the delta in binary values between adjacent timesteps (19 delta values for each cell): $s_t^{\Delta contacts\_binary} \in \{-1, 0, 1\}^{19 \times 96}$ (details in Appendix~\ref{Contacts}).

\textbf{Finger Joint Angles.} 
The robot can observe the history of all joint angles of the hand over the last 20 timesteps, as well as whether the delta in values between adjacent timesteps exceeds a small threshold $\delta_{joint\_angle\_threshold} = 0.05$ rad. Since there are 8 joints for the Barrett Hand: $s_t^{joint\_angles} \in \mathbb{R}^{20 \times 8}, s_t^{\Delta joint\_angles} \in \{0, 1\}^{19 \times 8}$.

\textbf{Tactile Contact Cartesian Positions.} 
The robot can also observe the history of the $[x, y, z]$ Cartesian positions of all positive tactile contacts of the hand over the last 20 timesteps: $s_t^{contacts\_xyz} \in \mathbb{R}^{20 \times 96 \times 3}$, as well as the delta in values between adjacent timesteps: $s_t^{\Delta contacts\_xyz} \in \mathbb{R}^{19 \times 96 \times 3}$. The Cartesian positions are obtained via forward kinematics and expressed in the end-effector frame. If the tactile contact is not positive, the Cartesian position will be $[0, 0, 0]$ by default.

\subsection{Action Space}\label{action}
Given the observations elaborated in Section~\ref{observation}, the robot learns a function approximator $f$ that generates an action comprising of 1) movement of each of the $n$ fingers, 2) decision to reopen all fingers or not, 3) the position and orientation adjustments of the end-effector pose in the case of reopening, and 4) decision to lift the hand for a pick-up attempt or not: $a_t = \{a_t^{finger_1}, a_t^{finger_2}, ..., a_t^{finger_n}, a_t^{reopen}, a_t^{wrist\_rotation}, a_t^{lift}\}$.

\textbf{Finger Movements.} The robot can decide whether to close each finger further or not: $a_t^{finger_i} \in \{0, 1\}, \text{where } i \in [1, n] $. If $a_t^{finger_i} = 1$, the $i^{th}$ finger will close by a small joint angle delta of $\delta_{finger\_angle}$ (explained in Section~\ref{curriculum}). The robot samples each finger movement from an independent Bernoulli distribution given a learned sigmoid-activated parameter: $a_t^{finger_i} \sim \pi (a_t^{finger_i} \mid s_t) = \operatorname{Bern}(\operatorname{sigmoid}(f^{finger_i}(s_t)))$. 

\textbf{Finger Reopening and End-Effector Pose Adjustment.} The robot's action can control whether to reopen or not: $a_t^{reopen} \in \{0, 1\}$. The robot samples $a_t^{reopen}$ from a Bernoulli distribution given a sigmoid-activated parameter: $a_t^{reopen} \sim \pi (a_t^{reopen} \mid s_t) = \operatorname{Bern}(\operatorname{sigmoid}(f^{reopen}(s_t)))$. $a_t^{reopen}$ is 1 also when no joints moved above $\delta_{joint\_angle\_threshold} = 0.05 \text{ rad}$ over the past 5 timesteps. If the robot decides to reopen, each finger movement action $a_t^{finger_i}$ is disabled for this timestep.

During a reopen maneuver, all fingers reopen to the pre-grasp joint angles, and the position and orientation of the end-effector adjusts. During position adjustment, the hand's $[x, y]$ coordinates re-locate to above the most recent center of all active finger-palm tactile centers. Each active finger-palm tactile center is the center of all Cartesian locations of the finger-palm's active tactile cells (details elaborated in Appendix~\ref{Position}). During orientation adjustment, the robot's wrist is rotated by a learned angle to generate a better grasp. This learned angle ranges from $-180\degree \text{ to } 180\degree$: $a_t^{wrist\_rotation} \in [-\pi, \pi]$. To generate this angle, the robot samples from a Gaussian distribution whose mean is a learned, tanh-activated parameter and then scales the sampled value by factor $\pi$: $a_t^{wrist\_rotation} \sim \pi (a_t^{wrist\_rotation} \mid s_t) = \mathcal{N}(\operatorname{tanh}(f^{wrist\_rotation}(s_t)), \sigma_{rotation}) \times \pi$. Here, the Gaussian distribution's standard deviation $\sigma_{rotation}$ is also a learned parameter.

\textbf{Lifting.} The robot's action can control whether to lift the hand for a pick-up attempt or not: $a_t^{lift} \in \{0, 1\}$. The robot samples $a_t^{lift}$ from a Bernoulli distribution given a sigmoid-activated parameter: $a_t^{lift} \sim \pi (a_t^{lift} \mid s_t) = \operatorname{Bern}(\operatorname{sigmoid}(f^{lift}(s_t)))$. During a lift maneuver, the arm is lifted up vertically by 25cm, and each finger movement action $a_t^{finger_i}$ is disabled. If the robot decides to both reopen and lift, the reopen maneuver takes higher priority and is performed instead of the lift maneuver. If the last timestep of the episode is reached: $t_{final} = H$, lifting automatically occurs.

\subsection{Reward Structure}\label{reward}
After a lift maneuver, the current episode is terminated because the robot has decided to attempt to pick-up an object. At this last timestep $t_{final}$ of the episode, a binary success reward is collected, indicating whether the robot successfully picked an object up: $r_{t_{final}} = \mathds{1} \{\text{pick-up is successful}\}$. In all timesteps earlier than the last timestep, the reward is zero if the robot decides not to reopen. If the robot decides to reopen, a penalty of $-0.05$ is given if no fingers closed beyond 0.2 rad: $r_t = -0.05 \times a_t^{reopen} \times (1 - \mathds{1}\{\max_{i \in grip\_joint\_indices} [s_t^{joint\_angles}]_i > 0.2 \text{ rad} \})$, where $t \in [1, t_{final} - 1]$, which penalizes the robot against reopening fingers too frequently. 

\subsection{Soft Proximal Policy Optimization}\label{softppo}
\begin{figure*}
\includegraphics[width=\columnwidth]{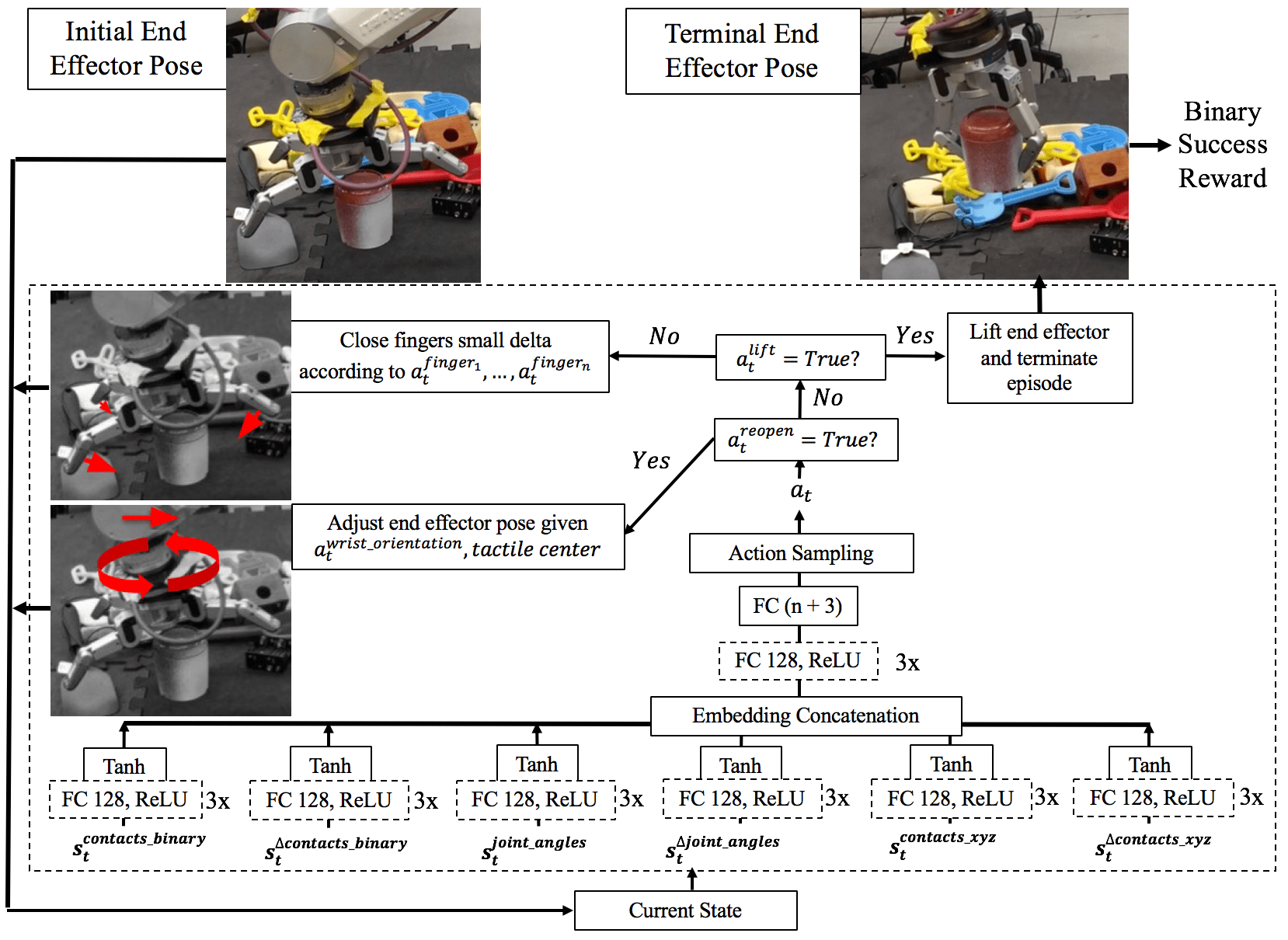}
\caption{\small \textbf{Multi-Fingered Adaptive Tactile Grasping.} Given an initial grasp pose obtained from a vision-based system, for example, the architecture uses soft proximal policy optimization to learn a grasping behavior. The state space consists of tactile contact readings (${s_t^{(\Delta)contacts\_binary}}$), contact Cartesian locations ($ s_t^{(\Delta)contacts\_xyz}$) and finger joint angles ($ {s_t^{(\Delta)joint\_angles}}$). Using a few deep neural networks, features are extracted from the six components of the state space, squashed to $[-1, 1]$ using $tanh$ activations, and finally concatenated into an embedding. This latent embedding is passed through another set of fully connected layers and then outputs a grasp action that specifies how to incrementally close each finger ($a_t^{finger_1}, \ldots a_t^{finger_n}$), whether to lift ($a_t^{lift}$), whether to reopen ($a_t^{reopen}$), and how to adjust the end-effector in case of reopening ($a_t^{wrist\_rotation}$).
The top portion of the figure shows how the robot's behavior is determined by $a_t$. First, it chooses to either continue current grasp or reopen and adjust the grasp. If it continues current grasp, it checks whether to lift, and if not, how to incrementally close the fingers.
A binary reward is obtained if lifting results in pick-up success. This reward, along with a small penalty for frequent reopening (Section~\ref{reward}), is the signal used to train the network in a deep RL manner. ``FC $m$, ReLU'' refers to a fully connected layer with output dimension of $m$ followed by a ReLU activation.}
\label{fig:architecture}
\vspace{-3mm}
\end{figure*}
Let $\theta$ be the parameter weights of the policy network and $\pi_\theta$ be the policy the robot is trying to learn: $\pi_\theta: \mathcal{S} \to \Pi(\mathcal{A})$. Shown in Figure~\ref{fig:architecture}, $\theta$ is a multi-input-branch deep neural network, where multiple observation modalities are handled by individual tanh-activated neural networks for feature extraction.
The robot's goal is to maximize the cumulative discounted sum of rewards: $\underset{\theta}{\text{maximize }}\mathbb{E}_{\pi_{\theta}}[\sum_{t}\gamma^{t-1} r_t]$. To begin, we follow the standard policy (SP) optimization objective:
\begin{equation}\label{sp_objective}
  \underset{\theta}{\text{maximize }} \mathcal{L}^{SP} = \mathbb{E}_{\rho_0, \pi_{\theta}}[\pi_{\theta}(a_t \mid s_t) Q_{\pi_{\theta}}(s_t, a_t)]
\end{equation}

Next, we introduce a baseline (BL) estimator parameterized by $\psi$ for state-value prediction and variance reduction, after which objective~\eqref{sp_objective} turns into~\cite{Schulman2015-nj}:
\begin{equation}
  \underset{\theta}{\text{maximize }} \mathcal{L}^{PG} = \mathbb{E}_{\rho_0, \pi_{\theta}}[\pi_{\theta}(a_t \mid s_t) \hat{A}_t]
\end{equation}
where $\hat{A}_t$ is an estimator of the advantage function~\cite{Baird1994-jl}. We optimize $\psi$ with the following loss:
\begin{equation}
  \underset{\psi}{\text{minimize }} \mathcal{L}^{BL} = \mathbb{E}\left[ \left\lVert V_{\psi} - V_{\pi_{\theta}} \right\rVert^2 \right]
\end{equation} Next, we substitute the action probability $\pi_{\theta}(a_t \mid s_t)$ with the Clipped Surrogate Objective~\cite{schulman2017proximal} and apply a soft advantage target to balance between exploration and exploitation~\cite{haarnoja2018soft}:
\begin{equation}
  \underset{\theta}{\text{maximize }} \mathcal{L}^{PG} = \mathbb{E}_{\rho_0, \pi_{\theta}}[\min (\lambda_t(\theta), \text{clip}(\lambda_t(\theta), 1-\epsilon, 1+\epsilon)) (\hat{A}_t - \alpha \log \pi_{\theta}(a_t \mid s_t))]
\end{equation}
where $\lambda_t(\theta) = \frac{\pi_{\theta}(a_t \mid s_t)}{\pi_{\theta_{old}}(a_t \mid s_t)}$. Hyperparameters are detailed in Appendix~\ref{sec:hyperparams}.

As shown in Figure~\ref{fig:architecture}, at every timestep, the robot decides whether to reopen fingers via $a_t^{reopen}$. If it does not reopen, $a_t^{lift}$ decides if it is safe to lift the object. If the robot does not reopen or lift, it decides how to close each finger~$a_t^{finger_i}$. If the current timestep reaches the finite horizon, lifting automatically occurs and no action component is effective. Therefore, the log action probability is: 
\begin{equation}
    \begin{split}
        \log \pi_{\theta}(a_t \mid s_t) &= [\log \pi_{\theta}(a_t^{reopen} \mid s_t) + (1-a_t^{reopen}) \times \log \pi_{\theta}(a_t^{lift} \mid s_t) \\&+ (1-a_t^{reopen}) \times (1-a_t^{lift}) \times \sum_{i=1}^{n} \log \pi_{\theta}(a_t^{finger_i} \mid s_t)] \times \mathds{1}\{t_{final} < H\}
    \end{split}
\end{equation}

\subsection{Curriculum Learning}\label{curriculum}
The finger-closing component of MAT is curriculum-learned. Compared to an open-loop approach that uniformly closes all fingers on the object and lifts after a preset time, MAT incrementally adjusts each finger and decides to lift when certain the object is in hand. An important parameter is the resolution or delta of the finger joint movement $\delta_{finger\_angle}$. In the extreme, closing all fingers by a large $\delta_{finger\_angle}$ reduces to the open-loop policy, however we desire a small $\delta_{finger\_angle}$ decoupled for each finger to achieve smooth grasping. The challenge is that no reward is received until a lift is attempted and for small $\delta_{finger\_angle}$ the reward signal will be too sparse. Initially during training, the policy decides to lift almost randomly: $\mathbb{E}[\pi_\theta (a_t^{lift} \mid s_t)] = 0.5$. Since picking an object up requires closing the fingers for many consecutive timesteps before lifting, the initial reward signal is extremely sparse under small $\delta_{finger\_angle}$. Empirically, learning becomes difficult when $\delta_{finger\_angle} < 0.4 \text{ rad}$ for the Barrett hand. Conversely, under large $\delta_{finger\_angle}$, the policy cannot control finer finger motions and becomes prone to failure.

To get benefits of both fine finger motions and short episode horizons, training in simulation is conducted using curriculum learning around $\delta_{finger\_angle}$. Initially, $\delta_{finger\_angle} = 0.4 \text{ rad}$. Subsequently, the joint angle delta is given by $\delta_{finger\_angle} = 0.1 + (0.4 - 0.1) \times (1 - current\_max\_success\_rate)$, where $current\_max\_success\_rate$ is the highest pick-up success rate that the robot has achieved thus far during training. This curriculum learning procedure allows the finger movements to become more and more granular and sophisticated as grasp success rate improves, effectively ``closing a tight loop'' for tactile grasping.

\section{Experiments}
We train MAT entirely in simulation and test in both simulation and real-world. During training, a single-object or multi-object cluttered scene is loaded with equal probability. We place one object in a single-object scene, a random number of objects from 2 to 30 for a simulated cluttered scene (Figure~\ref{fig:pbscene}), and 10 objects for a real-world cluttered scene. Leveraging the ShapeNet Repository~\cite{chang2015shapenet} in simulation, we use 200+ seen objects from the YCB and KIT datasets and 100+ novel objects (not seen during training) from the BigBIRD dataset. We train and test 500 grasp attempts per experiment in simulation. We evaluate real-world single-object performance across 10 trials for each of the 15 seen and novel objects, and real-world cluttered scene performance across 15 scenes. Appendix~\ref{sec:stattest} outlines how statistical significance is tested based on the numerical results.

\subsection{Results and Discussions}
\setlength\tabcolsep{5pt}
\begin{table}[H]
\vspace{-5mm}
\centering
\caption{Experimental Results (\% Grasp Success $\pm$ Standard-Dev)}
\label{tab:results}
\begin{tabular}{|c|cc|cc|}
\hline
 \multicolumn{1}{|c} {} & \multicolumn{2}{c|}{Single Object} & \multicolumn{2}{|c|} {Cluttered Scene}\\
\multicolumn{1}{|c} {Objects} & \multicolumn{1}{c} {Seen} & \multicolumn{1}{c|} {Novel} & \multicolumn{1}{c} {Seen} & \multicolumn{1}{c|} {Novel}\\ \hline
& \multicolumn{4}{|c|}{Simulation} \\ \hline
MAT & \textbf{98.2 $\pm$ 2.1} & \textbf{97.4 $\pm$ 1.6} & \textbf{97.7 $\pm$ 2.9} & \textbf{95.9 $\pm$ 3.9}\\
Open-Loop Baseline \cite{wu2019pixeliros} & 93.8 $\pm$ 2.6 & 94.9 $\pm$ 1.4 & 92.5 $\pm$ 1.8 & 91.1 $\pm$ 3.7 \\\hline
& \multicolumn{4}{|c|}{Real} \\ \hline
MAT & \textbf{98.7 $\pm$ 3.5} & \textbf{98.0 $\pm$ 4.1} & \textbf{96.4 $\pm$ 4.6} & \textbf{95.8 $\pm$ 4.7}\\
Open-Loop Baseline \cite{wu2019pixeliros} & 96.7 $\pm$ 6.2 & 93.3 $\pm$ 8.1 & 92.9 $\pm$ 5.8 & 91.9 $\pm$ 6.7\\ \hline
\end{tabular}
\vspace{-3mm}
\end{table}
Table~\ref{tab:results} compares 1) a high success rate (above 90\%) baseline open-loop vision-based approach~\cite{wu2019pixeliros} with 2) MAT using initial 6-DOF grasp pose provided by~\cite{wu2019pixeliros}, and reports the percentage of grasp success and standard deviation across scenes.
We fairly evaluate MAT against \cite{wu2019pixeliros} using the same robot setup, objects, and cluttered scenes.
The simulation results show that MAT gives a statistically significant improvement in grasp success rates in all cases: $4.4\%$, $2.5\%$, $5.2\%$, $4.8\%$ for single-seen, single-novel, cluttered-seen, cluttered-novel respectively. On the other hand, the real-world results reveal statistically similar grasp success rates compared to simulation, showing high-fidelity sim-to-real transfer.
While the vision-only grasping system already gives high success rates, MAT gives further improvement and is able to avoid or recover from failure cases that the vision-only system cannot handle. For example, MAT uses the tactile readings to finely control how the robot incrementally closes each finger and ensure that it does not lift the object until it is sure it has a firm grip. Also, MAT is able to re-generate a new grasp pose if the tactile readings suggest that the current grasp being executed would not result in a successful pick-up. Appendix~\ref{sec:ablation} details extensive ablation studies and comparison with a tactile baseline.

\subsection{Grasping under Calibration Noise}
\begin{wrapfigure}{r}{0.4\textwidth}
  \vspace{-10mm}
  \begin{center}
    \includegraphics[width=0.4\textwidth]{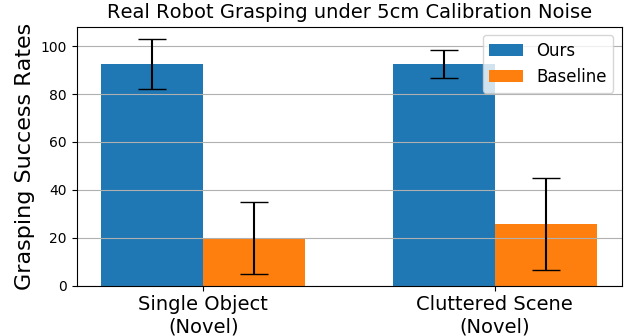}
  \end{center}
  \caption{\small Real-World Grasping of Novel Objects with 5cm Y-Axis Calibration Noise}
  \label{fig:results_calib_error_real}
  \vspace{-5mm}
\end{wrapfigure}
We observe robustness properties of MAT that are valuable if there is calibration error in the grasping setup-- a common occurrence in robotics.  To show this, we introduce varying levels of calibration noise to the grasping setup and measure how this affects the pick-up success rate. Calibration noise is added as an offset of $\delta$cm to the generated 6-DOF grasp pose (obtained from~\cite{wu2019pixeliros}) before grasp execution. 
On the real robot, we set the calibration noise to be $\delta = 5$cm and run the grasping experiments for the novel objects in both single-object and cluttered scene settings; the results are presented in  Figure~\ref{fig:results_calib_error_real}.
In simulation, we repeat the experiments with $\delta = 2.5, 5, 7.5$cm to analyze the performance across different noise levels as well as to compare between seen and novel objects.
Figure~\ref{fig:results_calib_error_sim} visualizes the simulation results, with detailed numerical results in Appendix~\ref{sec:results_table_appendix}.

The results show that the tactile policy is significantly more robust to calibration error compared to the vision-only system. On the real robot, the grasp success rate of the vision-only baseline~\cite{wu2019pixeliros} degrades under calibration noise to 20.0\% \& 25.6\% (single-object \& cluttered scenes), while our tactile-based policy still achieves 90\%+ success in both cases (Figure~\ref{fig:results_calib_error_real}).
Note that repeated trials do not increase the ability of the vision-only system to recover as the calibration stays an issue for such systems; as a result we terminate each run after 3 consecutive failed attempts per object or scene. For the novel cluttered scenes, the vision-only method picks up only a quarter of the objects present in the scenes and is unable to fully clear any of the scenes.
In simulation on the other hand, results in Figure~\ref{fig:results_calib_error_sim} reveal a degradation in performance for the vision-based system~\cite{wu2019pixeliros} as calibration noise increases; conversely, the performance of MAT stays robustly high.

\begin{figure}[!ht]
    \vspace{-5pt}
    \centering
    \begin{subfigure}{0.325\textwidth}
        \includegraphics[width=\linewidth]{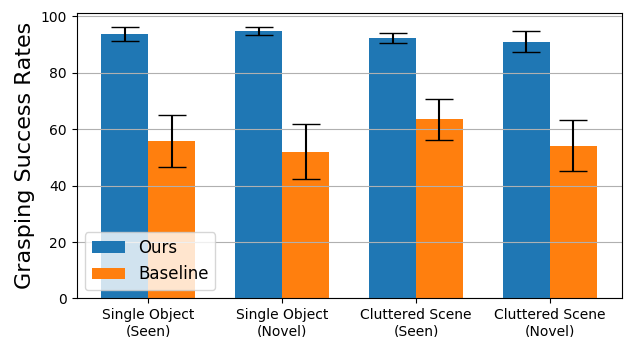}
        \caption{Calibration Noise: 2.5 cm}
    \end{subfigure}
    \begin{subfigure}{0.325\textwidth}
        \includegraphics[width=\linewidth]{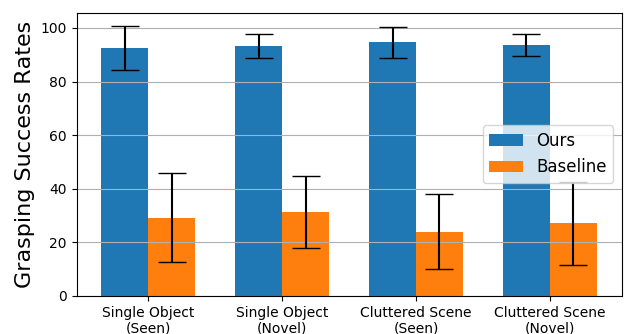}
        \caption{Calibration Noise: 5.0 cm}
    \end{subfigure}
    \begin{subfigure}{0.325\textwidth}
        \includegraphics[width=\linewidth]{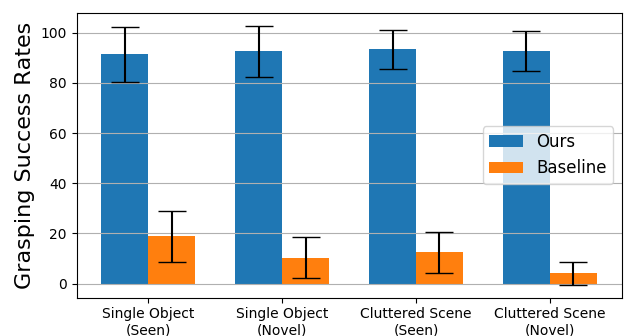}
        \caption{Calibration Noise: 7.5 cm}
    \end{subfigure}
    \caption{\small{\textbf{Grasping in Simulation under Calibration Noise.} MAT (blue) shows robustness under increasing calibration noise compared to a strong vision-only baseline~\cite{wu2019pixeliros} (orange) which degrades significantly.}}
    \label{fig:results_calib_error_sim}
    \vspace{-10pt}
\end{figure}

\section{Conclusion and Future Work}
This work presents MAT, an adaptive, closed-loop tactile algorithm to reinforcement-learn dexterous robotic grasping for multi-fingered hands with noisy tactile readings and no visual information. Entirely trained in simulation, MAT achieves mid-to-high 90\% success rates and significantly improves upon open-loop grasping under calibration error. As many vision-based grasping systems assume and require efforts to achieve near-perfect calibration, the robustness of MAT even under considerable calibration error reduces the need for a perfectly calibrated grasping setup. MAT can be easily added to any existing open-loop grasp planner to close the final gap between failed grasps and success. In the future, we hope to extend MAT to dexterous, tactile-based object manipulation.

{\footnotesize
\bibliography{references.bib}}
\newpage
\appendix
\section{Appendix}
\subsection{MAT Design Rationale}
\label{sec:rationale}
\subsubsection{Planning as opposed to learning wrist repositioning}
MAT essentially relocates the palm to the center of all finger tactile centers, which is very intuitive because these locations give direct hints on where the object is. If humans were to reposition the wrist given these sparse tactile locations, they would have relocated the wrist close to the tactile center as well. This makes MAT's planning strategy hard to beat using learning. We did implement learning wrist repositioning early in research. However, we later discovered that learning has empirically underperformed planning due to the observation stated earlier.

\subsubsection{Adopting discrete finger closing rather than continuous motor control}
Empirically, adopting continuous motion control creates a large dynamics sim-to-real gap. In contrast, discrete finger closing makes simulated and real-world finger actions identical and creates virtually no sim-to-real gap. Furthermore, our curriculum learning of the finger closing procedure in Section~\ref{curriculum} in the paper attempts to make discrete finger closing similar to continuous motor control.

\subsubsection{4-DOF as opposed to 6-DOF grasp pose adjustments}
MAT only adjusts the wrist rotation (the roll component in Roll-Pitch-Yaw) due to two important reasons. First, adjusting the pitch and yaw correctly requires accurate perception of the target object pose. While sparse tactile information gives hint on where the object is, it does not say much about the object pose. Adjusting the wrist rotation in the absence of object pose perception, on the other hand, is still useful since it allows the fingers to explore previously unvisited cartesian spaces. More practically, changing the pitch-yaw orientation in cluttered scenes has been empirically dangerous in our experimental setup since at least one finger will move downward, possibly collide with the clutter, and break. 

Collision is detected by checking if the lateral-spread of the Barrett hand is changing involuntarily, indicating that the table or the objects are colliding with the hand sideways. Collision is also detected if any tactile cells or fingers on the Barrett hand are experiencing unusually significant forces.
To protect the hand from breaking due to collision with cluttered objects, MAT lifts the end-effector incrementally by 0.1cm until the hand is free of collision. This protection mechanism is triggered only during the finger-reopening and the end-effector pose adjustment phases. In other words, during the finger-closing phase and the lifting phase, this mechanism is disabled since virtually no collision is possible. 

However, this does not mean that MAT grasp poses will always be top-down. MAT is designed to work on top of visual grasping algorithms that generate initial 6-DOF grasp poses. The initial vision system~\cite{wu2019pixeliros} we used in experiments is a 6-DOF grasp pose algorithm that also generates non-top-down grasps. As a result, even though MAT performs 4-DOF grasp adjustments, the resulting 6-DOF grasp pose can still be non-top-down. 

\subsection{Statistical Significance Testing Mechanism}
\label{sec:stattest}
In this paper, we presented mean and standard deviation statistics for each experiment, which enabled us to examine whether the difference in performance between different experiments are statistically significant or not. We consider the performance difference statistically significant if the difference in \textbf{mean} performance of the two experiments are at least 1 standard deviation away. For example, in Table~\ref{tab:results} in the paper, the performances for single-object simulation experiment between MAT ($98.2 \pm 2.1\%$) and Open-Loop Baseline \cite{wu2019pixeliros} ($93.8 \pm 2.6\%$) are statistically significantly different (as opposed to similar) because the difference in mean ($4.4\%$) is higher than the standard deviation of either statistics ($2.1\%$ or $2.6\%$).

\subsection{Lower Real-World Statistical Significance due to High Standard Deviation}
\label{sec:realss}
In Table~\ref{tab:results} in the paper, the real-world results of MAT compared to Open-Loop Baseline~\cite{wu2019pixeliros} are less statistically significant mainly due to \textbf{higher standard deviations} as a result of limited number of objects and scenes. The mean statistics however stay robustly high. The main purpose of real-world results is to prove high-fidelity sim-to-real transfer. In contrast, the simulation experiments are more suitable for statistical tests since they use well recognized datasets (ShapeNet~\cite{chang2015shapenet}), many more objects (200+ seen and 100+ novel objects), many more trials (500 trials per experiment), and a uniformly distributed clutter level.

\subsection{Extensive Tactile Baseline and Ablation Experiments}
\label{sec:ablation}
\begin{table}[H]
\vspace{-5mm}
\centering
\caption{Ablation and Tactile Baseline Results (\% Grasp Success $\pm$ Standard-Dev)}
\label{tab:ablation}
\begin{tabular}{|c|c|cc|cc|}
\hline
 & & \multicolumn{2}{c|}{Single Object} & \multicolumn{2}{|c|} {Cluttered Scene}\\
Objects & Noise & \multicolumn{1}{c} {Seen} & \multicolumn{1}{c|} {Novel} & \multicolumn{1}{c} {Seen} & \multicolumn{1}{c|} {Novel}\\ \hline
& & \multicolumn{4}{|c|}{Simulation} \\ \hline
MAT & \multirow{2}{*}{0cm} & \textbf{98.2 $\pm$ 2.1} & \textbf{97.4 $\pm$ 1.6} & \textbf{97.7 $\pm$ 2.9} & \textbf{95.9 $\pm$ 3.9}\\
Tactile Baseline \cite{jentoft2014limits} &  &94.4 $\pm$ 2.3 & 95.0 $\pm$ 1.9 & 93.3 $\pm$ 1.0 & 91.3 $\pm$ 3.1 \\ \hline
& & \multicolumn{4}{|c|}{Ablation (Simulation)} \\ \hline
Finger-Closing Only & \multirow{4}{*}{0cm} &96.6 $\pm$ 2.9 & 96.2 $\pm$ 2.6& 95.3 $\pm$ 1.1 & 94.7 $\pm$ 3.6\\
Regrasping Only &  &96.2 $\pm$ 1.8 & 96.0 $\pm$ 1.8 & 94.4 $\pm$ 1.8 & 93.5 $\pm$ 1.6  \\
Position Adjustment Only &  &96.9 $\pm$ 3.8 & 96.4 $\pm$ 4.6 & 96.2 $\pm$ 4.0 & 94.8 $\pm$ 3.5\\ 
Orientation Adjustment Only &  &96.7 $\pm$ 3.4 & 96.4 $\pm$ 2.8 & 95.7 $\pm$ 4.6 & 94.8 $\pm$ 1.3\\ \hline
& & \multicolumn{4}{|c|}{Simulation} \\ \hline
MAT & \multirow{2}{*}{2.5cm} & \textbf{92.5 $\pm$ 8.1} & \textbf{93.3 $\pm$ 7.2} & \textbf{94.6 $\pm$ 5.8} & \textbf{93.7 $\pm$ 4.2} \\
Tactile Baseline  \cite{jentoft2014limits} &  & 64.1 $\pm$ 5.9 & 68.2 $\pm$ 5.6 & 66.1 $\pm$ 5.9 & 66.8 $\pm$ 4.6 \\ \hline
& & \multicolumn{4}{|c|}{Ablation (Simulation)} \\ \hline
Finger Closing Only & \multirow{4}{*}{2.5cm} & 75.2 $\pm$ 6.0 & 73.1 $\pm$ 5.3 & 73.6 $\pm$ 5.2 & 73.9 $\pm$ 4.8 \\
Regrasping Only &  & 75.9 $\pm$ 2.6 & 78.5 $\pm$ 3.7 & 76.0 $\pm$ 2.0 & 74.4 $\pm$ 2.0 \\
Position Adjustment Only &  & 76.9 $\pm$ 5.5 & 78.5 $\pm$ 7.2 & 80.4 $\pm$ 3.5 & 75.8 $\pm$ 7.5 \\
Orientation Adjustment Only &  & 81.2 $\pm$ 6.4 & 81.6 $\pm$ 6.4 & 83.1 $\pm$ 4.3 & 77.8 $\pm$ 6.0 \\\hline
\end{tabular}
\vspace{-3mm}
\end{table}

\subsubsection{Tactile baseline comparison}
To the best of our knowledge, MAT is among the first tactile methods that are multi-fingered, free of force-torque usage and compliance, and experimentally tested on cluttered scenes. This unfortunately makes most related works less suitable as a baseline. For example, \cite{hsiao2010contact} focuses on parallel-jaw hands. \cite{felip2012contact} and \cite{felip2013manipulation} require force-torque sensor in both the alignment phase and the sliding grasp phase. In addition, their force adaptation phases did not specify the numerical threshold values for joint angles and tactile contact forces used to detect grasp success versus failure, making them less feasible to re-implement fairly. Lastly, \cite{jentoft2014limits} requires compliance. Nevertheless, we re-implemented the non-compliant version of \cite{jentoft2014limits} as a tactile baseline comparison, in which each finger stops upon detecting initial contact and then all fingers close in unison after all are in contact. Lifting is subsequently performed after a certain period of time.

By comparing Row ``MAT'' to Row ``Tactile Baseline \cite{jentoft2014limits}'' for 0cm calibration noise in Table~\ref{tab:ablation}, we observed statistically significant improvement of MAT's performance over the tactile baseline.

By comparing Row ``MAT'' to Row ``Tactile Baseline \cite{jentoft2014limits}'' for 2.5cm calibration noise in Table~\ref{tab:ablation}, we observed a substantially more significant improvement of MAT's performance over the tactile baseline, mainly due to MAT's ability to adjust the robot's end-effector pose.

\subsubsection{Closed-loop vision baseline comparison}
Some closed-loop vision-only baselines \cite{kalashnikov2018qt}\cite{levine2018learning} seem comparable to MAT. However, both works focus on real-world learning with parallel-jaw robots while MAT focuses on multi-fingered hands and can train purely in simulation. Extending \cite{kalashnikov2018qt} and \cite{levine2018learning} to 6-DOF grasping and multi-fingered hands will increase real-world sample complexity and make real-world robot learning less tractable.

\subsubsection{Ablation studies to disentangle different components of MAT}
We also performed a series of ablation experiments to answer the following valuable questions.

\textbf{How much benefit is provided by just using tactile feedback during closing?} 

By comparing Row ``MAT'' to Row ``Finger Closing Only'' for 0cm calibration noise in Table~\ref{tab:ablation}, we observe that only using finger closing degrades MAT success rate in simulation by 1.6\%, 1.2\%, 2.4\%, 1.2\% for single-seen, single-novel, cluttered-seen, cluttered-novel. 

By comparing Row ``MAT'' to Row ``Finger Closing Only'' for 2.5cm calibration noise in Table~\ref{tab:ablation}, we see that the performance degradations are much higher: 17.3\%, 20.2\%, 21.0\%, 19.8\% respectively.

\textbf{How much benefit is provided by pure regrasping?}

By comparing Row ``MAT'' to Row ``Regrasping Only'' for 0cm calibration noise in Table~\ref{tab:ablation}, we observe that only using regrasping degrades MAT success rate in simulation by 2.0\%, 1.4\%, 3.3\%, 2.4\% for single-seen, single-novel, cluttered-seen, cluttered-novel. 

By comparing Row ``MAT'' to Row ``Regrasping Only'' for 2.5cm calibration noise in Table~\ref{tab:ablation}, we see that the performance degradations are much higher: 16.6\%, 14.8\%, 18.6\%, 19.3\% respectively.

\textbf{How does learning just a new position compare to learning both new position and orientation instead?}

By comparing Row ``MAT'' to Row ``Position Adjustment Only'' for 0cm calibration noise in Table~\ref{tab:ablation}, we observe that disabling orientation adjustment degrades MAT success rate in simulation by 1.3\%, 1.0\%, 1.5\%, 1.1\% for single-seen, single-novel, cluttered-seen, cluttered-novel. 

By comparing Row ``MAT'' to Row ``Position Adjustment Only'' for 2.5cm calibration noise in Table~\ref{tab:ablation}, we see that the performance degradations are much higher: 15.6\%, 14.8\%, 14.2\%, 17.9\% respectively.

\textbf{How does learning just a new orientation compare to learning both new position and orientation instead?}

By comparing Row ``MAT'' to Row ``Orientation Adjustment Only'' for 0cm calibration noise in Table~\ref{tab:ablation}, we observe that disabling position adjustment degrades MAT success rate in simulation by 1.5\%, 1.0\%, 2.0\%, 1.1\% for single-seen, single-novel, cluttered-seen, cluttered-novel. 

By comparing Row ``MAT'' to Row ``Orientation Adjustment Only'' for 2.5cm calibration noise in Table~\ref{tab:ablation}, we see that the performance degradations are much higher: 11.3\%, 11.7\%, 11.5\%, 15.9\% respectively.

\subsection{Significance of Calibration Noise Experimental Results}
While the experimental results demonstrate the robustness of MAT under various levels of calibration noises, it is important to point out that the calibration noise experiments reflect the same types of symptoms caused by other similarly common challenges in grasping, such as:
\begin{itemize}
    \item suboptimality in the grasp pose generated by a learning or planning method
    \item partial observability of the 3D point cloud under a monocular camera setup
    \item inaccurate object pose prediction by an object pose estimation algorithm
    \item perceptual difficulties dealing with transparent or reflective objects
    \item performance degradation caused by low-fidelity sim-to-real transfer
    \item unexpected object pose disturbance
\end{itemize}

\subsection{Grasping in Simulation under Calibration Noise}
\label{sec:results_table_appendix}
Table \ref{tab:results_calib_error_sim} shows the grasp success rate of MAT compared to a strong vision baseline, at varying calibration noises. In simulation, we generate 100 different grasping scenes in each category-- both for seen and novel objects as well as for single-object and cluttered-scene scenarios. Calibration noise is introduced in a random direction on the X-Y plane. We record the number of pick-ups per experiment and report the grasp success rates. The results show that the performance of MAT stays high even under significant calibration noise. The performance of the vision-only baseline~\cite{wu2019pixeliros} suffers increasingly and considerably as the calibration noise increases. The robustness demonstrated by MAT can be used to augment any vision-based system.
\begin{table}[H]
\vspace{-3mm}
\centering
\caption{Grasping in Simulation under Calibration Noise (\% Grasp Success $\pm$ Standard-Dev)}
\label{tab:results_calib_error_sim}
\begin{tabular}{|c|c|cc|cc|}
\hline
 \multicolumn{1}{|c} {} & \multicolumn{1}{|c} {} & \multicolumn{2}{|c|}{Single Object} & \multicolumn{2}{|c|} {Cluttered Scene}\\
\multicolumn{1}{|c} {Objects} & \multicolumn{1}{|c|} {Noise} & \multicolumn{1}{c} {Seen} & \multicolumn{1}{c|} {Novel} & \multicolumn{1}{c} {Seen} & \multicolumn{1}{c|} {Novel}\\ \hline
MAT & 2.5cm & \textbf{92.5 $\pm$ 8.1} & \textbf{93.3 $\pm$ 7.2} & \textbf{94.6 $\pm$ 5.8} & \textbf{93.7 $\pm$ 4.2} \\
Open-Loop Baseline \cite{wu2019pixeliros} & 2.5cm & 55.8 $\pm$ 9.1 & 52.1 $\pm$ 9.7 & 63.5 $\pm$ 7.2 & 54.2 $\pm$ 9.1 \\\hline
MAT & 5cm & \textbf{91.4 $\pm$ 11.0} & \textbf{92.5 $\pm$ 10.2} & \textbf{93.3 $\pm$ 7.8} & \textbf{92.7 $\pm$ 8.0} \\
Open-Loop Baseline \cite{wu2019pixeliros} & 5cm & 29.2 $\pm$ 16.7 & 31.3 $\pm$ 13.6 & 24.0 $\pm$ 14.1 & 27.1 $\pm$ 15.5 \\\hline
MAT & 7.5cm & \textbf{82.5 $\pm$ 9.5} & \textbf{80.8 $\pm$ 12.3} & \textbf{85.4 $\pm$ 9.5} & \textbf{81.3 $\pm$ 9.0} \\
Open-Loop Baseline \cite{wu2019pixeliros} & 7.5cm & 18.8 $\pm$ 10.3 & 10.4 $\pm$ 8.0 & 12.5 $\pm$ 8.1 & 4.2 $\pm$ 4.6 \\\hline
\end{tabular}
\end{table}

\subsection{Real-Robot Grasping under Calibration Noise}
On the real robot, we generate 7 different grasping scenes using novel objects for single-object and cluttered-scene scenarios.
Table \ref{tab:results_calib_error_real} shows that the grasp success rate of MAT stays high ($>$90\%) under calibration noise of 5cm compared to a strong vision-only baseline,  whose performance drops significantly from 93.3\% to 20.0\% and 91.9\% to 25.6\% for single-object and cluttered-scene scenarios respectively. 
\begin{table}[H]
\centering
\caption{Real-World Grasping of Novel Objects with 5cm Y-Axis Calibration Noise (\% Grasp Success $\pm$ Standard-Dev)}
\label{tab:results_calib_error_real}
\begin{tabular}{|c|c|c|}
\hline
& Single Object & Cluttered Scene\\\hline
MAT & \textbf{92.5 $\pm$ 10.4} & \textbf{92.4 $\pm$ 5.9} \\
Open-Loop Baseline \cite{wu2019pixeliros}  & 20.0 $\pm$ 15.1 & 25.6 $\pm$ 19.2 \\ \hline
\end{tabular}
\end{table}
\subsection{Real-World Experimental Objects}
\begin{figure}[H]
    \centering
    \includegraphics[trim={0 0 0 50},clip, width=\linewidth]{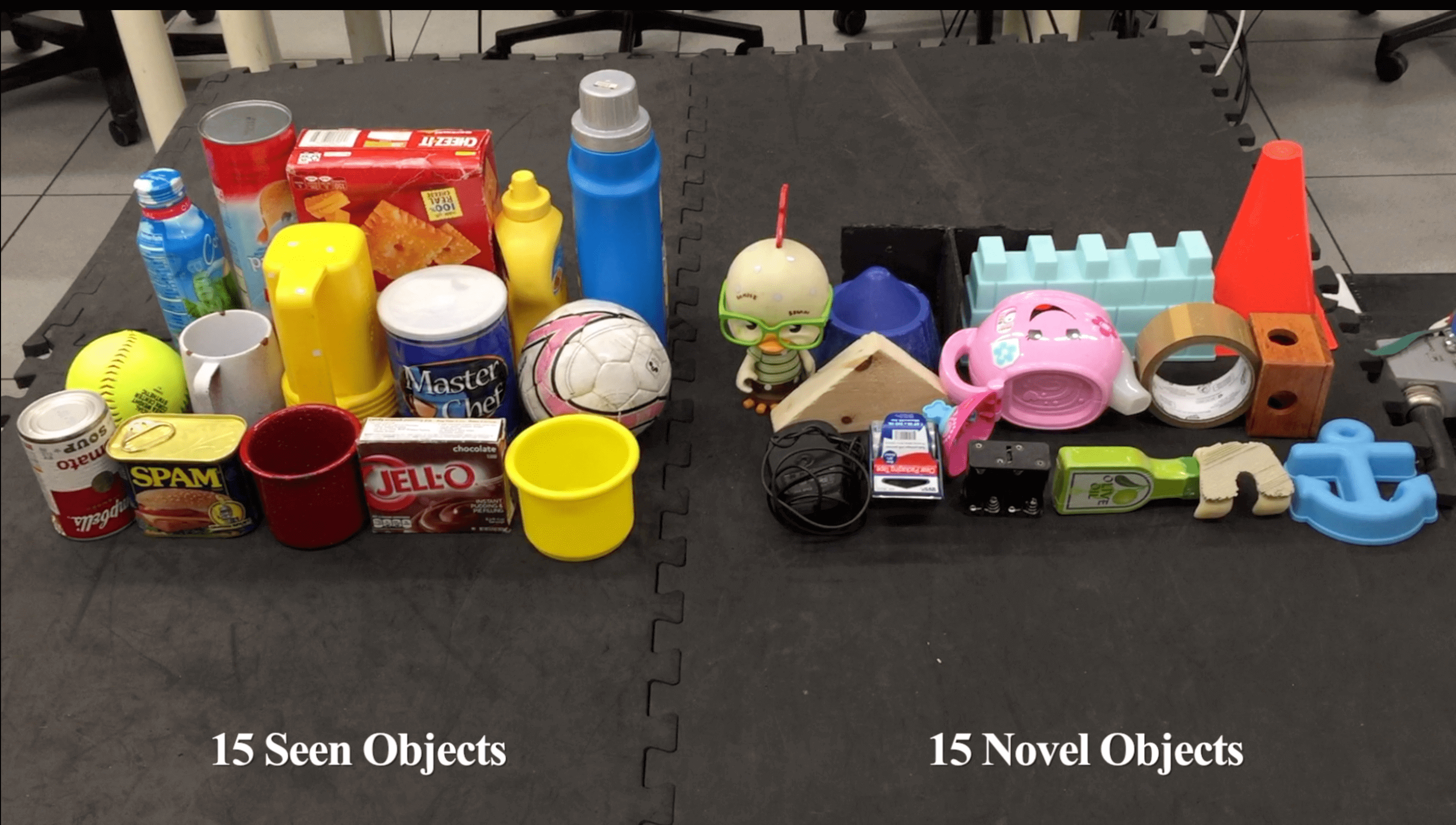}
    \caption{\textbf{15 Seen and 15 Novel Objects Used in Real-World Experiments.} Left half of image: 15 seen objects. Right half of image: 15 novel objects.}
    \label{fig:objects}
\end{figure}

Figure~\ref{fig:objects} displays the 15 seen and 15 novel objects used in real-world experiments.
\subsection{Learned Closed-Loop Behaviors from MAT}
\begin{figure}[H]
    \centering
    \begin{subfigure}[t]{0.16\textwidth}
        \captionsetup{skip=0pt}\includegraphics[trim={675 0 50 0},clip,width=\linewidth]{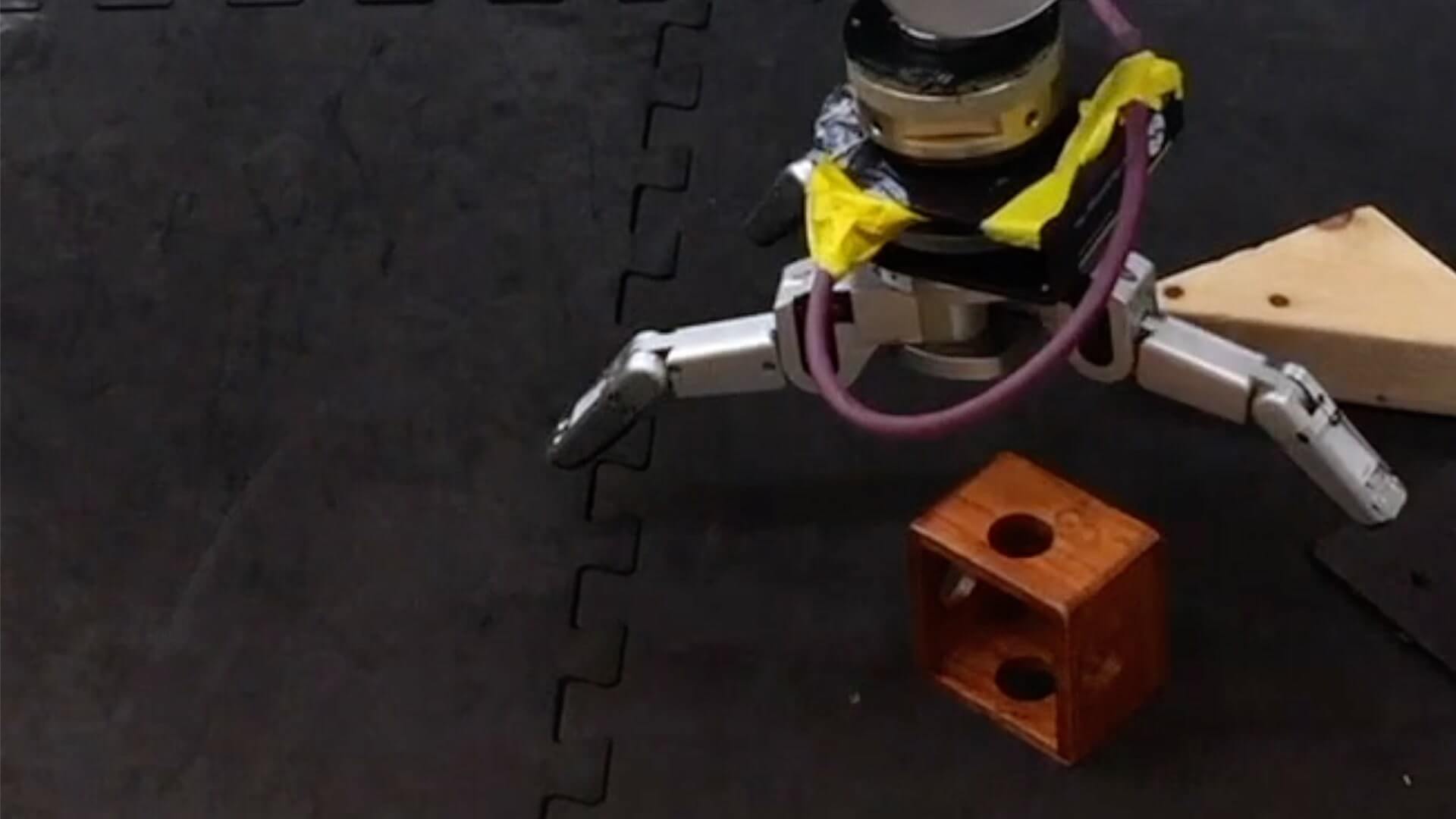}\caption{}
    \end{subfigure}
    \begin{subfigure}[t]{0.16\textwidth}
        \captionsetup{skip=0pt}\includegraphics[trim={675 0 50 0},clip,width=\linewidth]{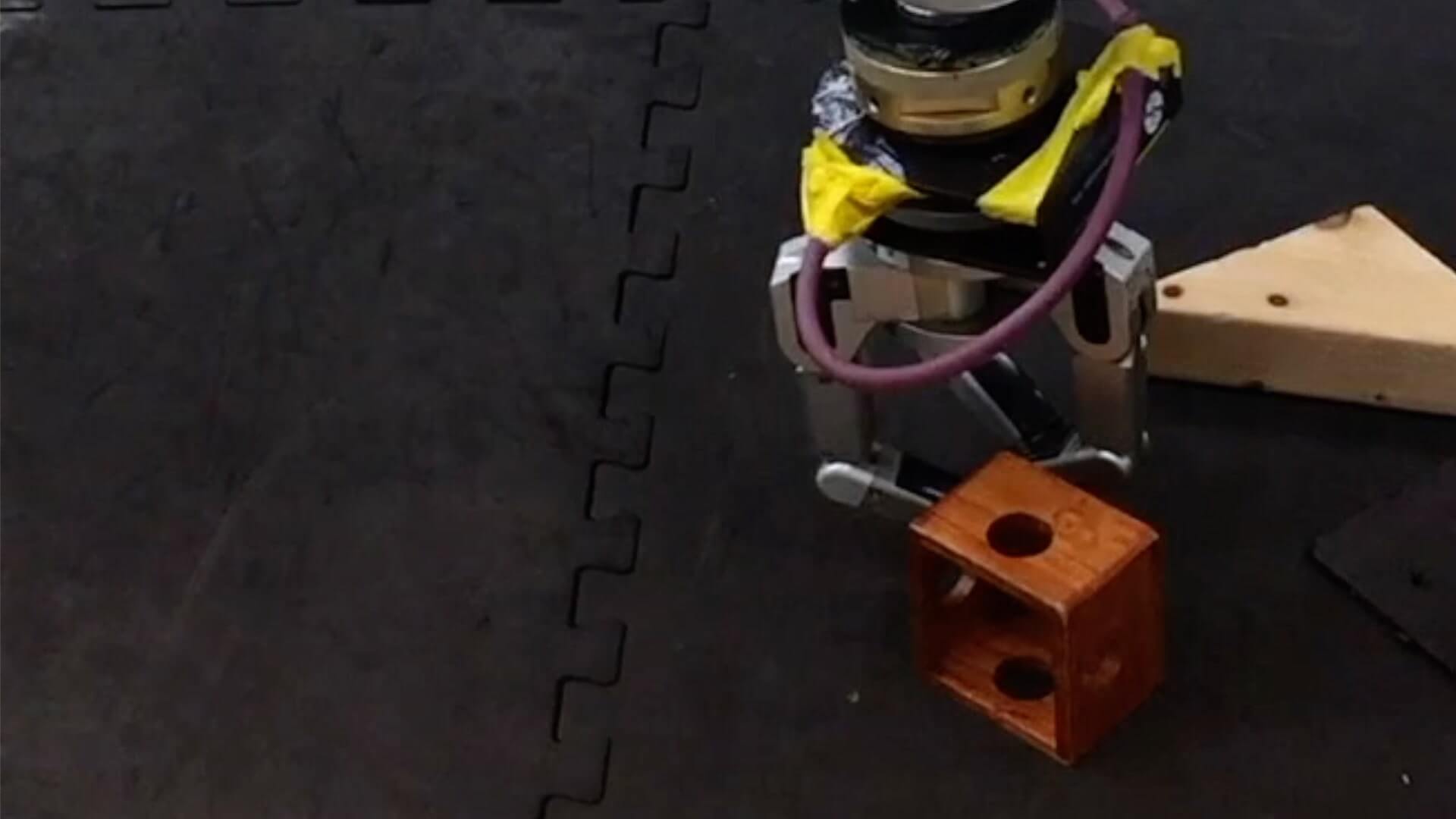}\caption{}
    \end{subfigure}
    \begin{subfigure}[t]{0.16\textwidth}
        \captionsetup{skip=0pt}\includegraphics[trim={675 0 50 0},clip,width=\linewidth]{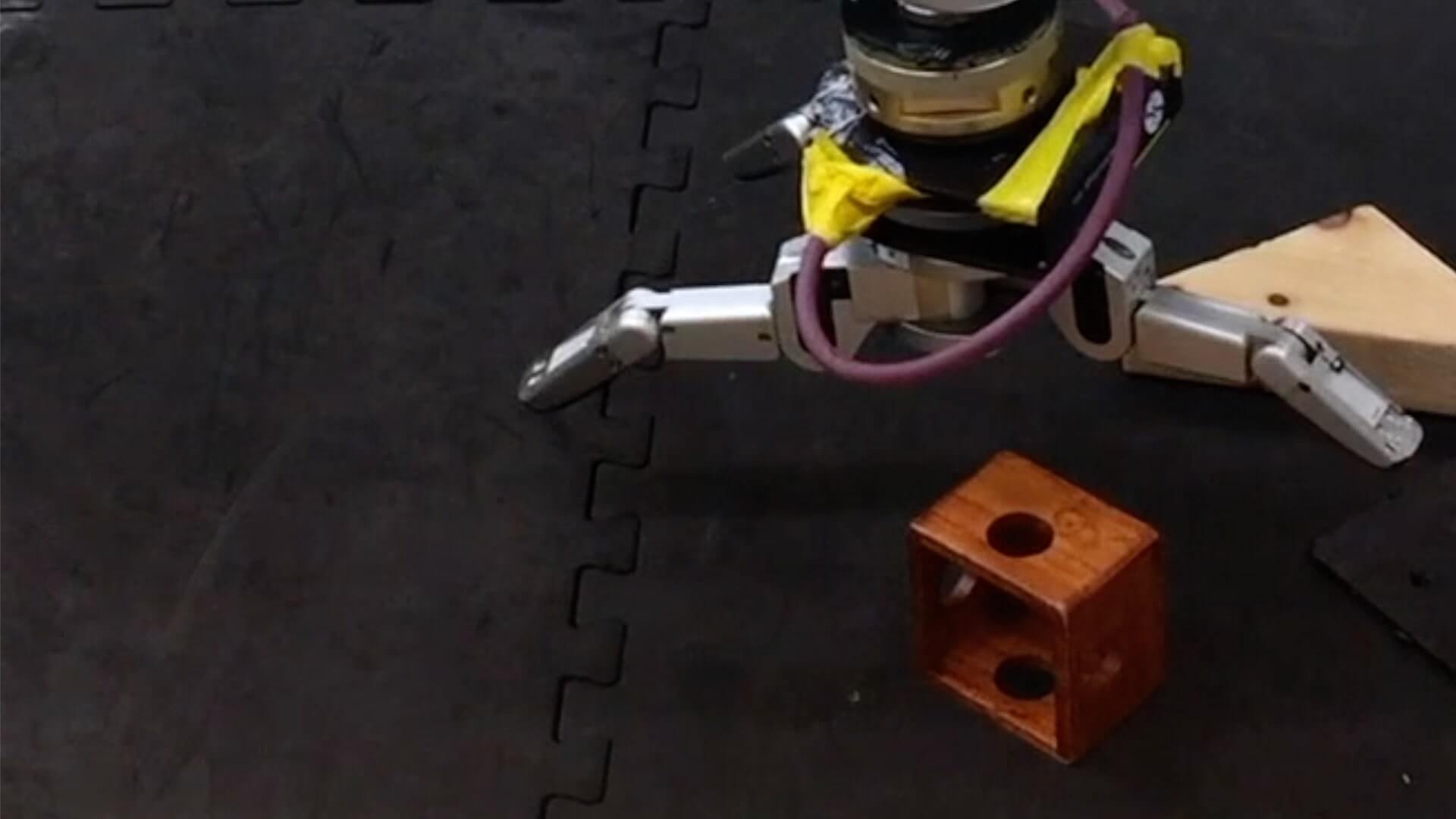}\caption{}
    \end{subfigure}
    \begin{subfigure}[t]{0.16\textwidth}
        \captionsetup{skip=0pt}\includegraphics[trim={675 0 50 0},clip,width=\linewidth]{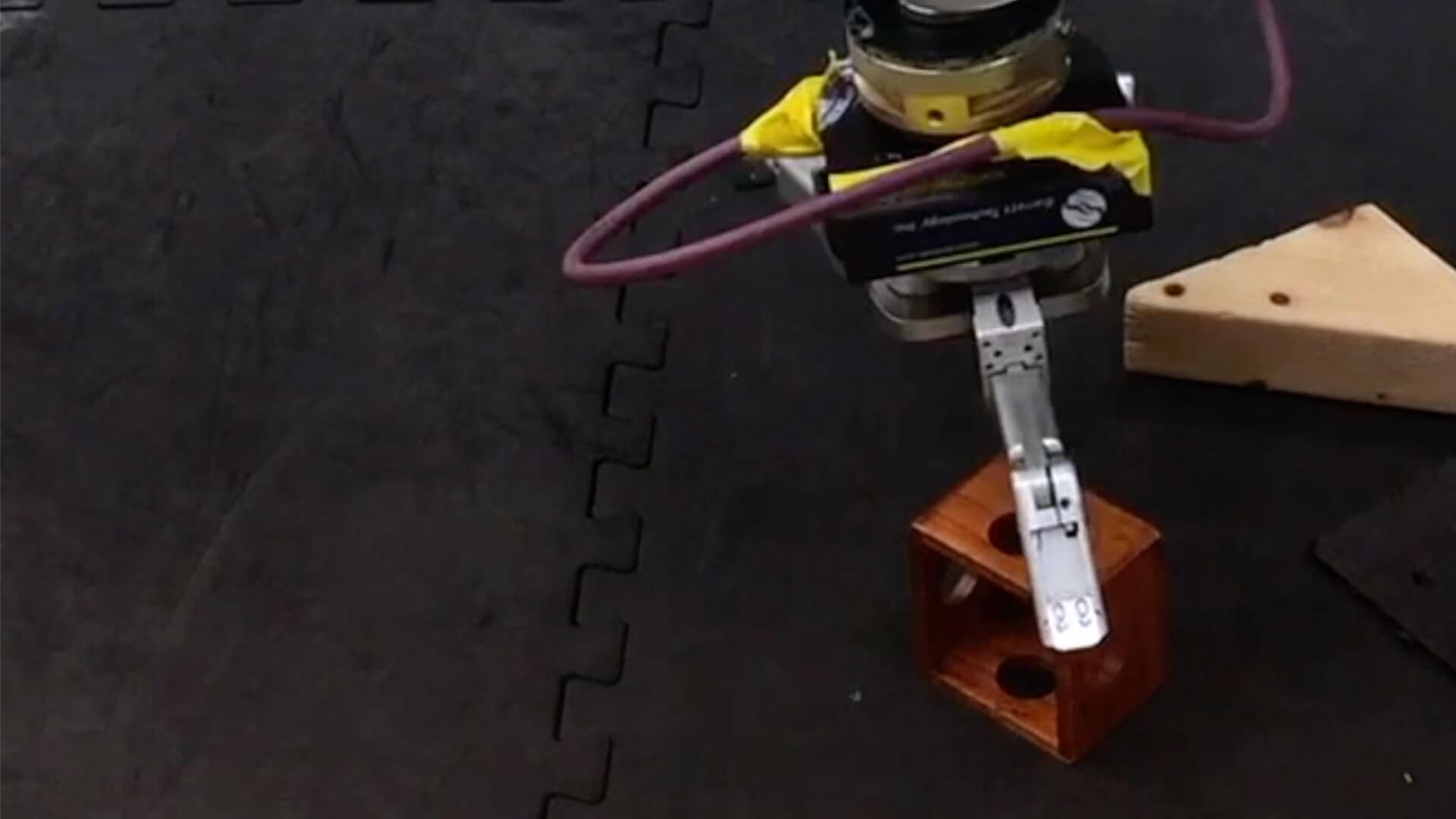}\caption{}
    \end{subfigure}
    \begin{subfigure}[t]{0.16\textwidth}
        \captionsetup{skip=0pt}\includegraphics[trim={675 0 50 0},clip,width=\linewidth]{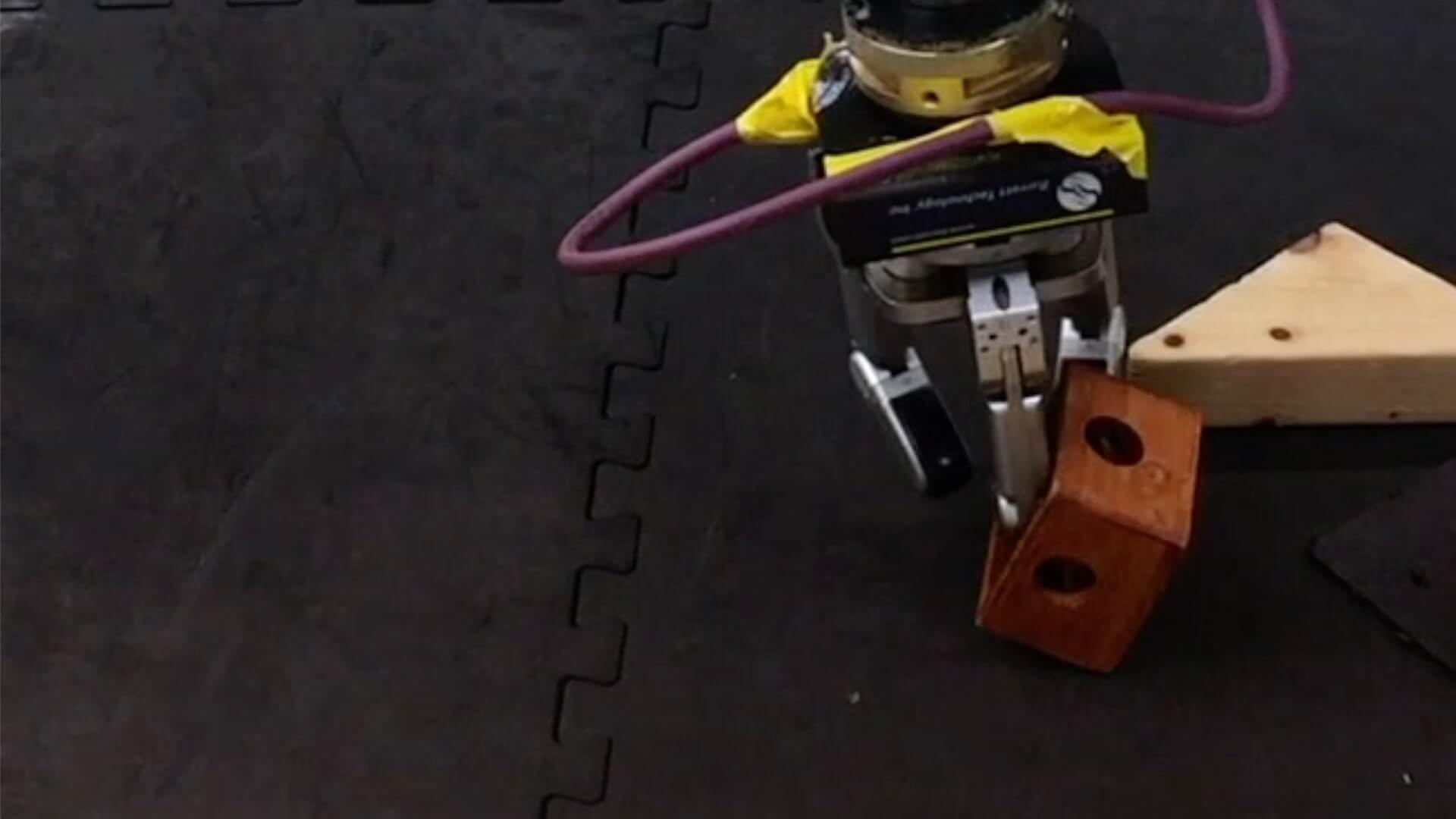}\caption{}
    \end{subfigure}
    \begin{subfigure}[t]{0.16\textwidth}
        \captionsetup{skip=0pt}\includegraphics[trim={675 0 50 0},clip,width=\linewidth]{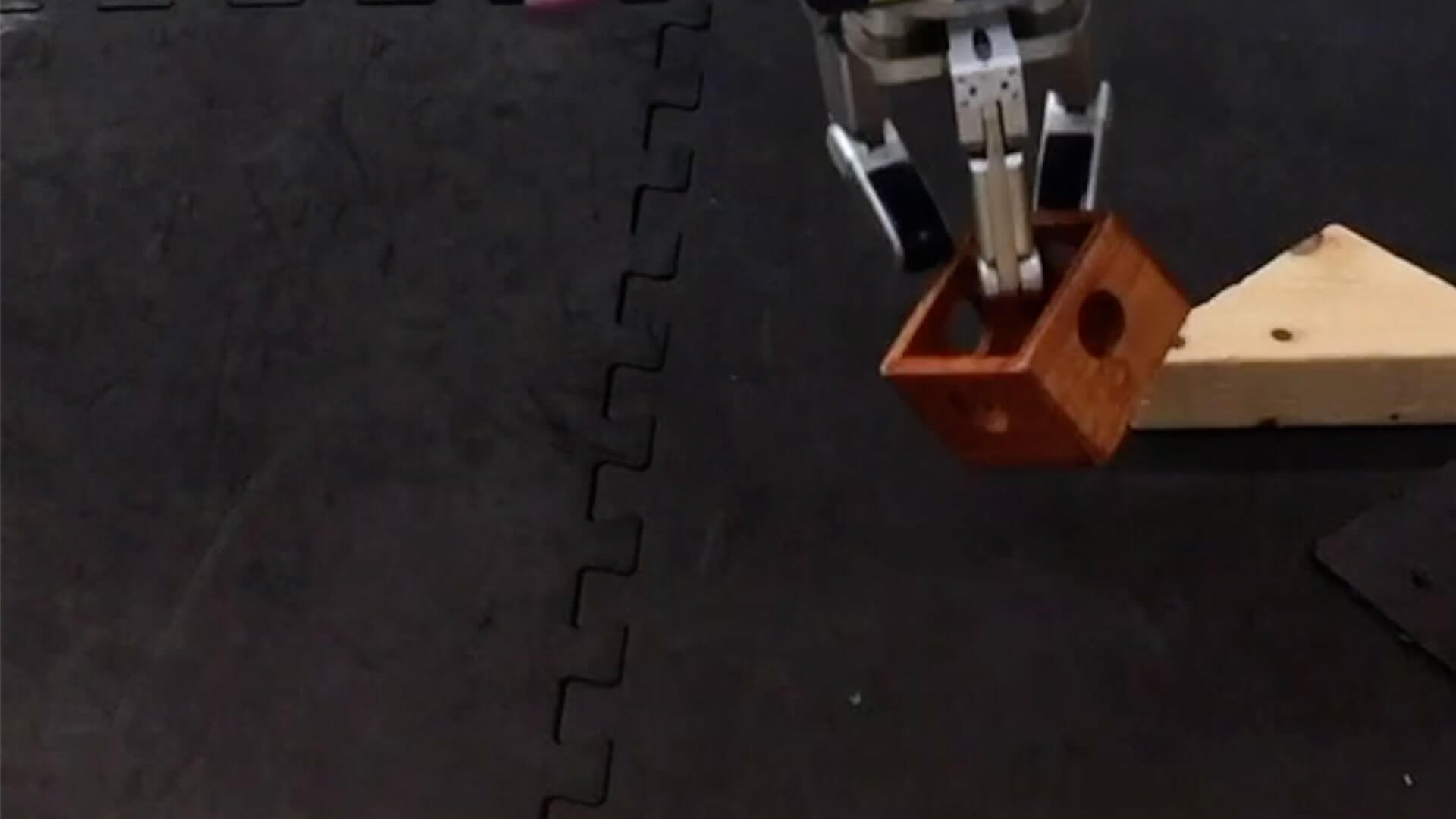}\caption{}
    \end{subfigure}
    \setcounter{subfigure}{0}
    \centering
    \begin{subfigure}[t]{0.16\textwidth}
        \captionsetup{skip=0pt}\includegraphics[trim={300 0 300 0},clip,width=\linewidth]{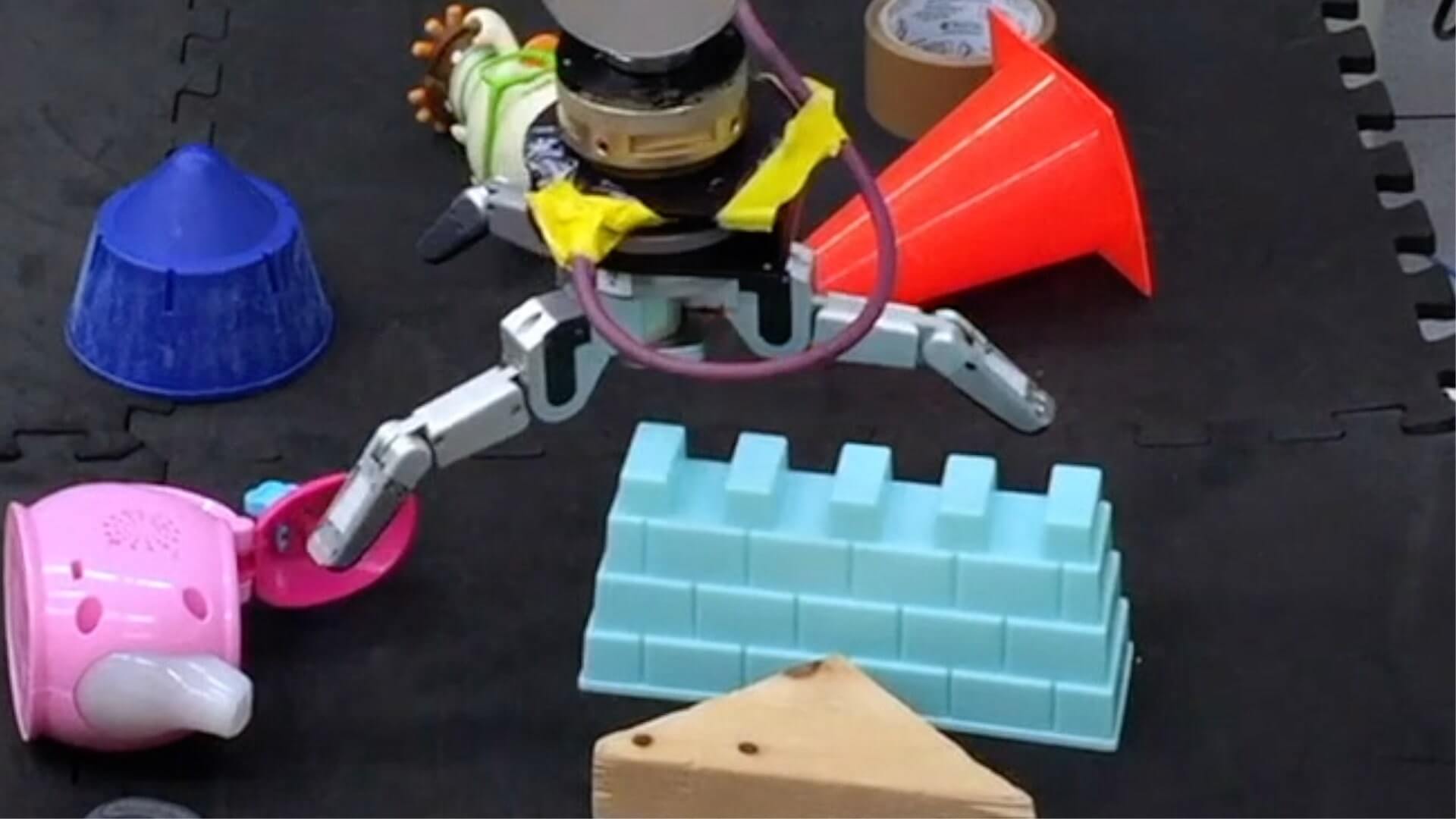}\caption{}
    \end{subfigure}
    \begin{subfigure}[t]{0.16\textwidth}
        \captionsetup{skip=0pt}\includegraphics[trim={300 0 300 0},clip,width=\linewidth]{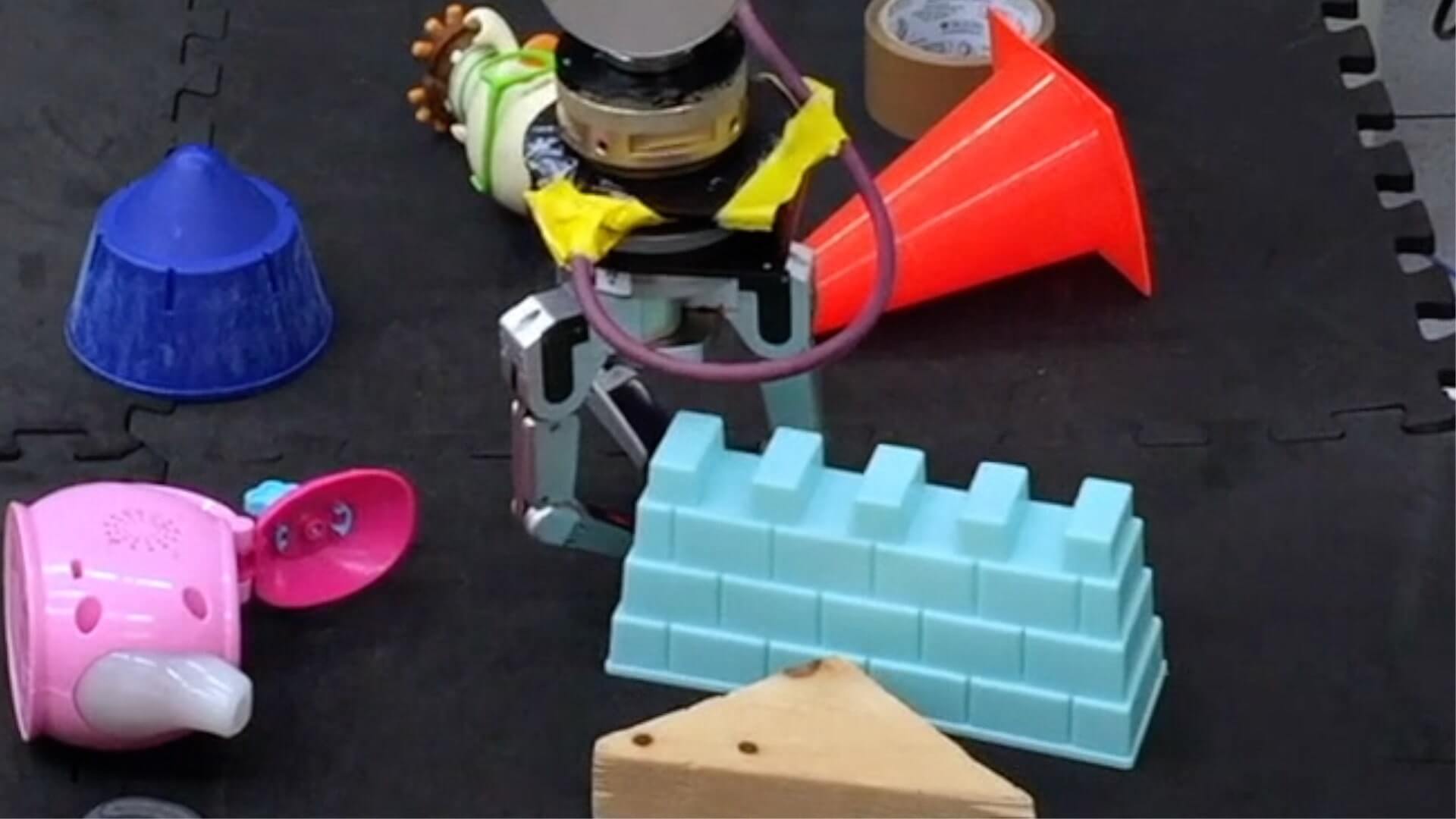}\caption{}
    \end{subfigure}
    \begin{subfigure}[t]{0.16\textwidth}
        \captionsetup{skip=0pt}\includegraphics[trim={300 0 300 0},clip,width=\linewidth]{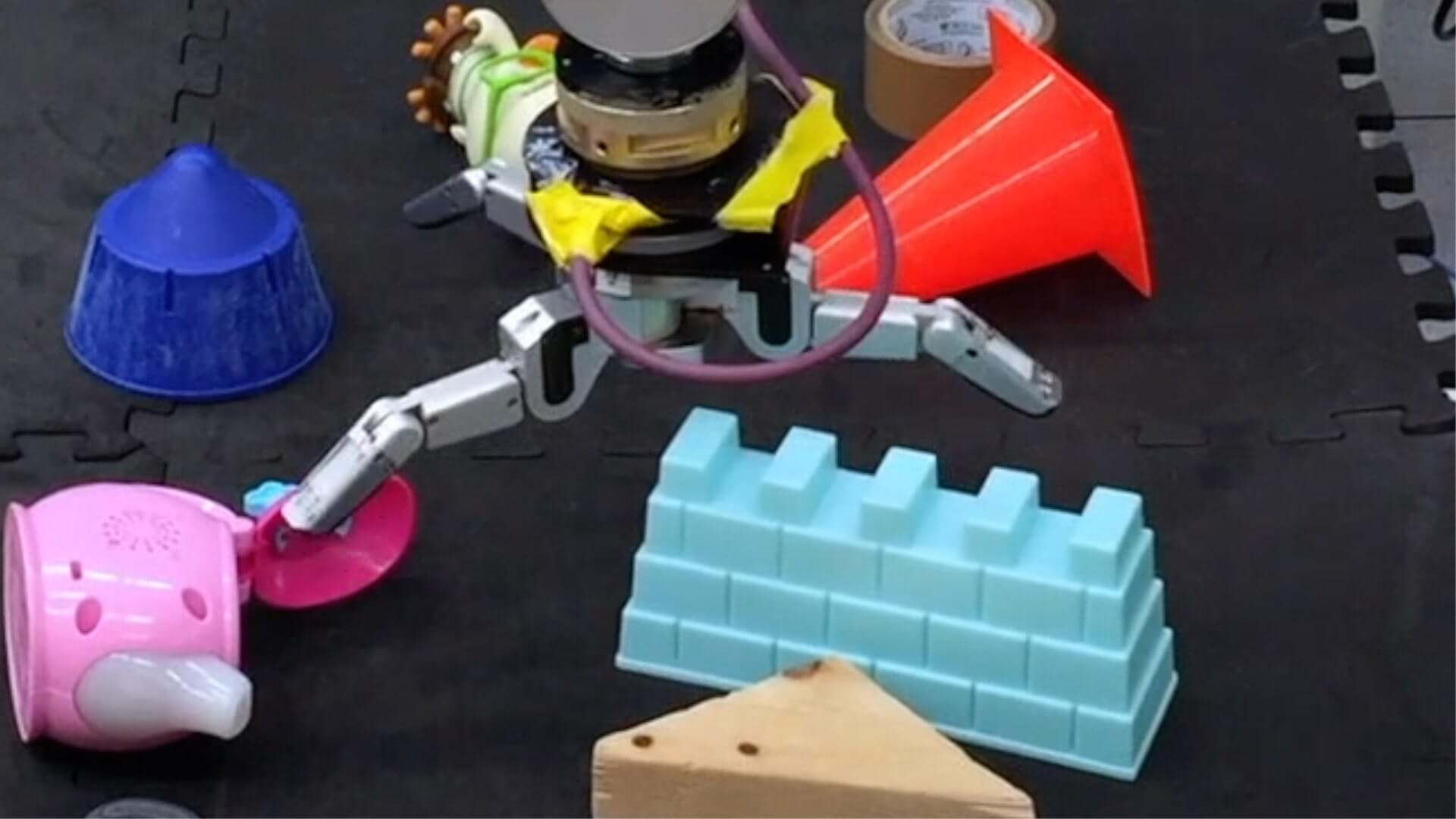}\caption{}
    \end{subfigure}
    \begin{subfigure}[t]{0.16\textwidth}
        \captionsetup{skip=0pt}\includegraphics[trim={300 0 300 0},clip,width=\linewidth]{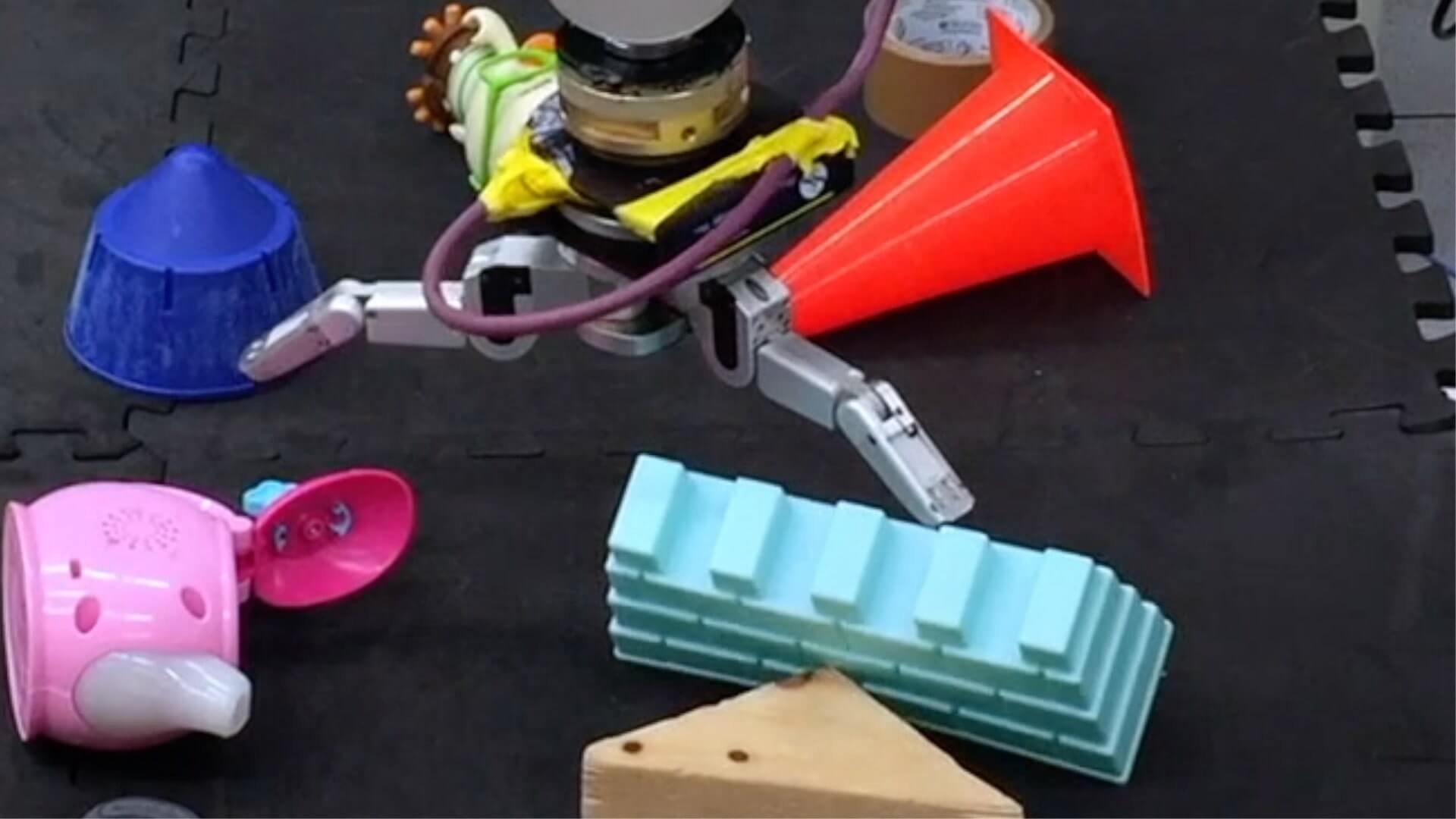}\caption{}
    \end{subfigure}
    \begin{subfigure}[t]{0.16\textwidth}
        \captionsetup{skip=0pt}\includegraphics[trim={300 0 300 0},clip,width=\linewidth]{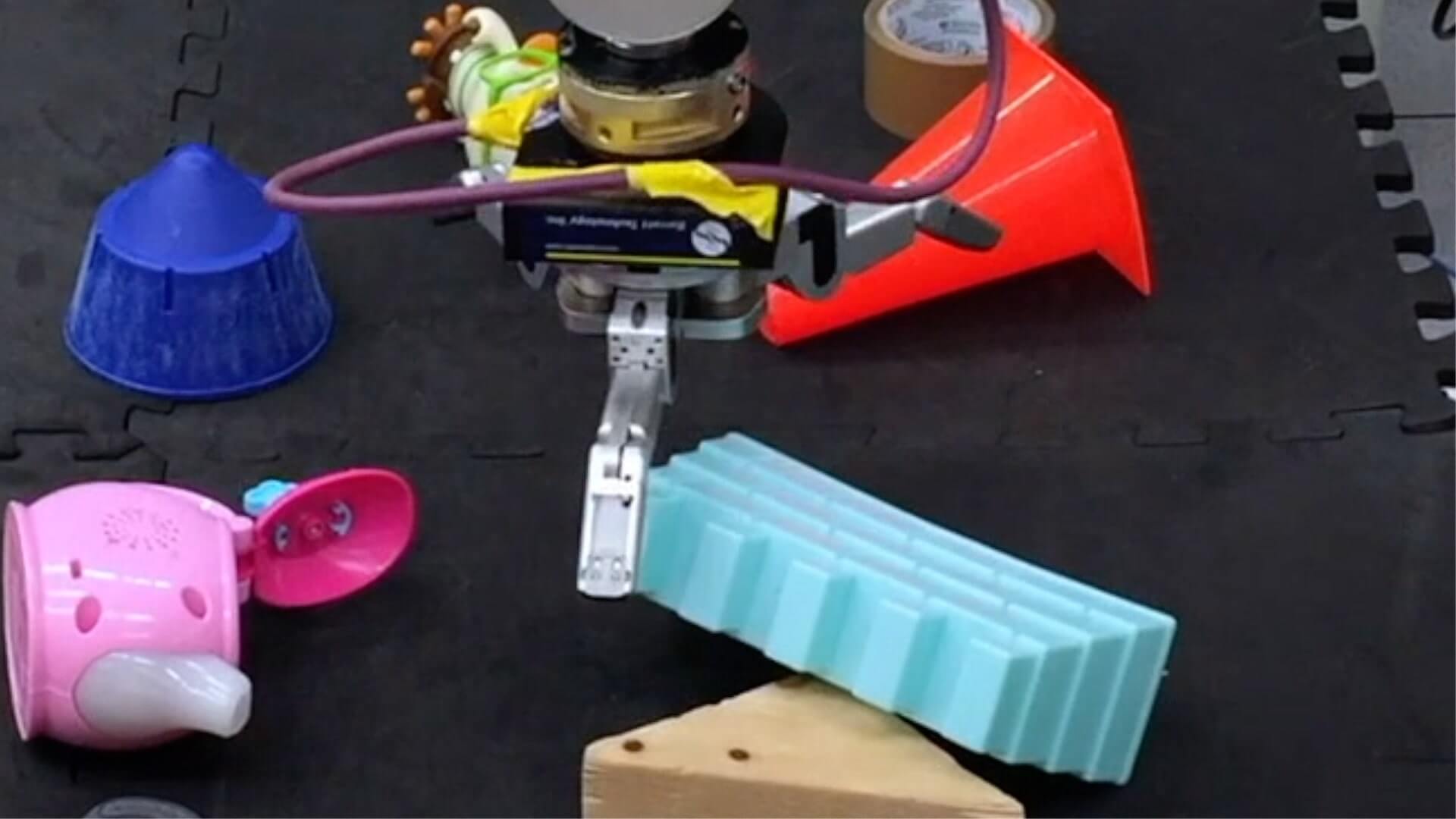}\caption{}
    \end{subfigure}
    \begin{subfigure}[t]{0.16\textwidth}
        \captionsetup{skip=0pt}\includegraphics[trim={300 0 300 0},clip,width=\linewidth]{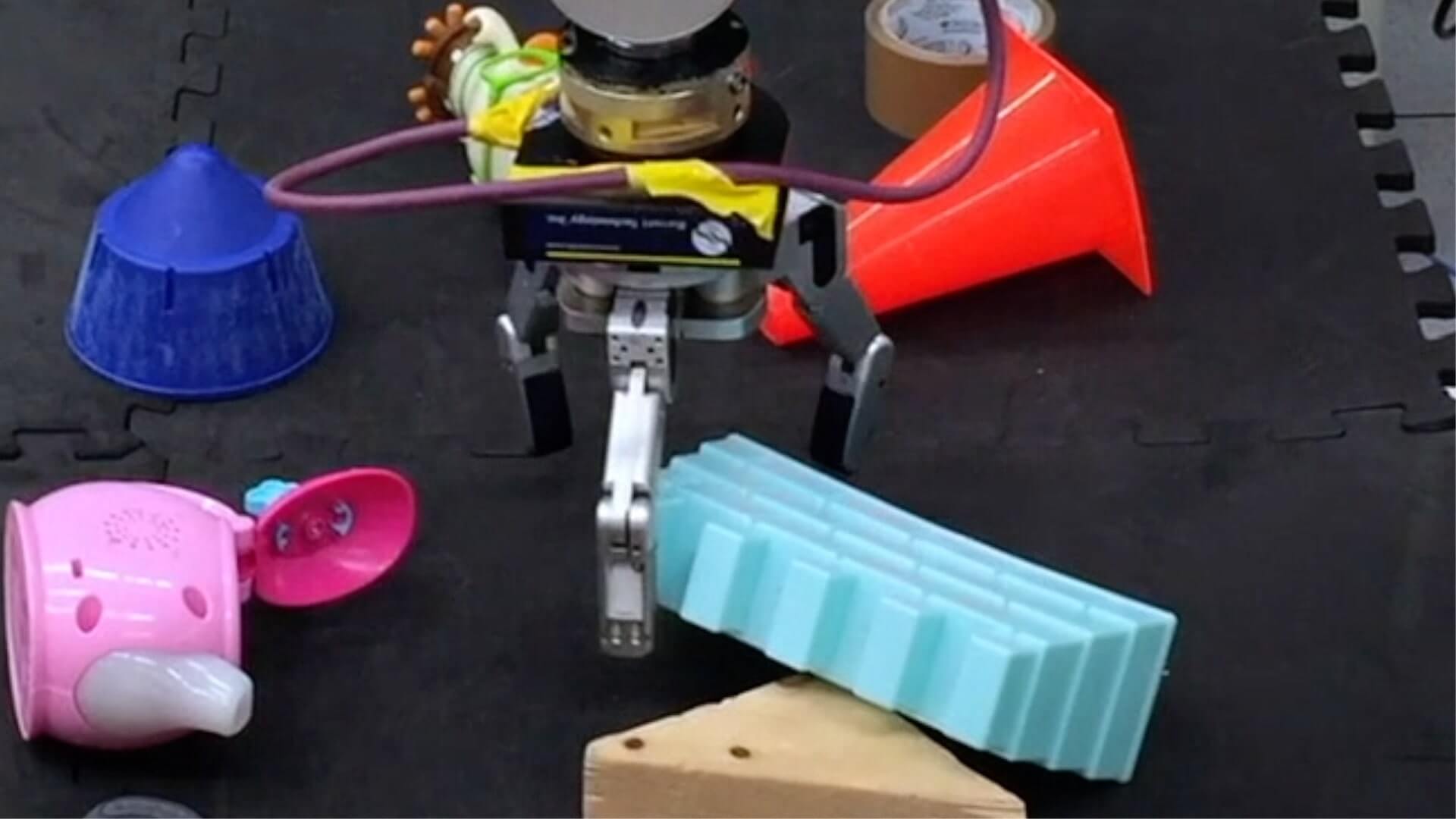}\caption{}
    \end{subfigure}
    \begin{subfigure}[t]{0.16\textwidth}
        \captionsetup{skip=0pt}\includegraphics[trim={300 0 300 0},clip,width=\linewidth]{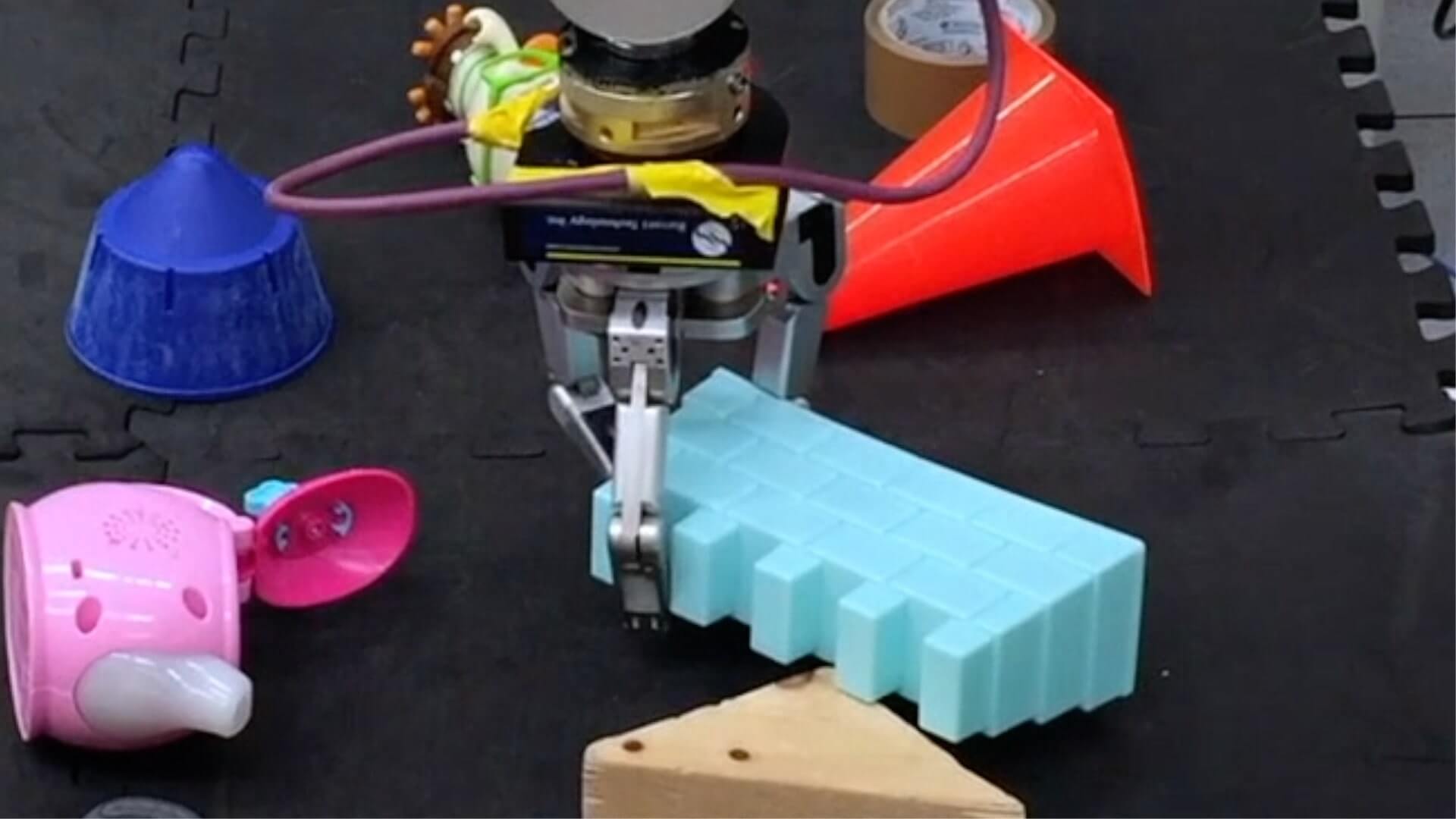}\caption{}
    \end{subfigure}
    \begin{subfigure}[t]{0.16\textwidth}
        \captionsetup{skip=0pt}\includegraphics[trim={300 0 300 0},clip,width=\linewidth]{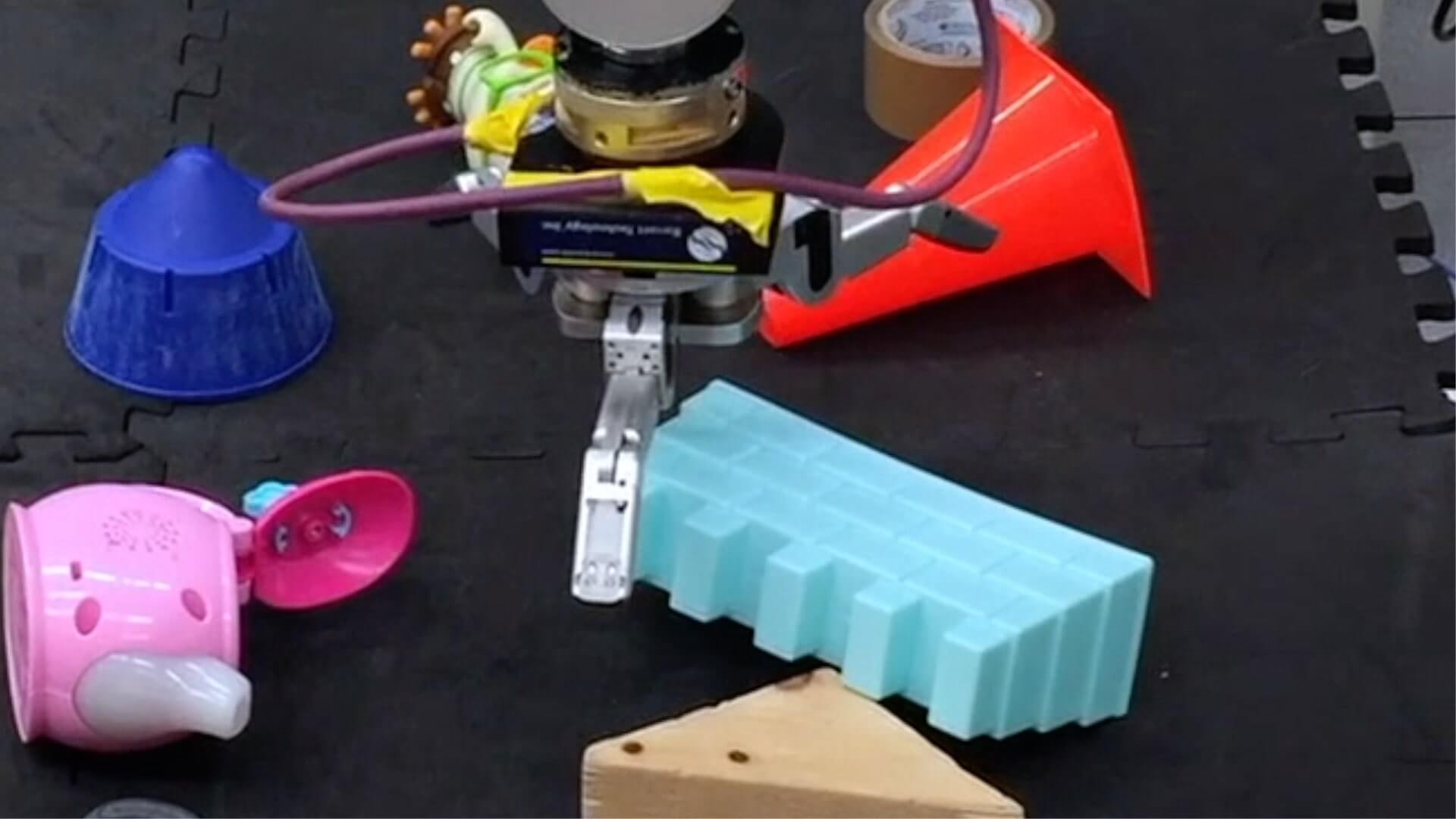}\caption{}
    \end{subfigure}
    \begin{subfigure}[t]{0.16\textwidth}
        \captionsetup{skip=0pt}\includegraphics[trim={300 0 300 0},clip,width=\linewidth]{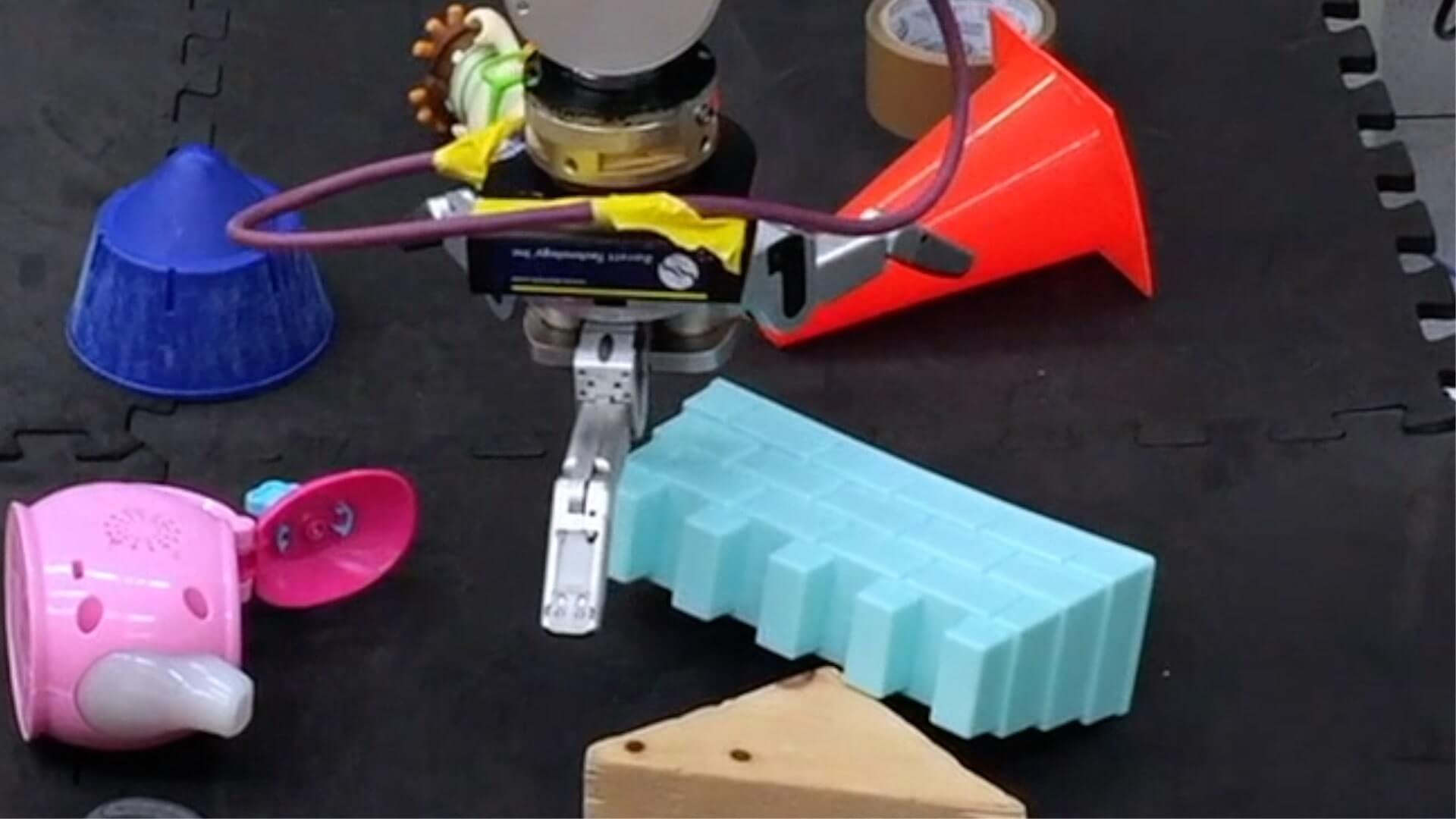}\caption{}
    \end{subfigure}
    \begin{subfigure}[t]{0.16\textwidth}
        \captionsetup{skip=0pt}\includegraphics[trim={300 0 300 0},clip,width=\linewidth]{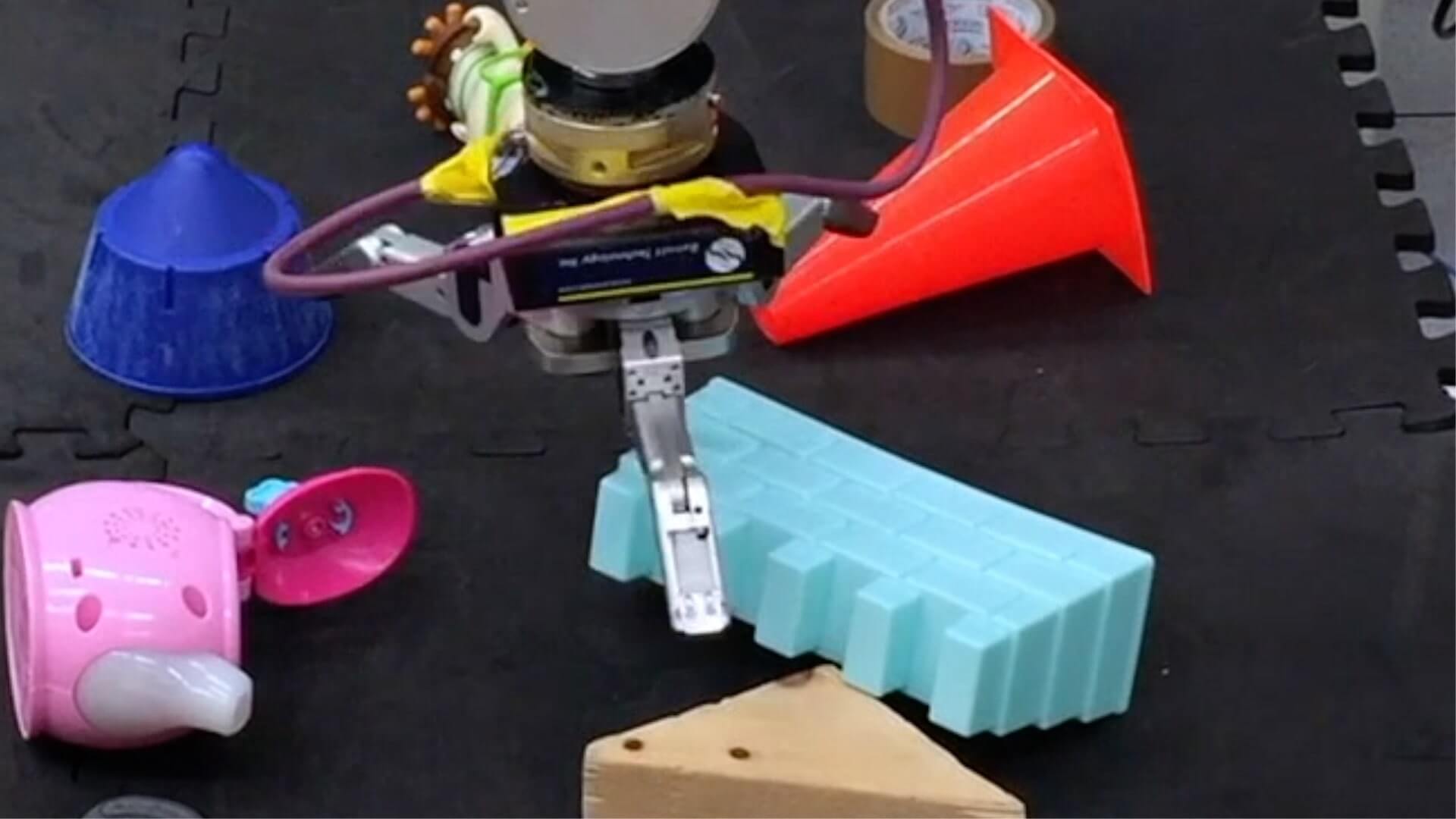}\caption{}
    \end{subfigure}
    \begin{subfigure}[t]{0.16\textwidth}
        \captionsetup{skip=0pt}\includegraphics[trim={300 0 300 0},clip,width=\linewidth]{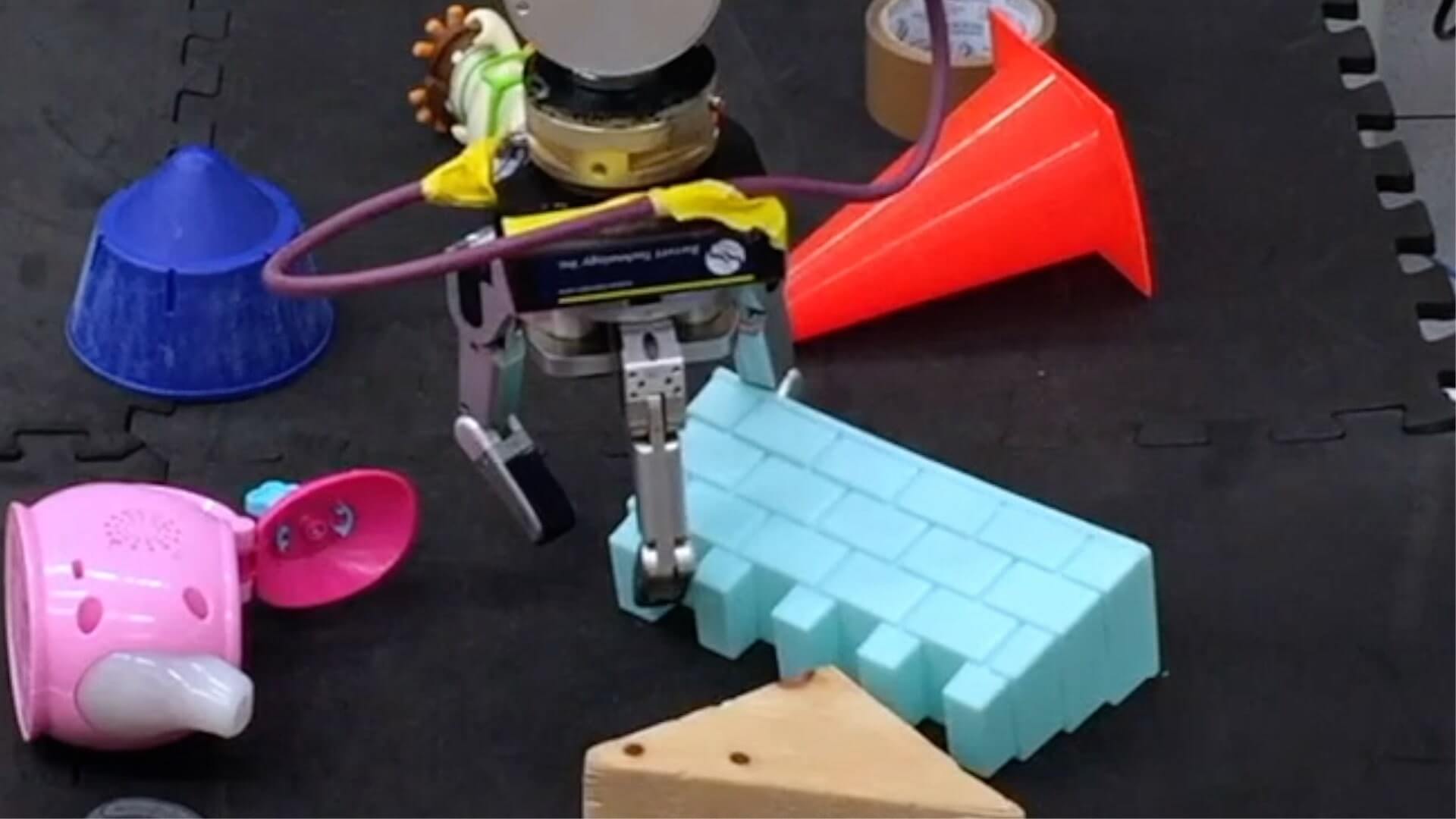}\caption{}
    \end{subfigure}
    \begin{subfigure}[t]{0.16\textwidth}
        \captionsetup{skip=0pt}\includegraphics[trim={300 0 300 0},clip,width=\linewidth]{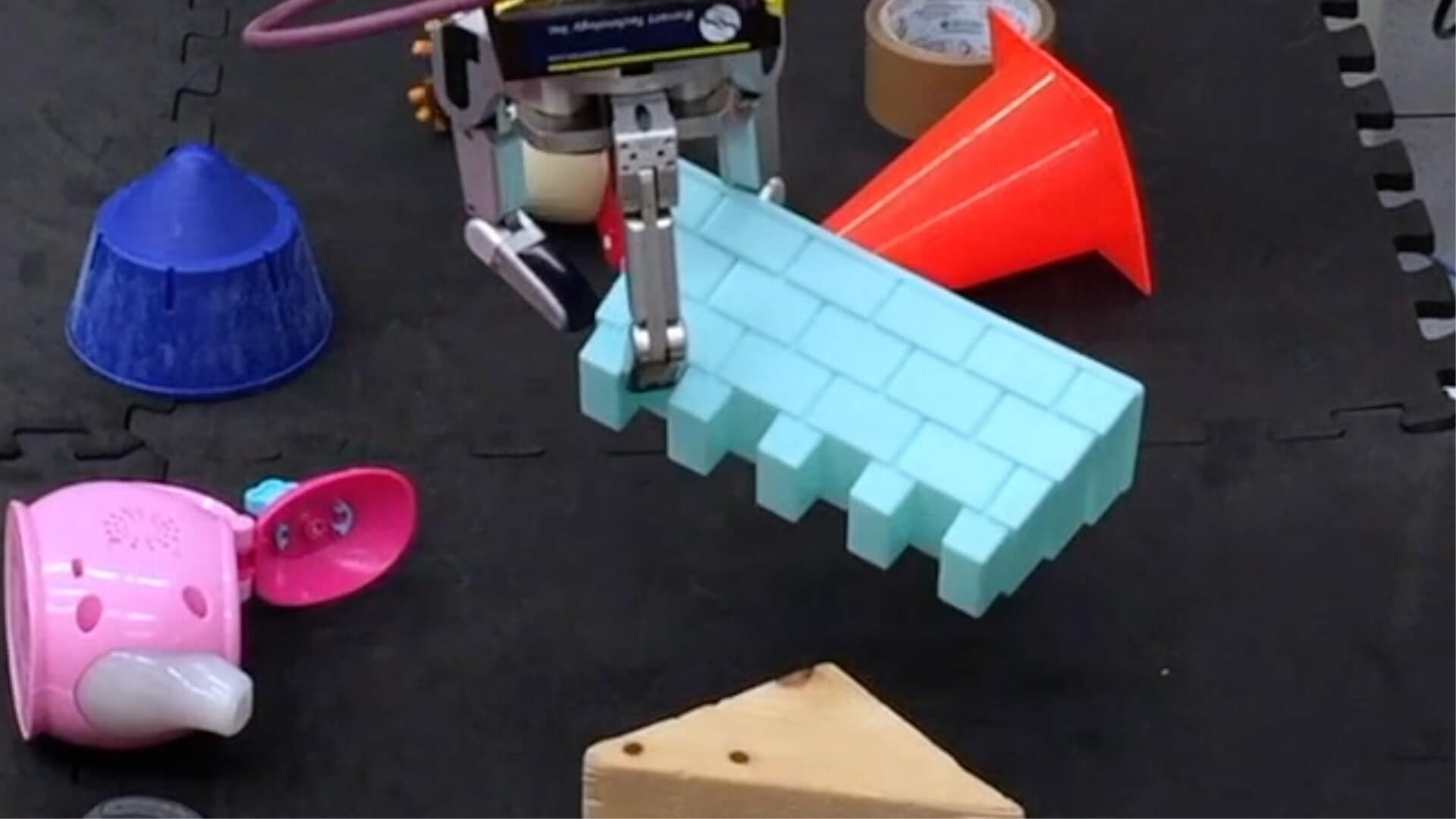}\caption{}
    \end{subfigure}
\setcounter{subfigure}{0}
    \begin{subfigure}[t]{0.16\textwidth}
        \captionsetup{skip=0pt}\includegraphics[trim={500 0 150 0},clip,width=\linewidth]{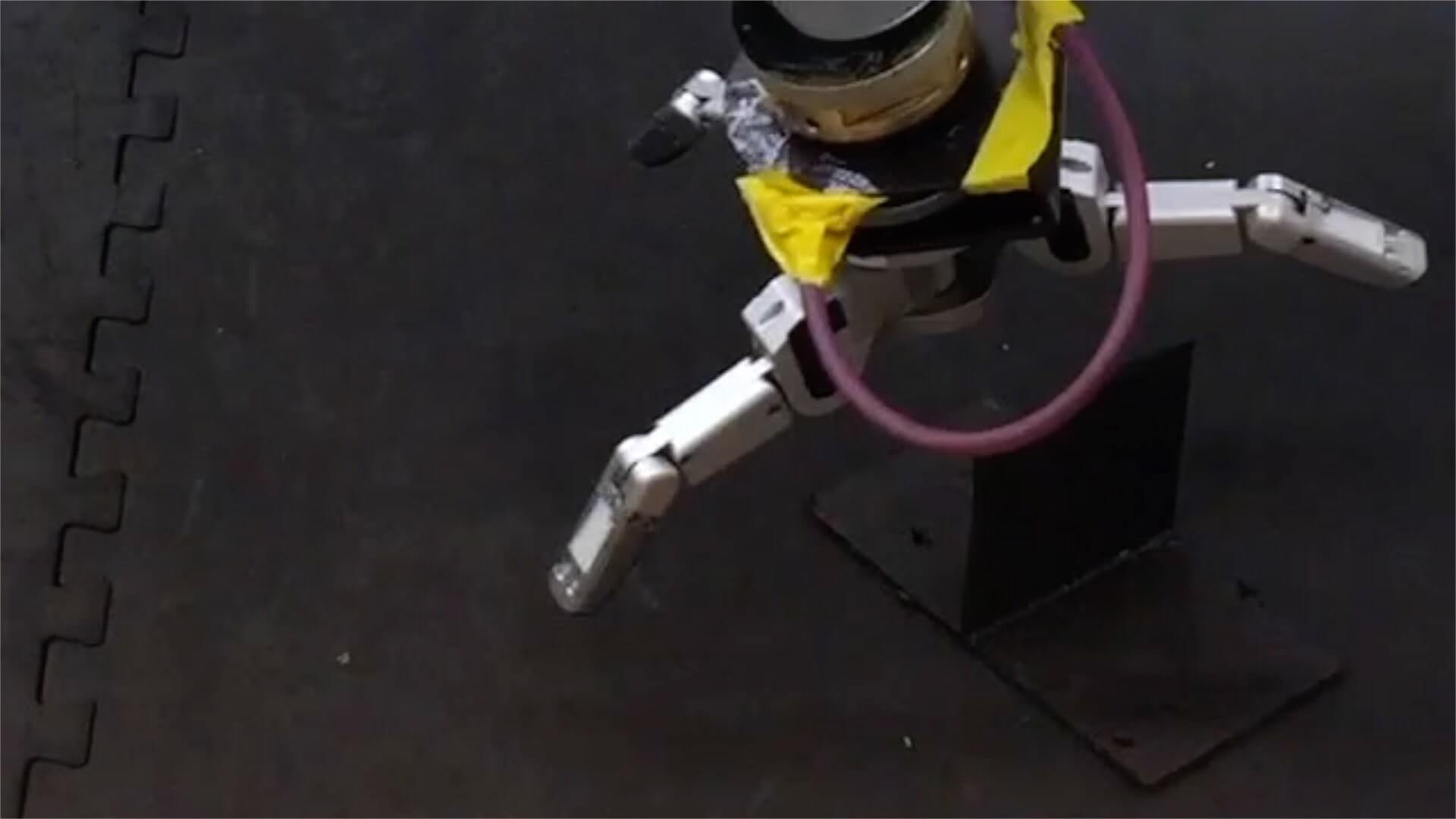}\caption{}
    \end{subfigure}
    \begin{subfigure}[t]{0.16\textwidth}
        \captionsetup{skip=0pt}\includegraphics[trim={500 0 150 0},clip,width=\linewidth]{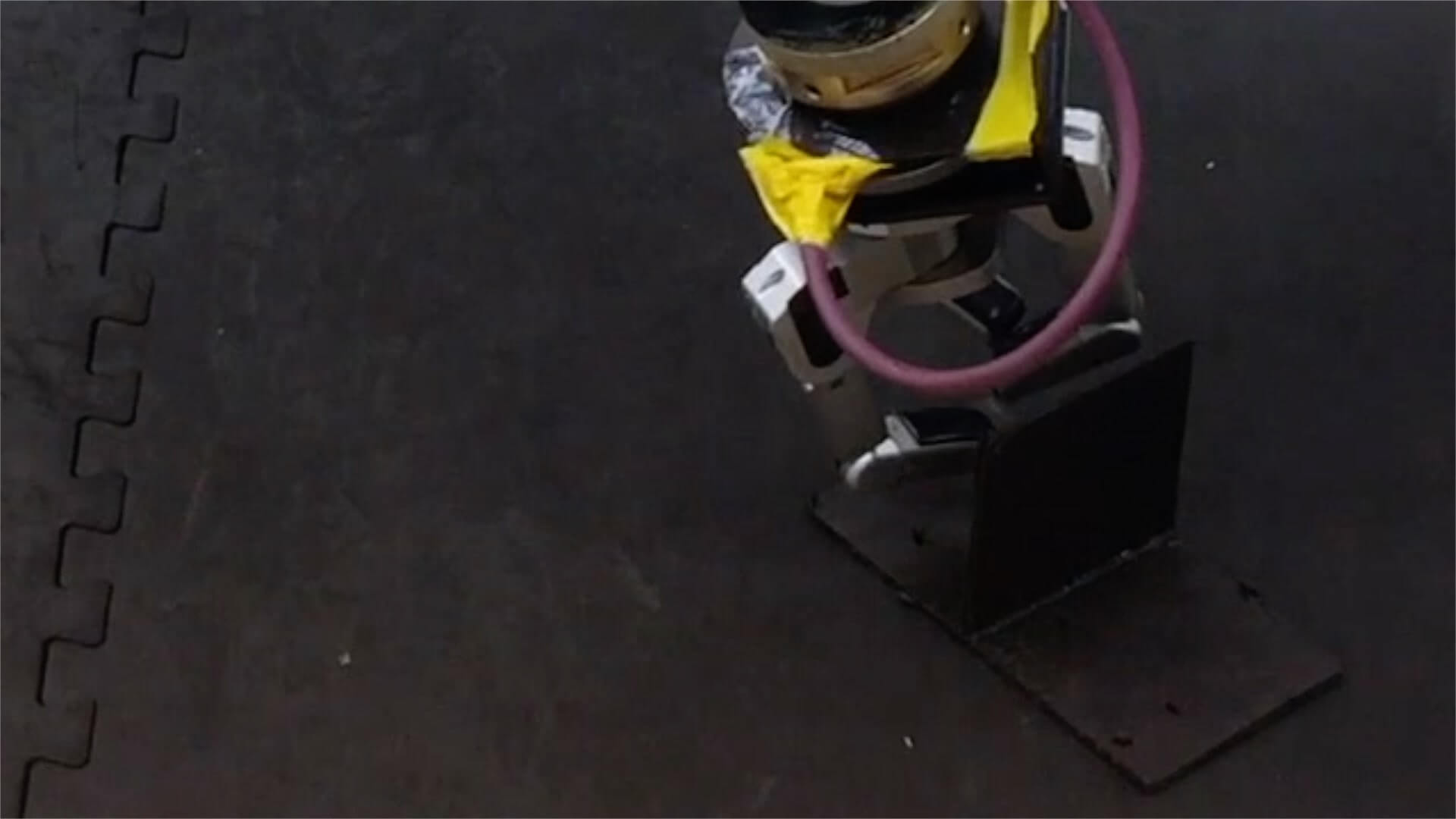}\caption{}
    \end{subfigure}
    \begin{subfigure}[t]{0.16\textwidth}
        \captionsetup{skip=0pt}\includegraphics[trim={500 0 150 0},clip,width=\linewidth]{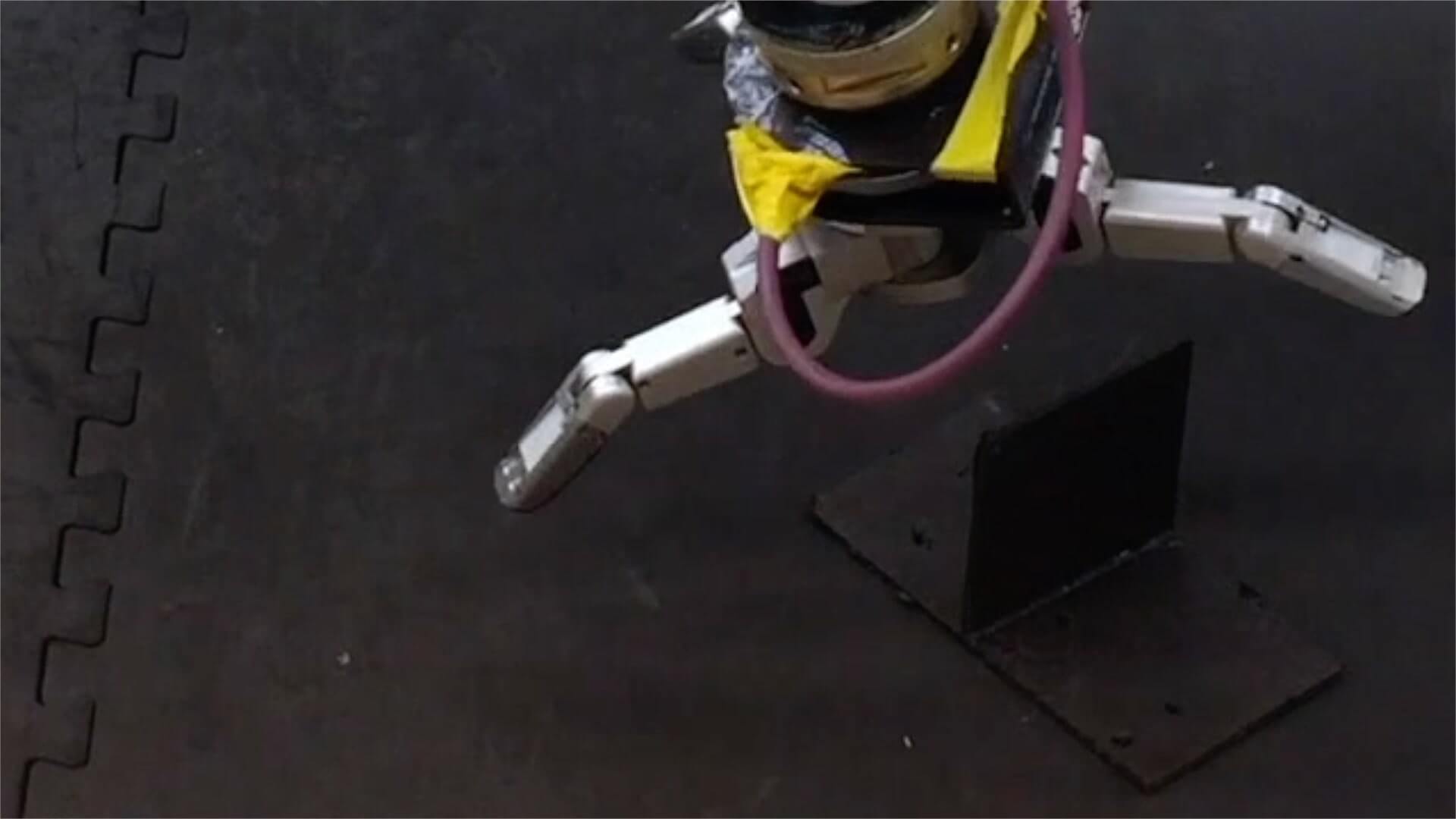}\caption{}
    \end{subfigure}
    \begin{subfigure}[t]{0.16\textwidth}
        \captionsetup{skip=0pt}\includegraphics[trim={500 0 150 0},clip,width=\linewidth]{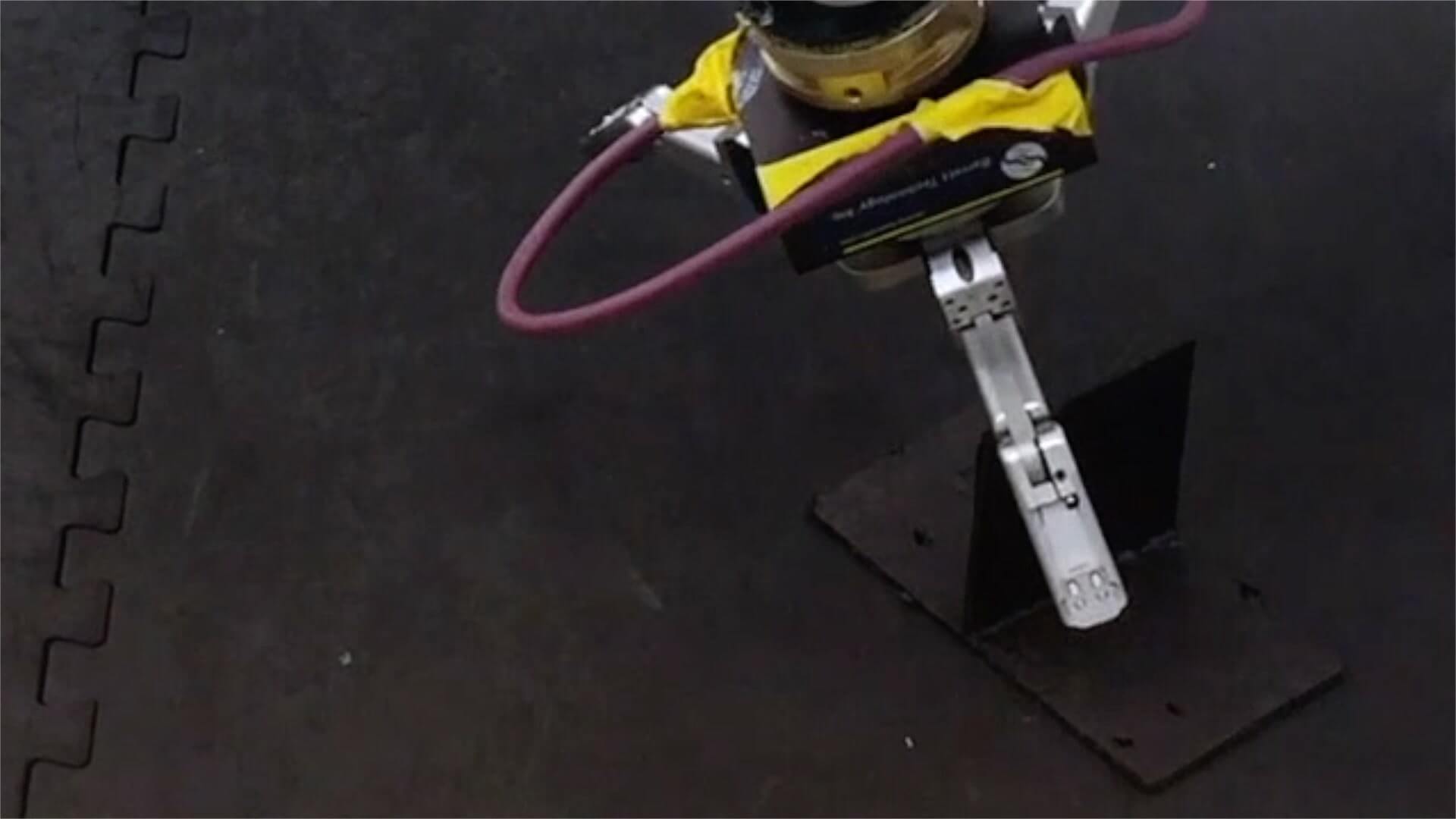}\caption{}
    \end{subfigure}
    \begin{subfigure}[t]{0.16\textwidth}
        \captionsetup{skip=0pt}\includegraphics[trim={500 0 150 0},clip,width=\linewidth]{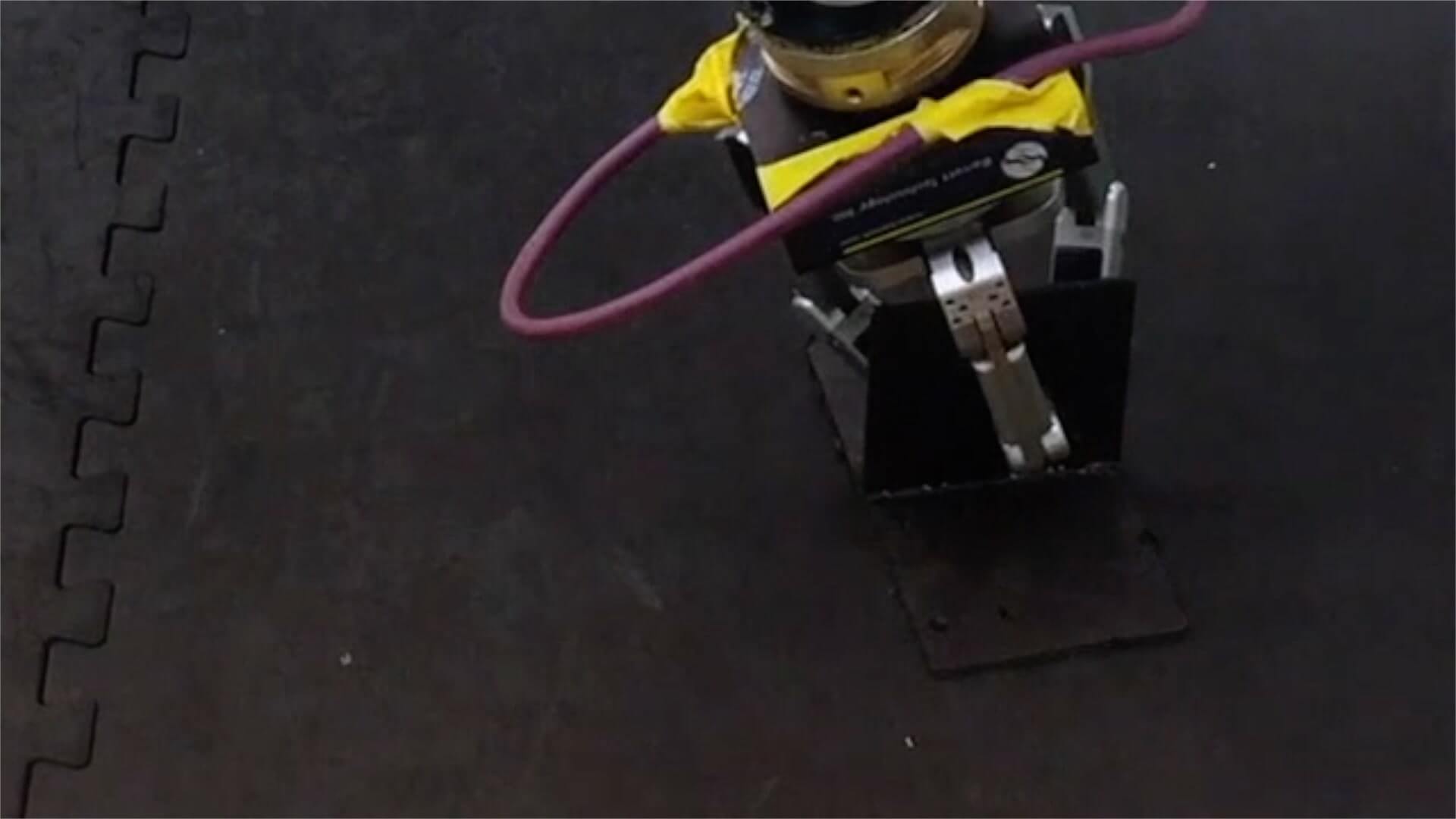}\caption{}
    \end{subfigure}
    \begin{subfigure}[t]{0.16\textwidth}
        \captionsetup{skip=0pt}\includegraphics[trim={500 0 150 0},clip,width=\linewidth]{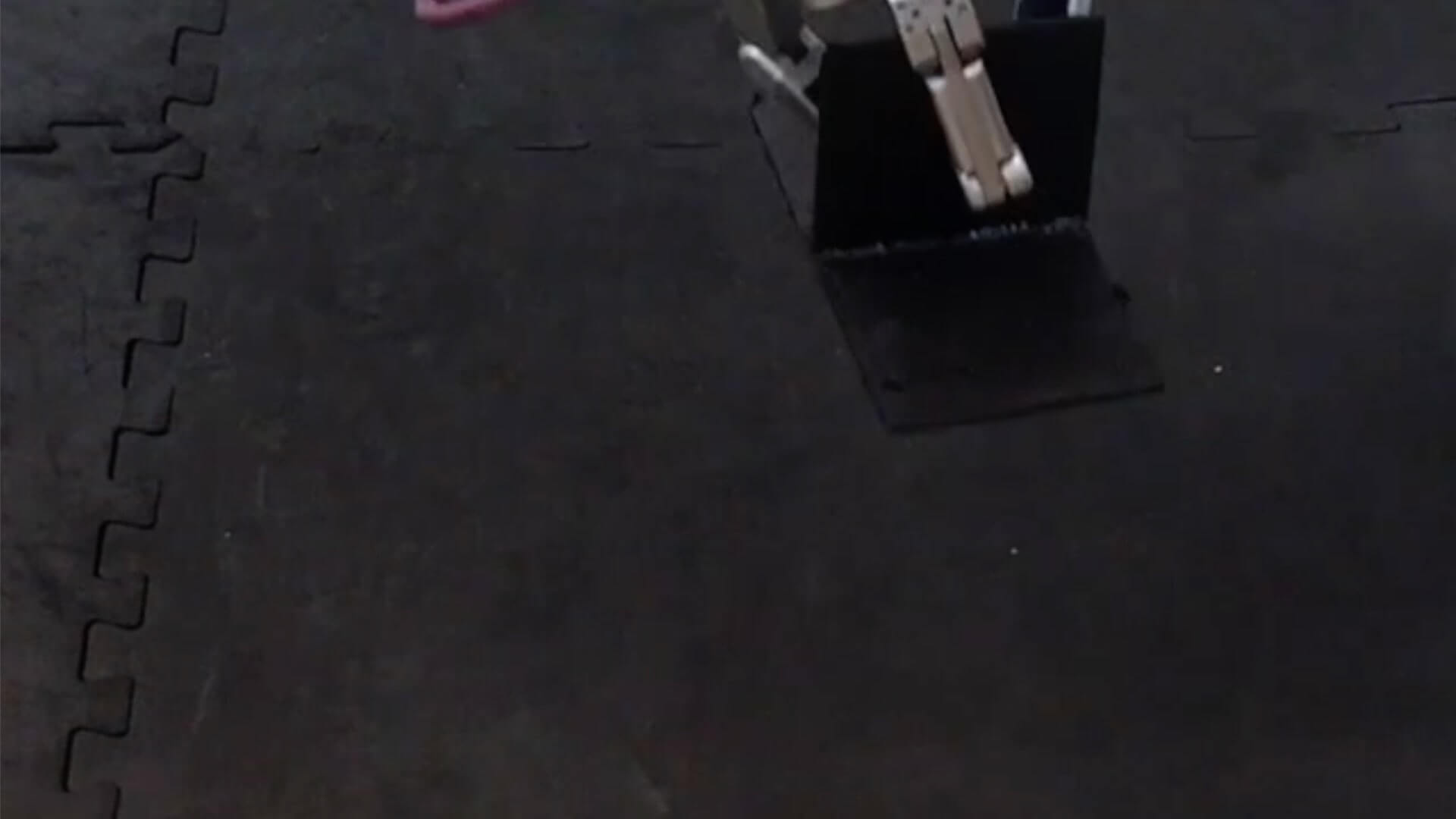}\caption{}
    \end{subfigure}
\setcounter{subfigure}{0}
    \begin{subfigure}[t]{0.16\textwidth}
        \captionsetup{skip=0pt}\includegraphics[trim={200 0 200 0},clip,width=\linewidth]{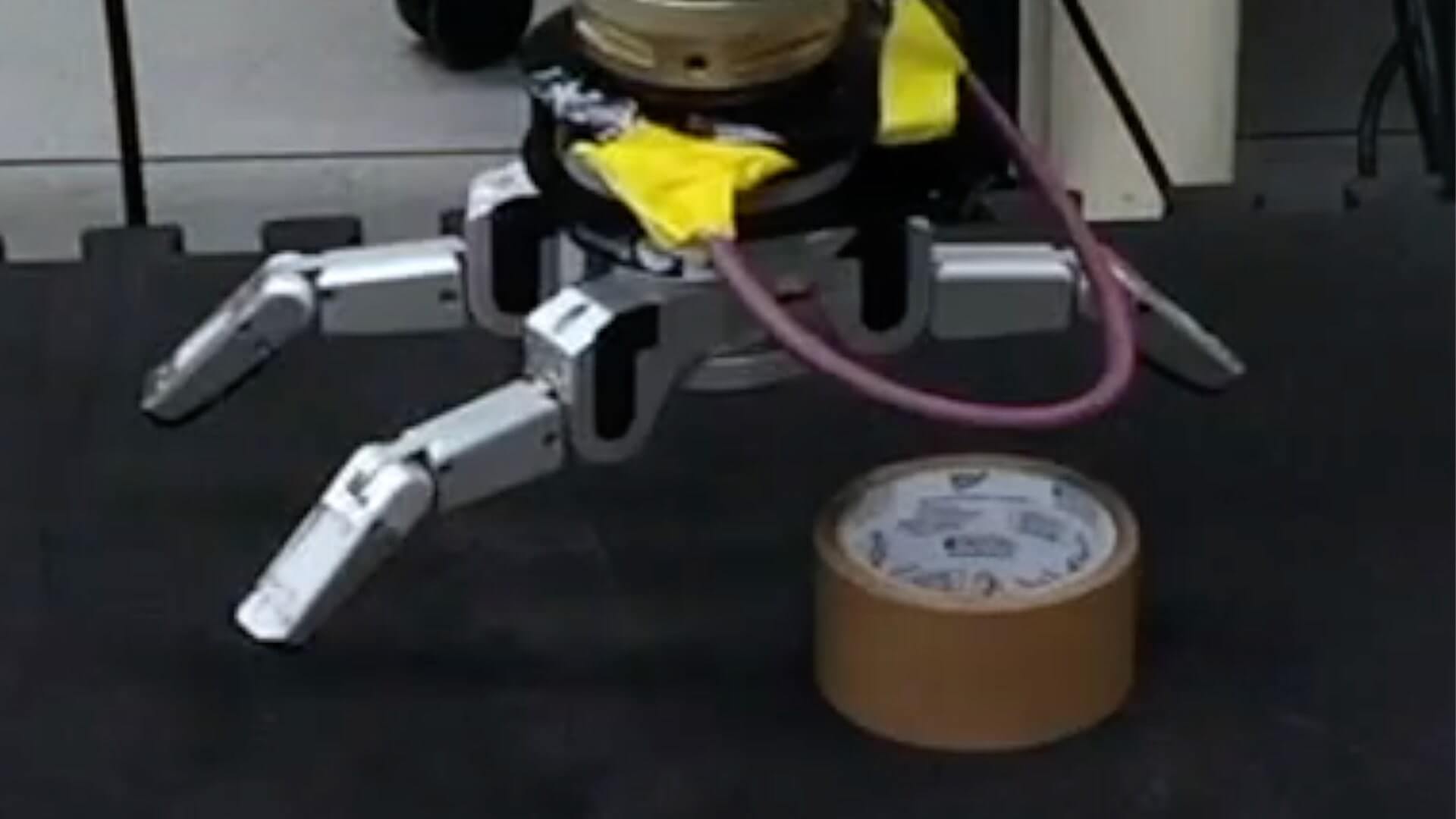}\caption{}
    \end{subfigure}
    \begin{subfigure}[t]{0.16\textwidth}
        \captionsetup{skip=0pt}\includegraphics[trim={200 0 200 0},clip,width=\linewidth]{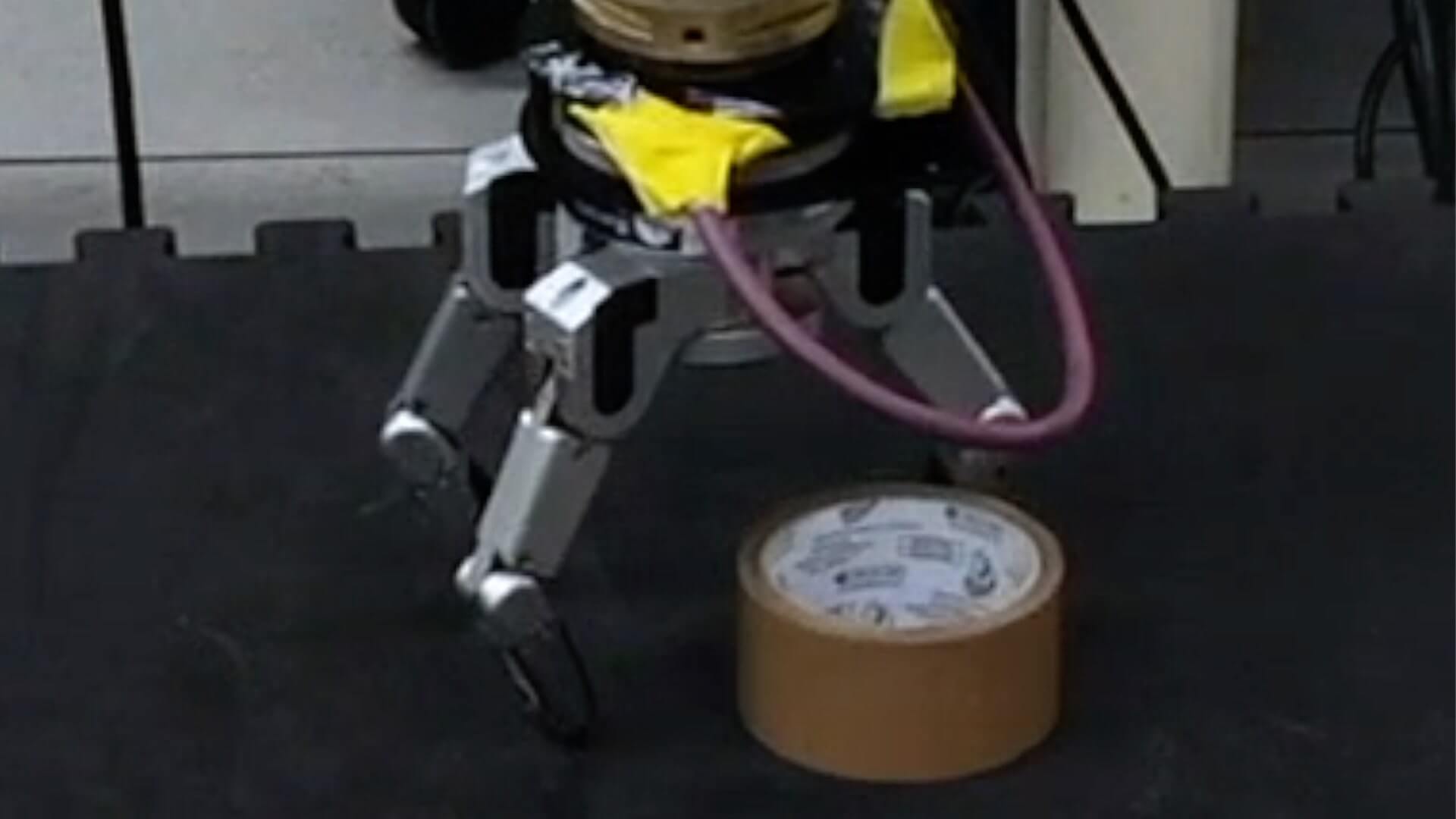}\caption{}
    \end{subfigure}
    \begin{subfigure}[t]{0.16\textwidth}
        \captionsetup{skip=0pt}\includegraphics[trim={200 0 200 0},clip,width=\linewidth]{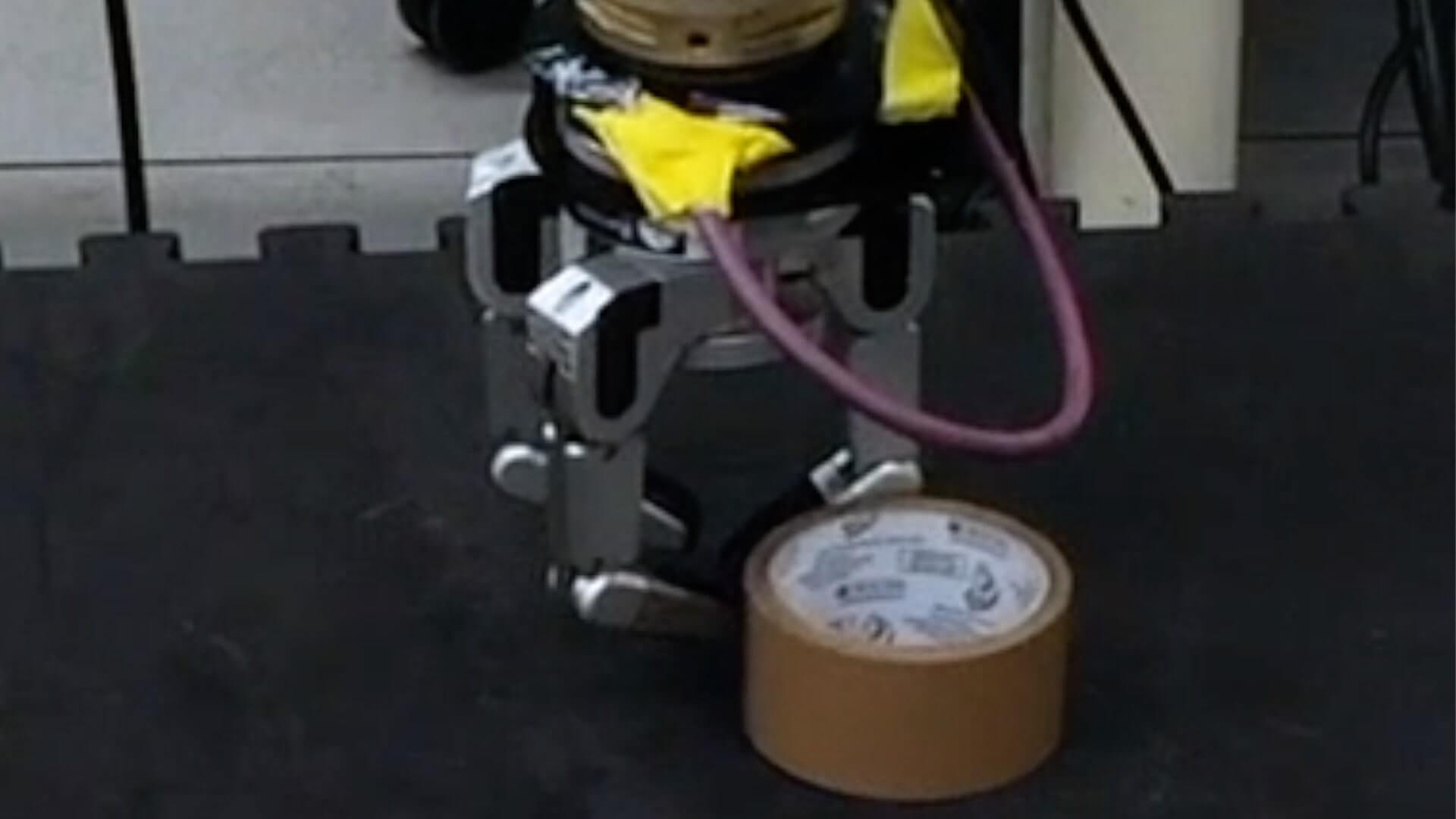}\caption{}
    \end{subfigure}
    \begin{subfigure}[t]{0.16\textwidth}
        \captionsetup{skip=0pt}\includegraphics[trim={200 0 200 0},clip,width=\linewidth]{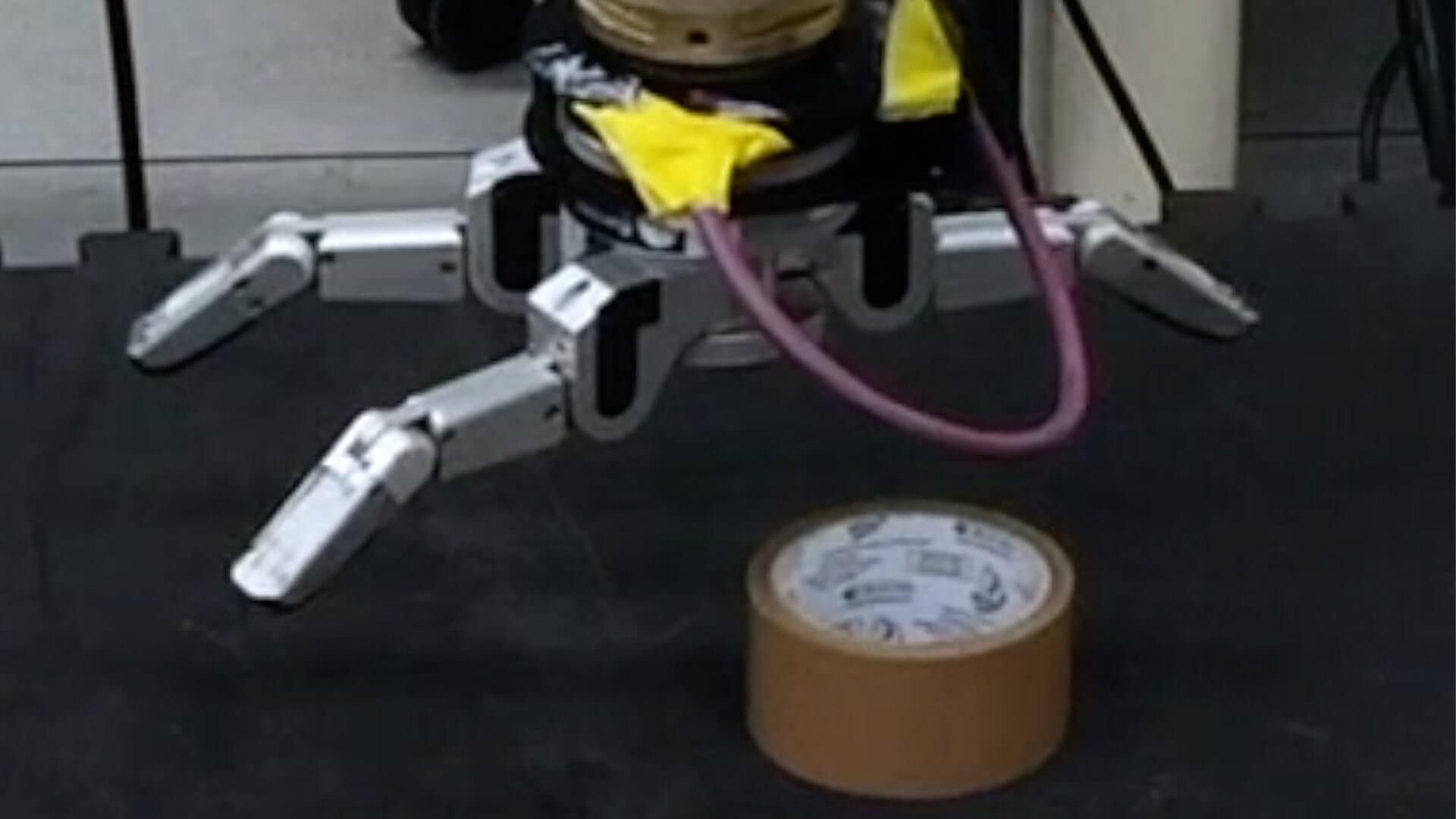}\caption{}
    \end{subfigure}
    \begin{subfigure}[t]{0.16\textwidth}
        \captionsetup{skip=0pt}\includegraphics[trim={200 0 200 0},clip,width=\linewidth]{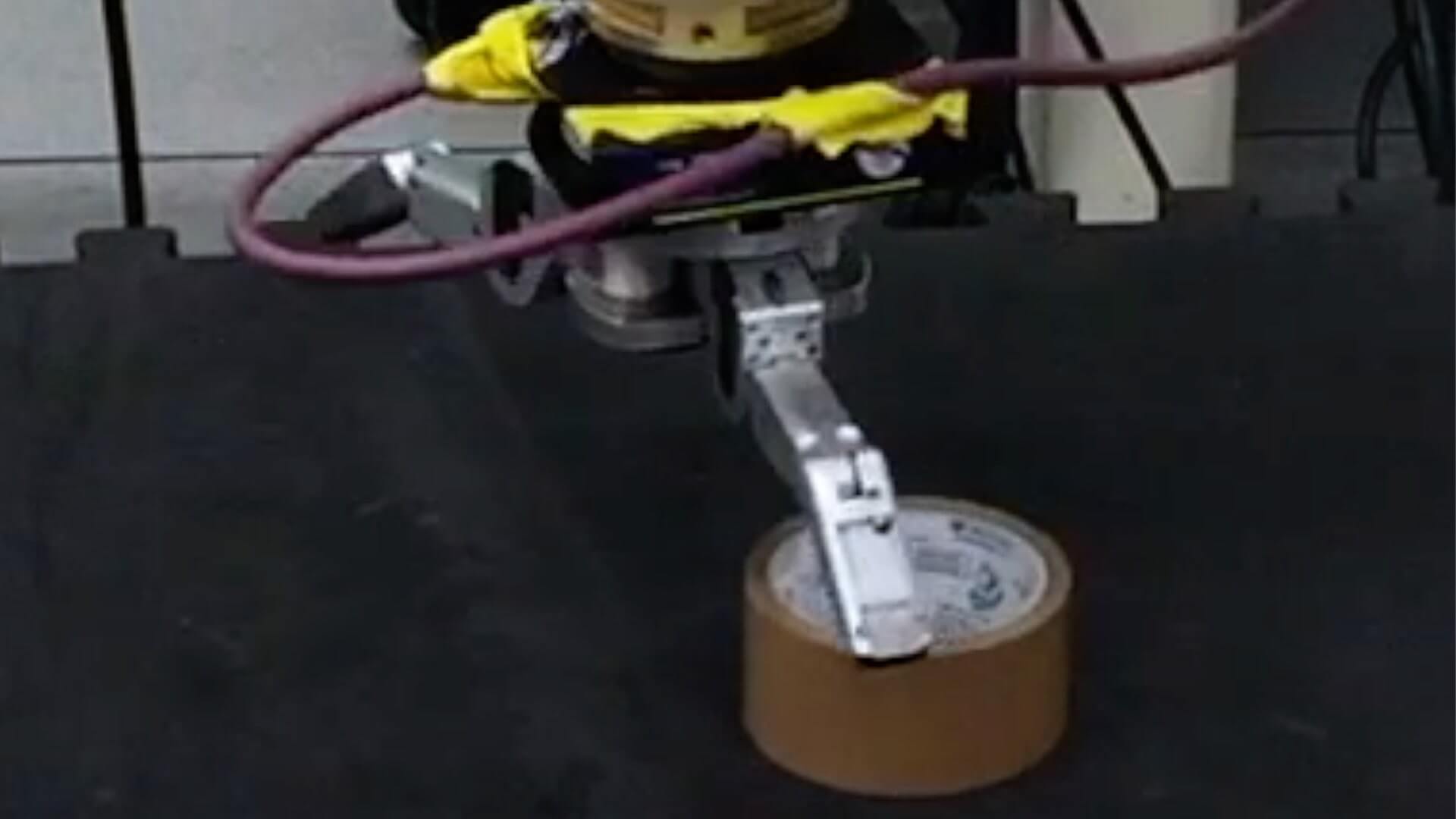}\caption{}
    \end{subfigure}
    \begin{subfigure}[t]{0.16\textwidth}
        \captionsetup{skip=0pt}\includegraphics[trim={200 0 200 0},clip,width=\linewidth]{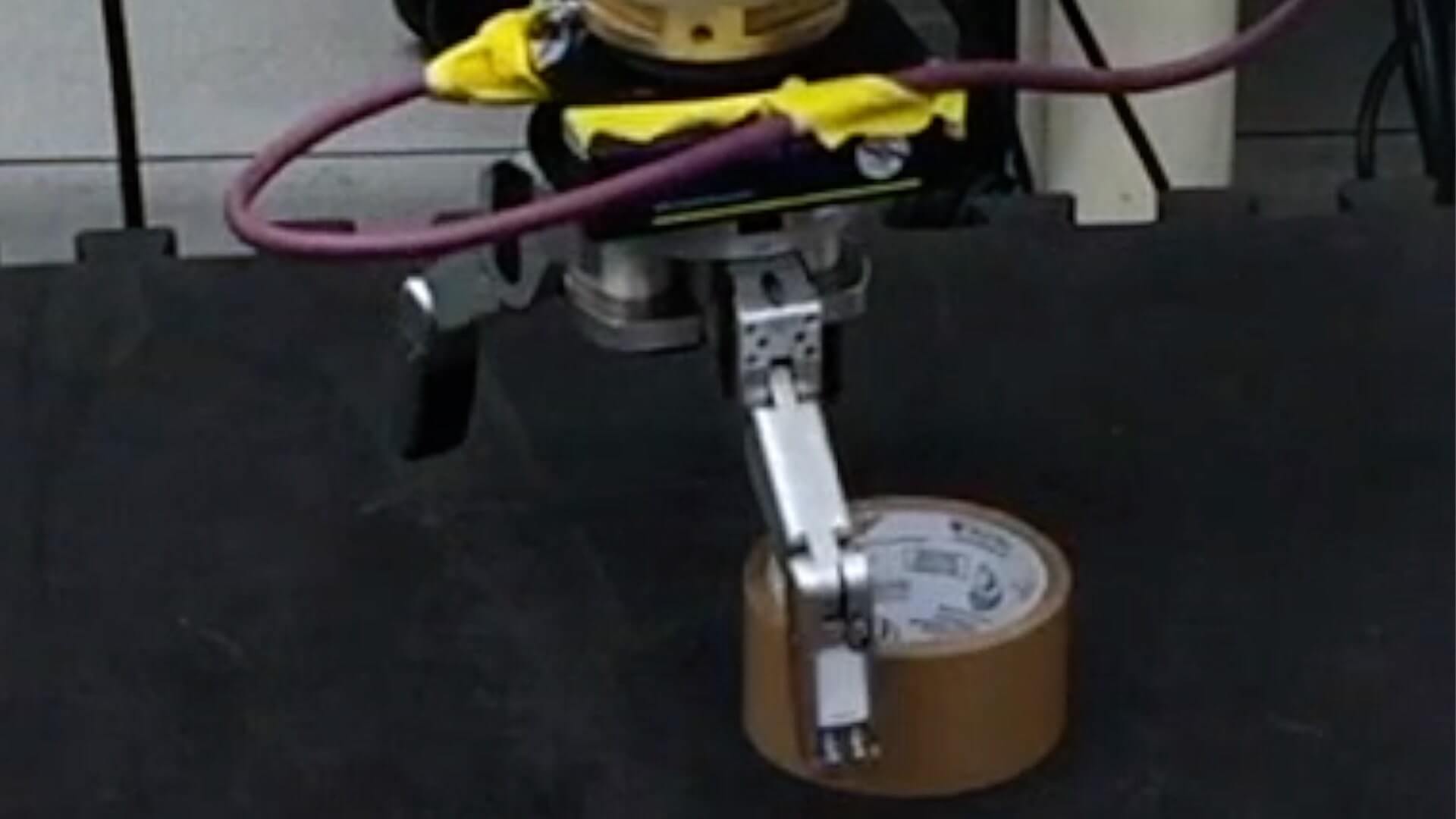}\caption{}
    \end{subfigure}
    \begin{subfigure}[t]{0.16\textwidth}
        \captionsetup{skip=0pt}\includegraphics[trim={200 0 200 0},clip,width=\linewidth]{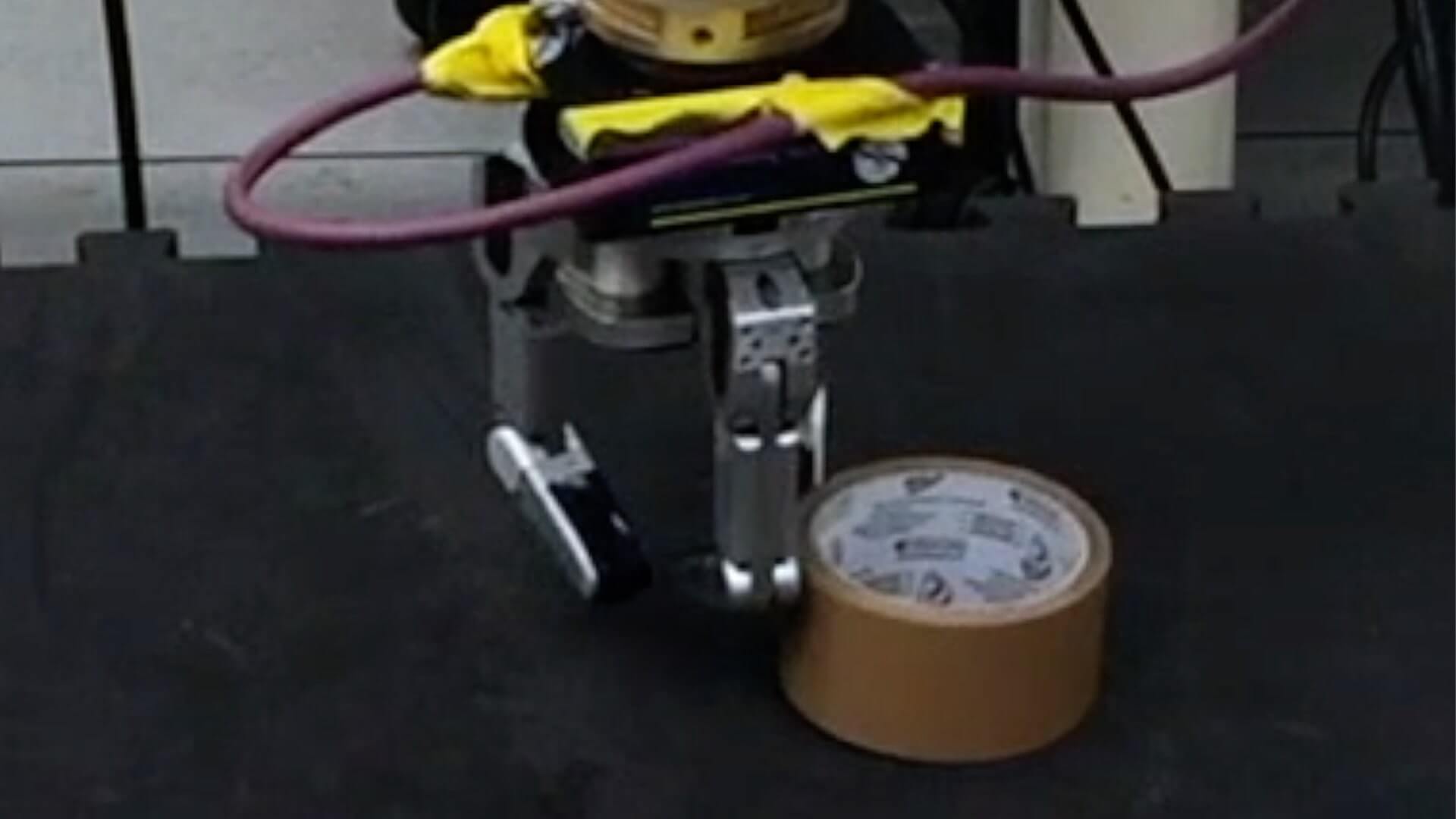}\caption{}
    \end{subfigure}
    \begin{subfigure}[t]{0.16\textwidth}
        \captionsetup{skip=0pt}\includegraphics[trim={200 0 200 0},clip,width=\linewidth]{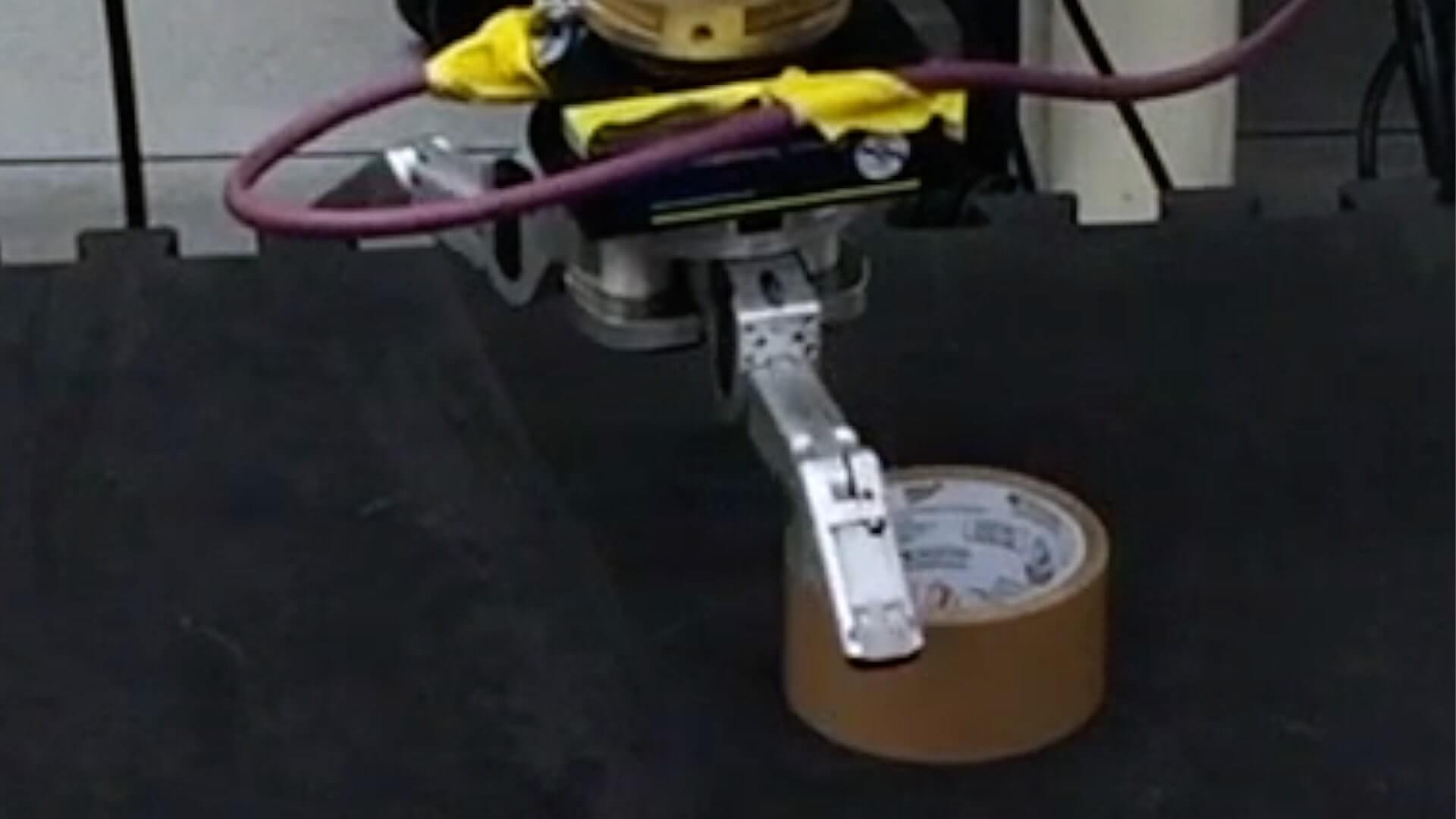}\caption{}
    \end{subfigure}
    \begin{subfigure}[t]{0.16\textwidth}
        \captionsetup{skip=0pt}\includegraphics[trim={200 0 200 0},clip,width=\linewidth]{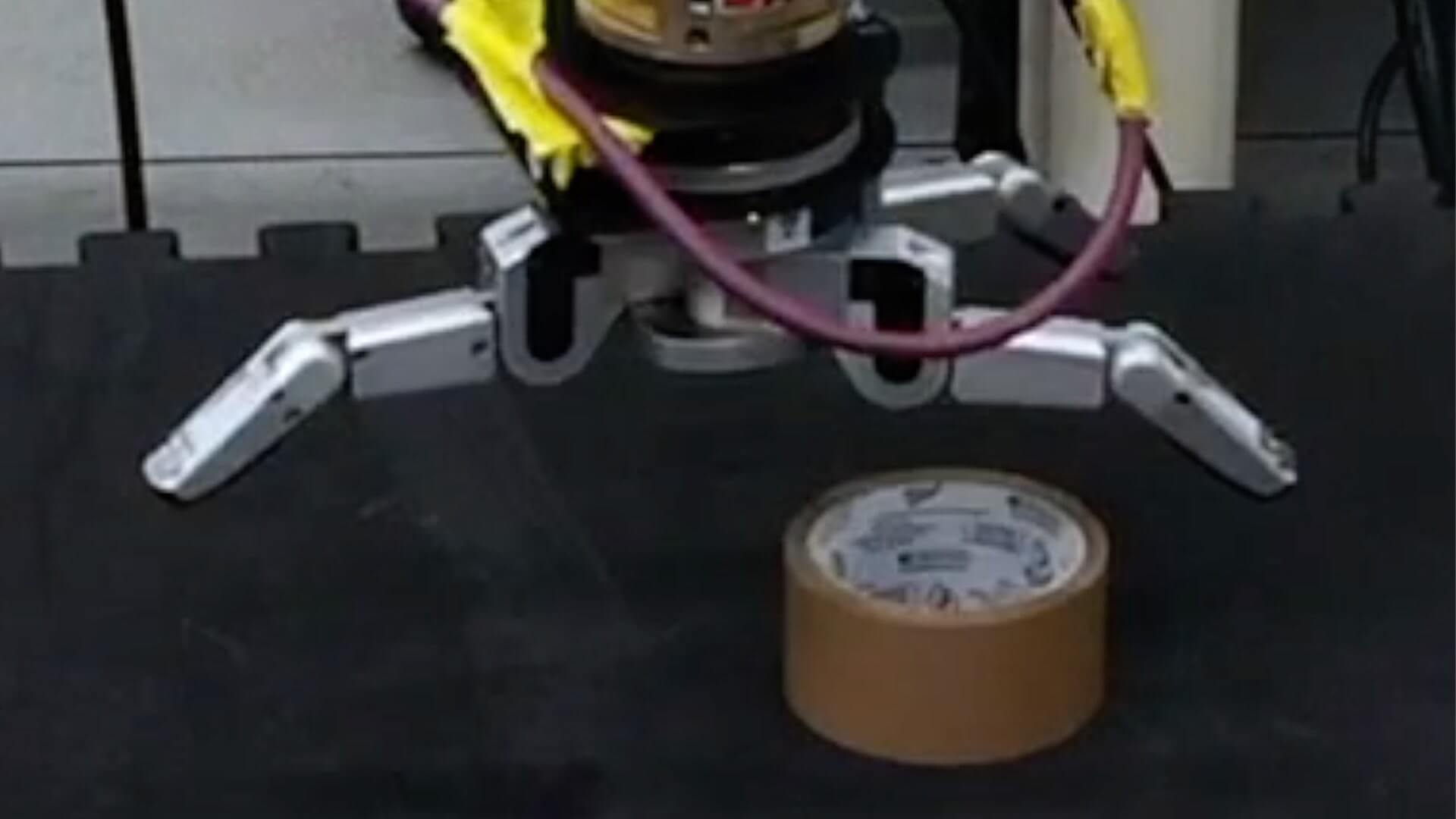}\caption{}
    \end{subfigure}
    \begin{subfigure}[t]{0.16\textwidth}
        \captionsetup{skip=0pt}\includegraphics[trim={200 0 200 0},clip,width=\linewidth]{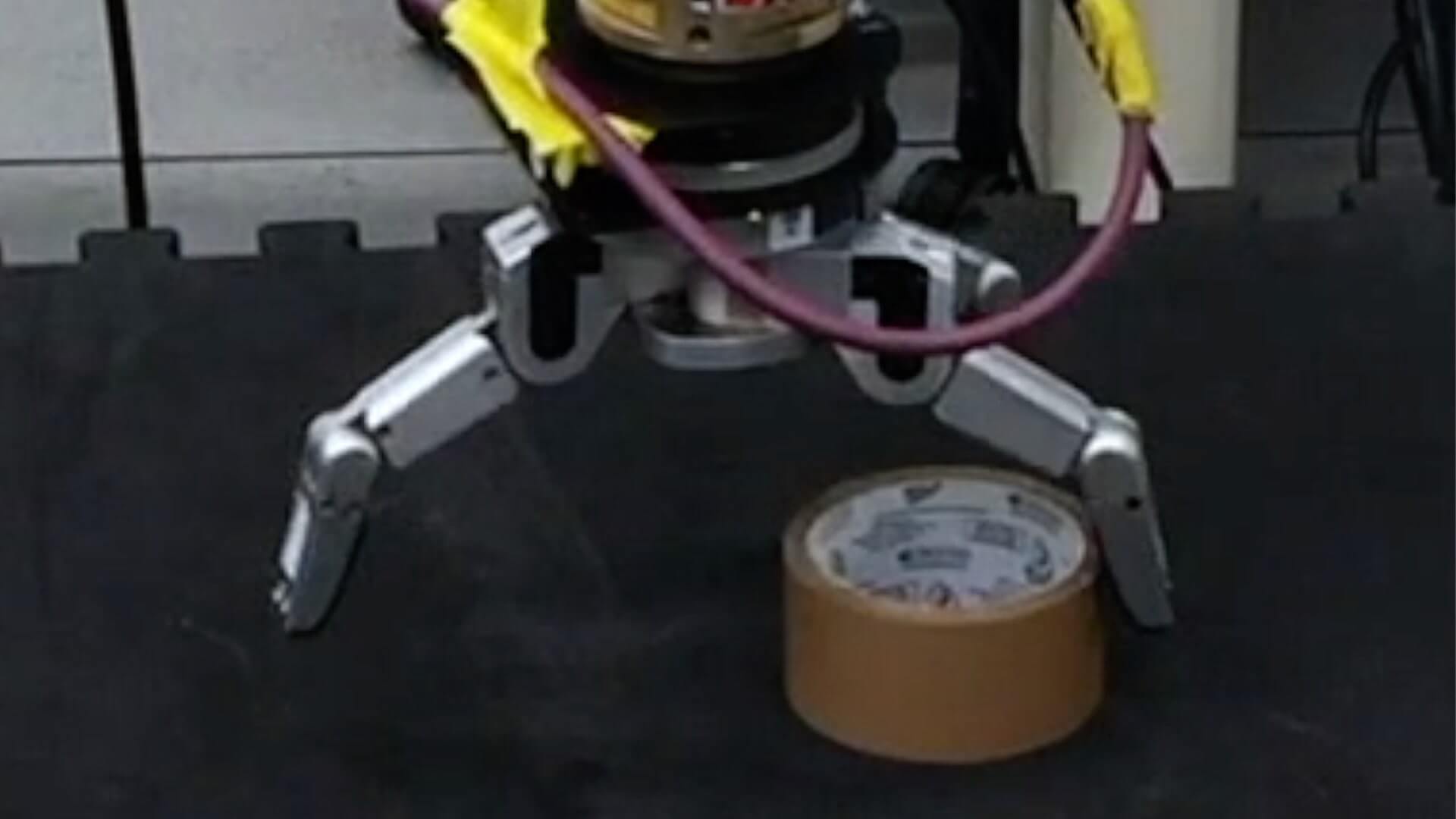}\caption{}
    \end{subfigure}
    \begin{subfigure}[t]{0.16\textwidth}
        \captionsetup{skip=0pt}\includegraphics[trim={200 0 200 0},clip,width=\linewidth]{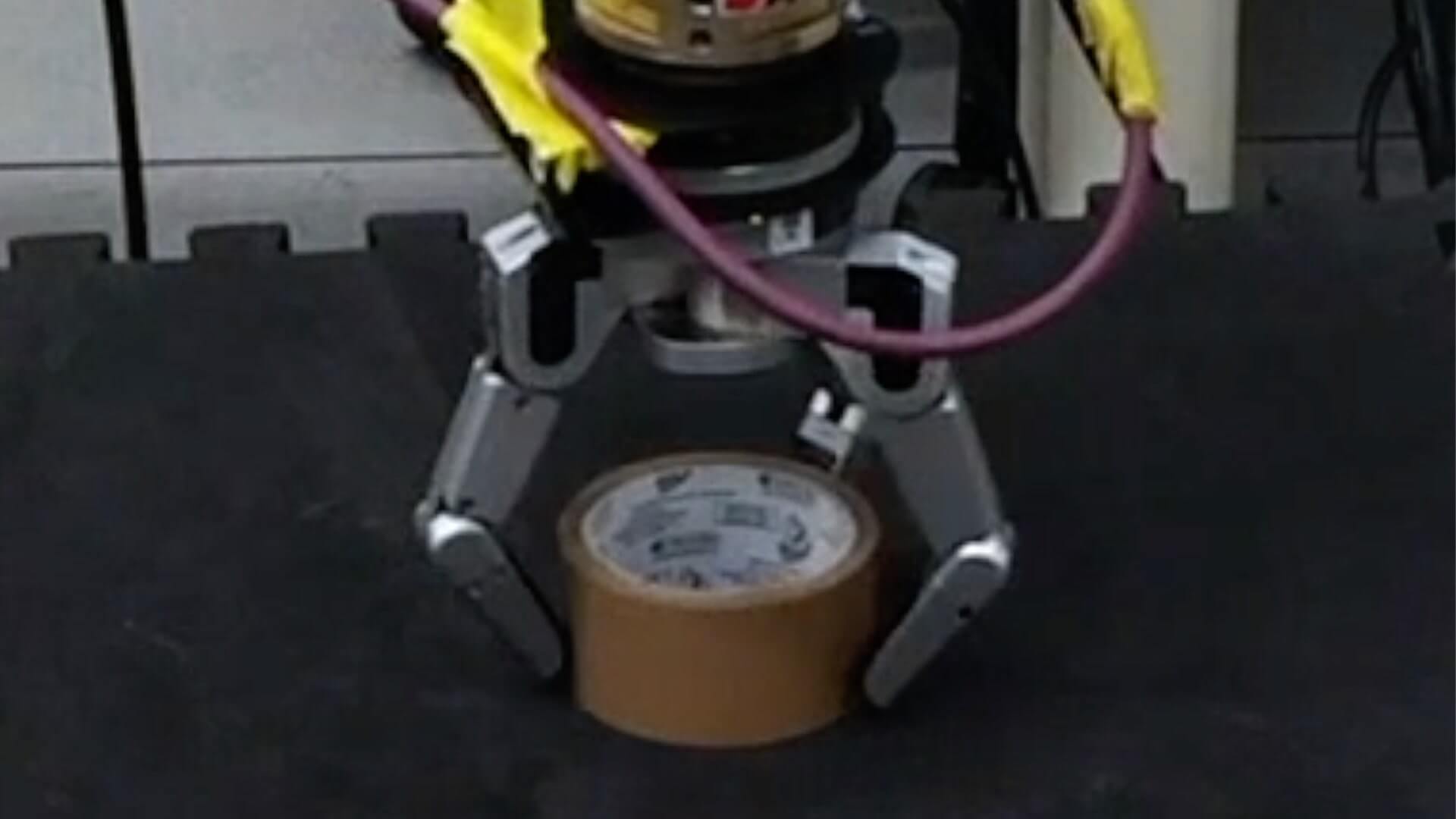}\caption{}
    \end{subfigure}
    \begin{subfigure}[t]{0.16\textwidth}
        \captionsetup{skip=0pt}\includegraphics[trim={200 0 200 0},clip,width=\linewidth]{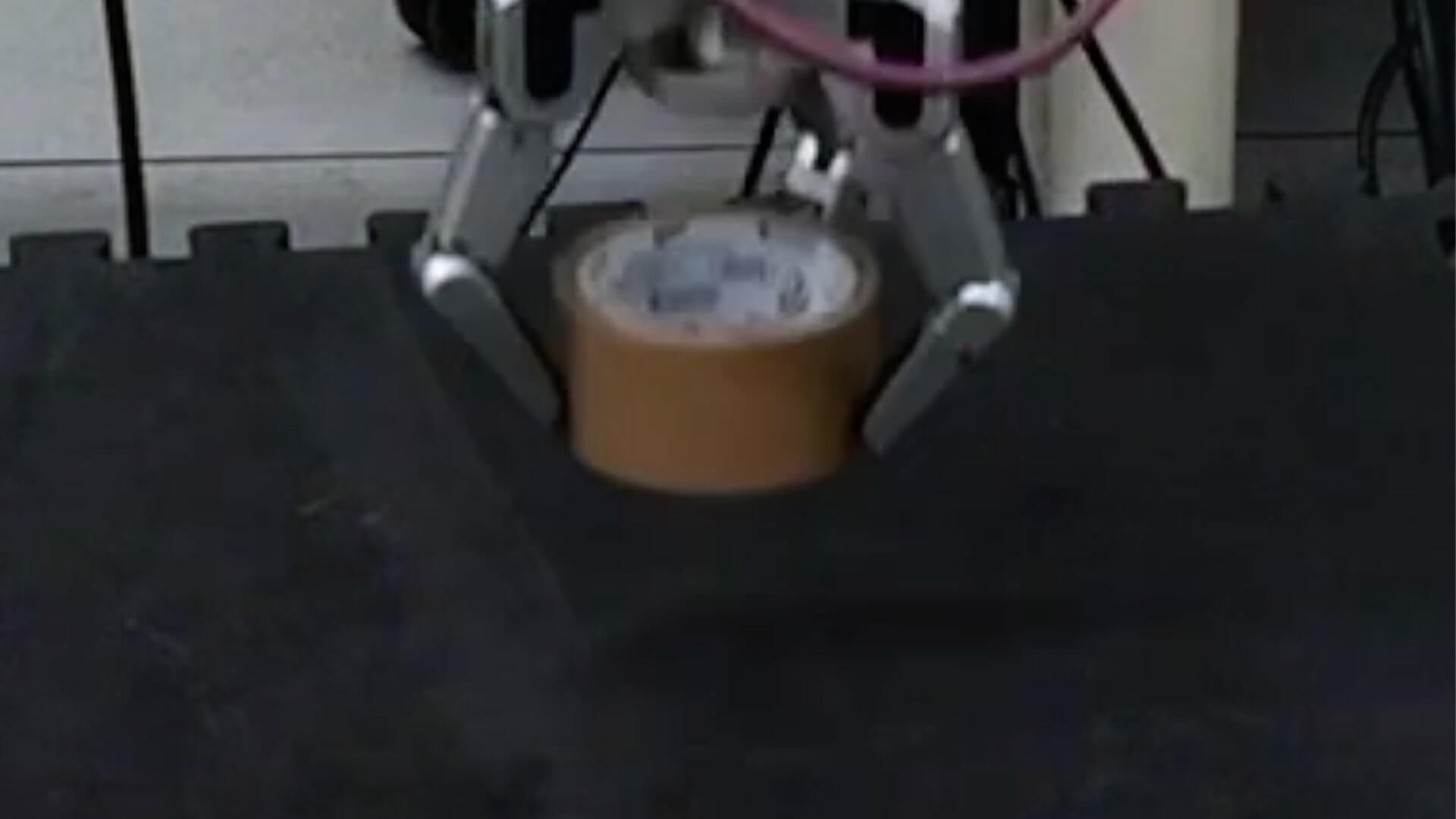}\caption{}
    \end{subfigure}
\setcounter{subfigure}{0}
    \begin{subfigure}[t]{0.16\textwidth}
        \captionsetup{skip=0pt}\includegraphics[trim={500 0 150 0},clip,width=\linewidth]{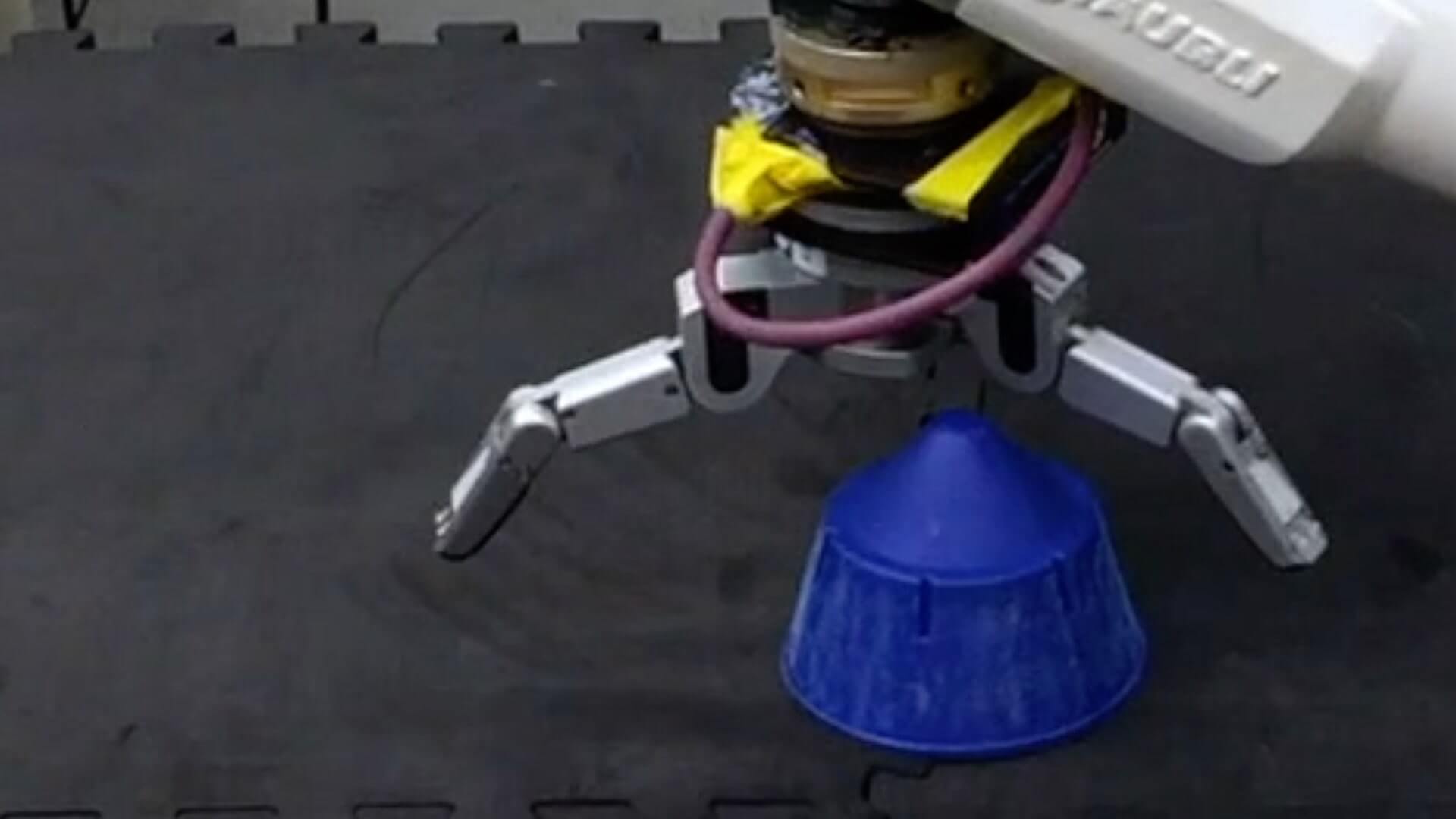}\caption{}
    \end{subfigure}
    \begin{subfigure}[t]{0.16\textwidth}
        \captionsetup{skip=0pt}\includegraphics[trim={500 0 150 0},clip,width=\linewidth]{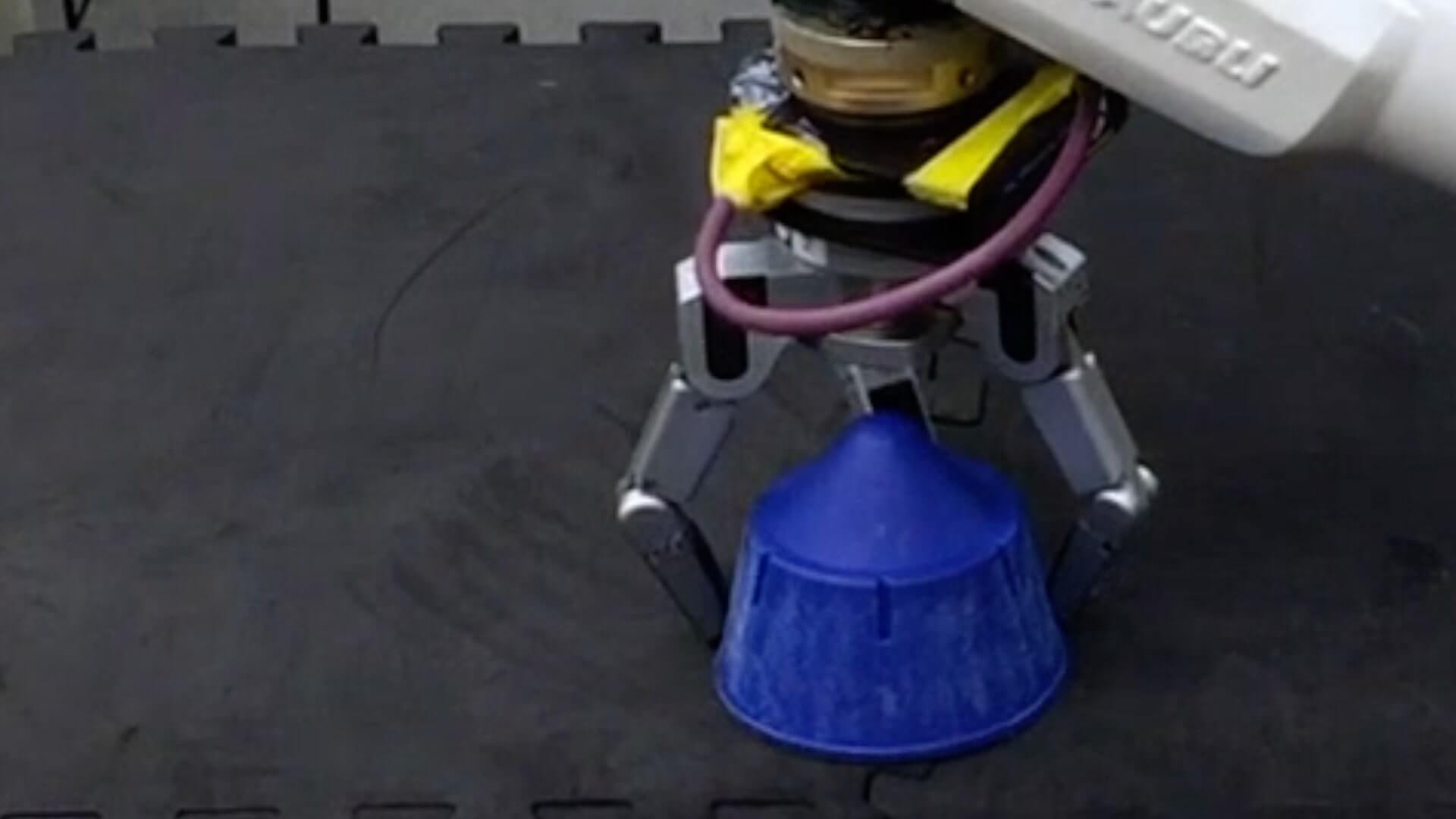}\caption{}
    \end{subfigure}
    \begin{subfigure}[t]{0.16\textwidth}
        \captionsetup{skip=0pt}\includegraphics[trim={500 0 150 0},clip,width=\linewidth]{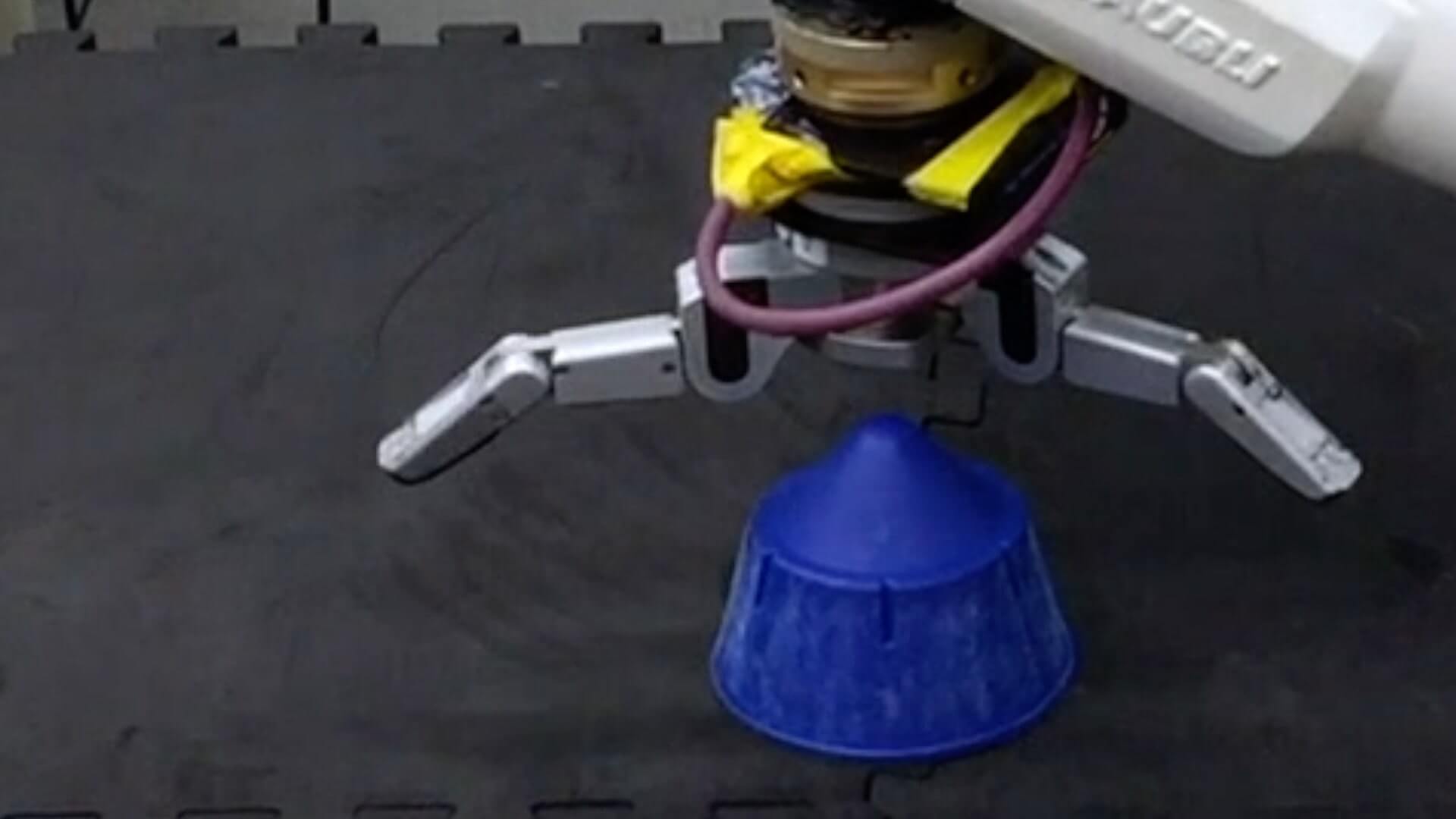}\caption{}
    \end{subfigure}
    \begin{subfigure}[t]{0.16\textwidth}
        \captionsetup{skip=0pt}\includegraphics[trim={500 0 150 0},clip,width=\linewidth]{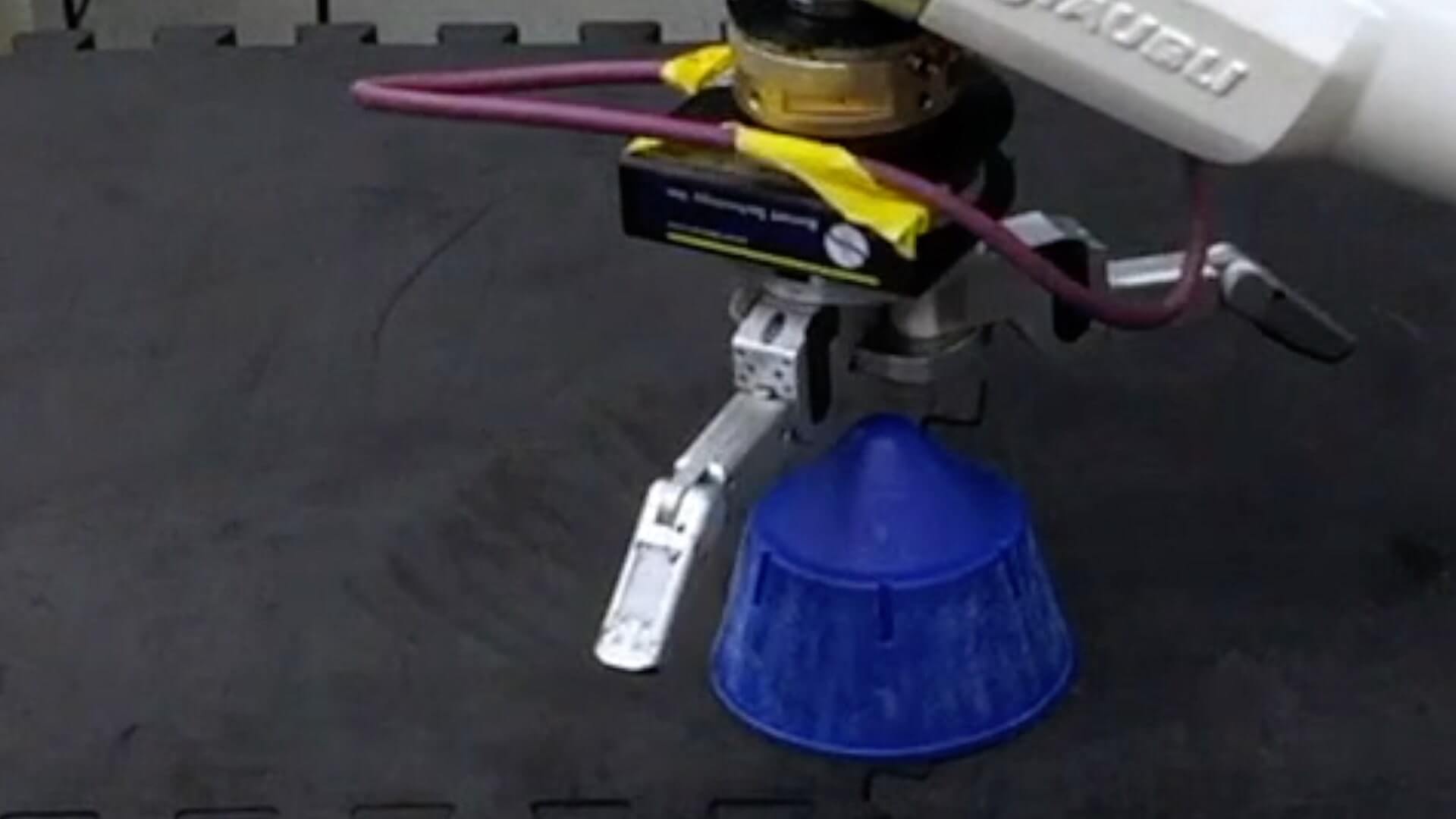}\caption{}
    \end{subfigure}
    \begin{subfigure}[t]{0.16\textwidth}
        \captionsetup{skip=0pt}\includegraphics[trim={500 0 150 0},clip,width=\linewidth]{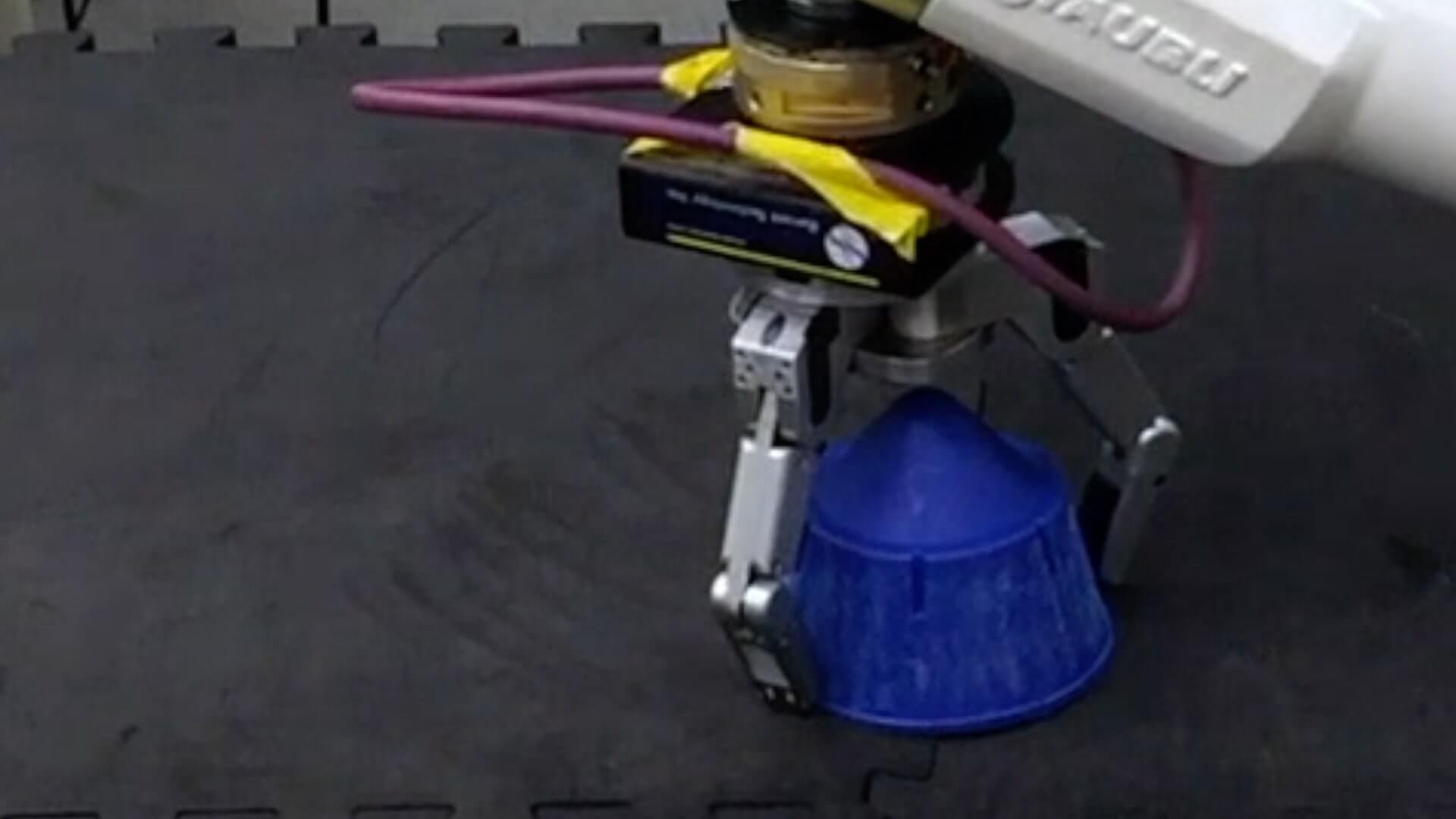}\caption{}
    \end{subfigure}
    \begin{subfigure}[t]{0.16\textwidth}
        \captionsetup{skip=0pt}\includegraphics[trim={500 0 150 0},clip,width=\linewidth]{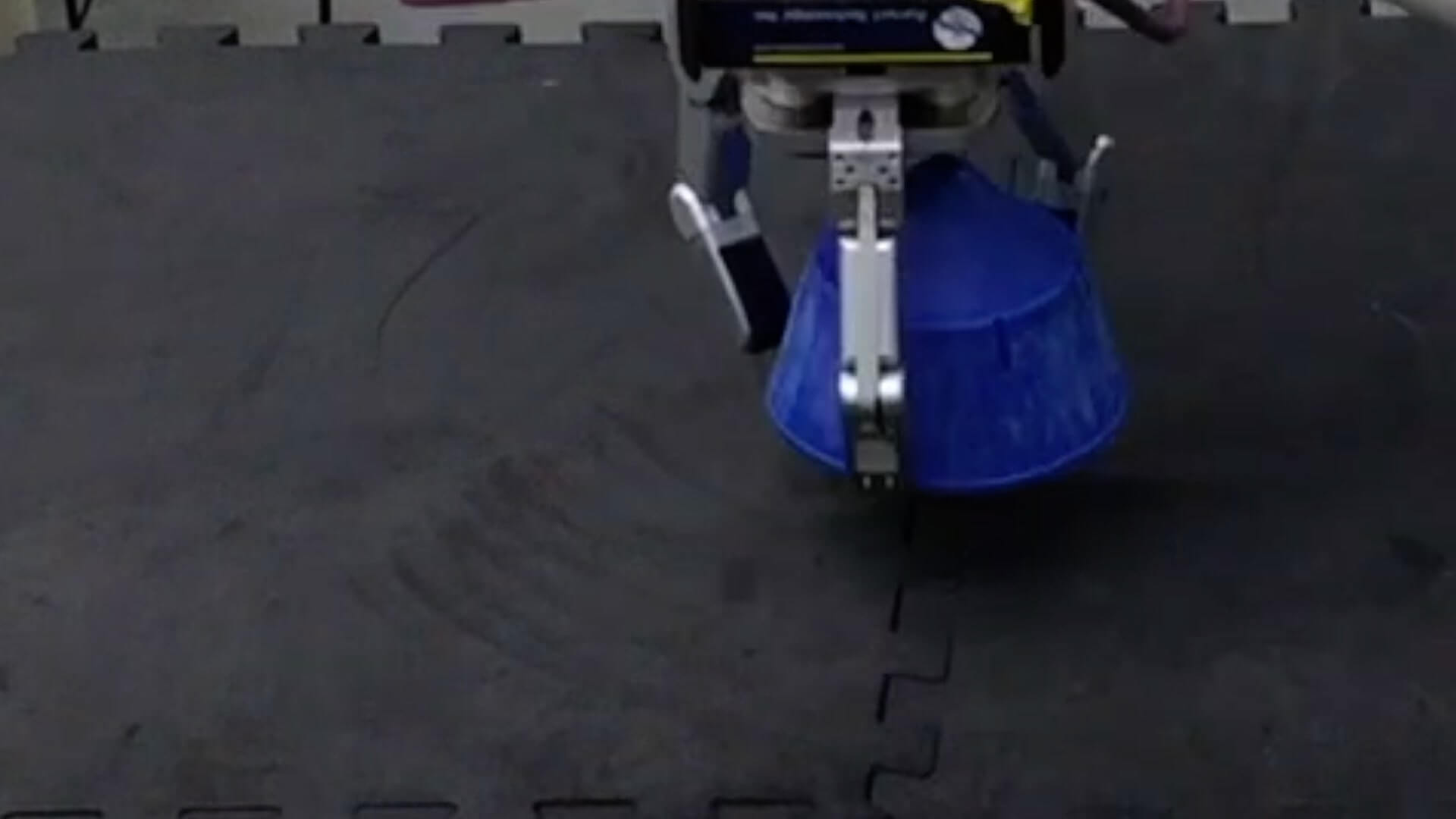}\caption{}
    \end{subfigure}
\setcounter{subfigure}{0}
    \begin{subfigure}[t]{0.16\textwidth}
        \captionsetup{skip=0pt}\includegraphics[trim={500 0 150 0},clip,width=\linewidth]{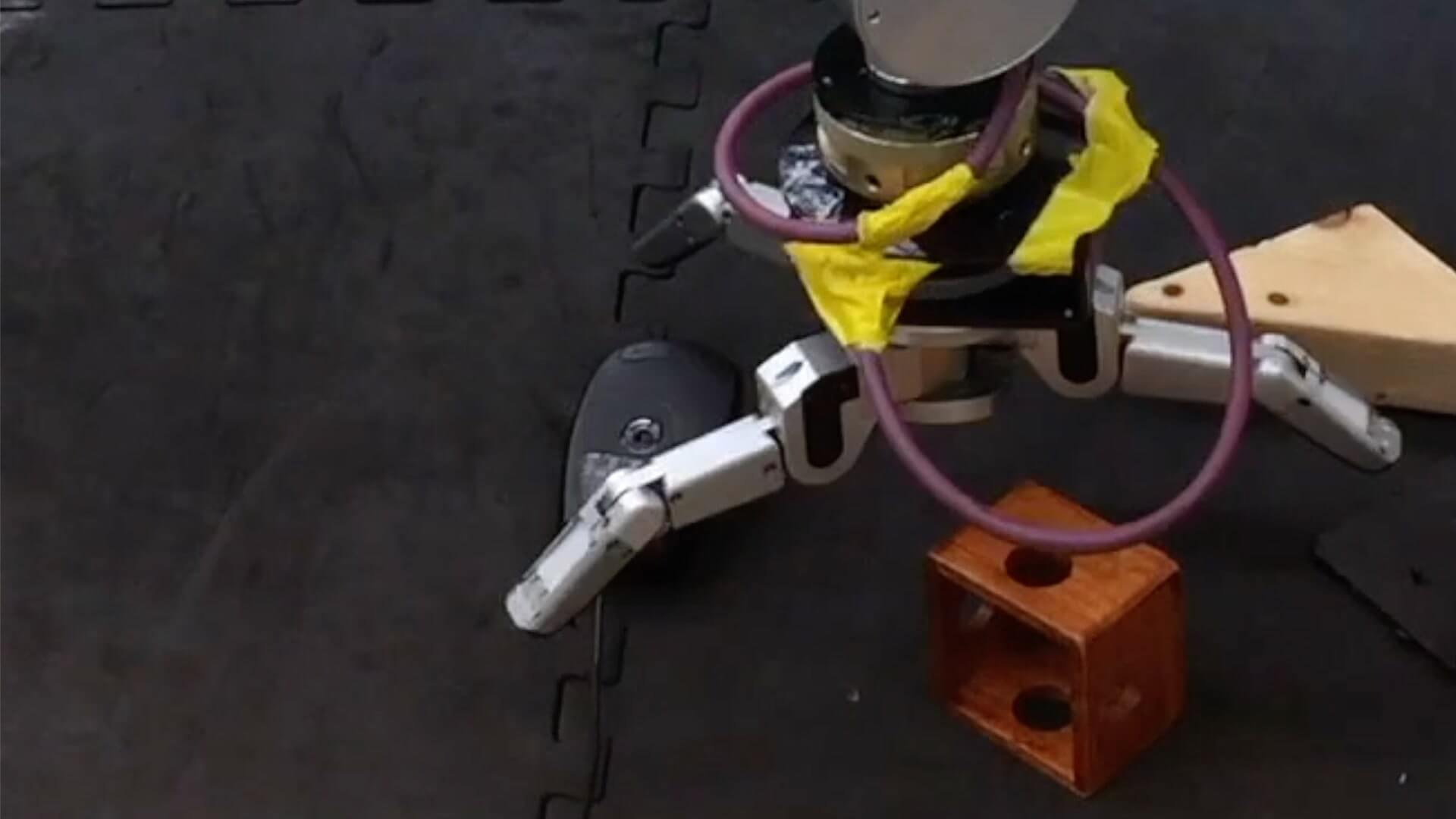}\caption{}
    \end{subfigure}
    \begin{subfigure}[t]{0.16\textwidth}
        \captionsetup{skip=0pt}\includegraphics[trim={500 0 150 0},clip,width=\linewidth]{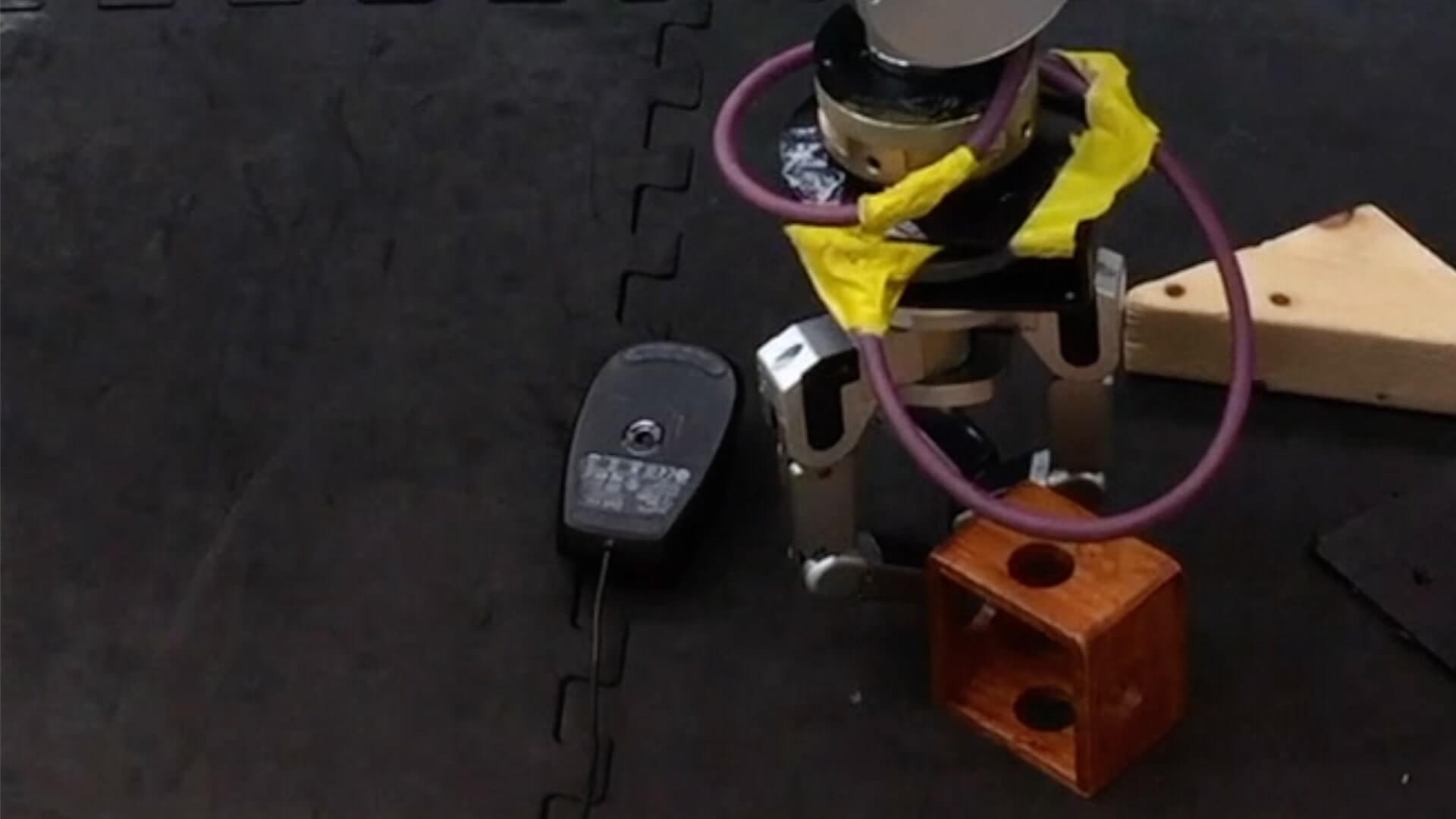}\caption{}
    \end{subfigure}
    \begin{subfigure}[t]{0.16\textwidth}
        \captionsetup{skip=0pt}\includegraphics[trim={500 0 150 0},clip,width=\linewidth]{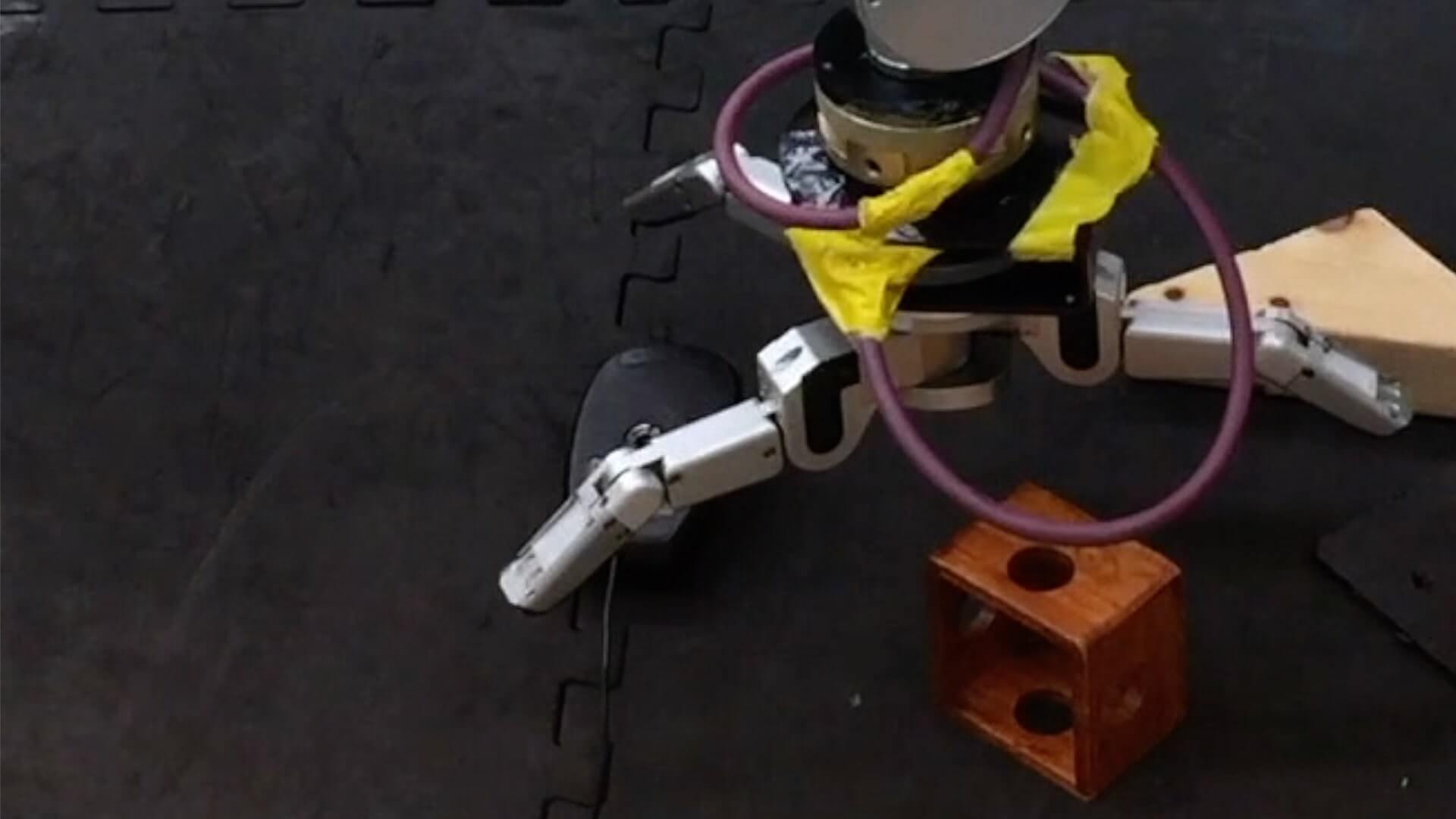}\caption{}
    \end{subfigure}
    \begin{subfigure}[t]{0.16\textwidth}
        \captionsetup{skip=0pt}\includegraphics[trim={500 0 150 0},clip,width=\linewidth]{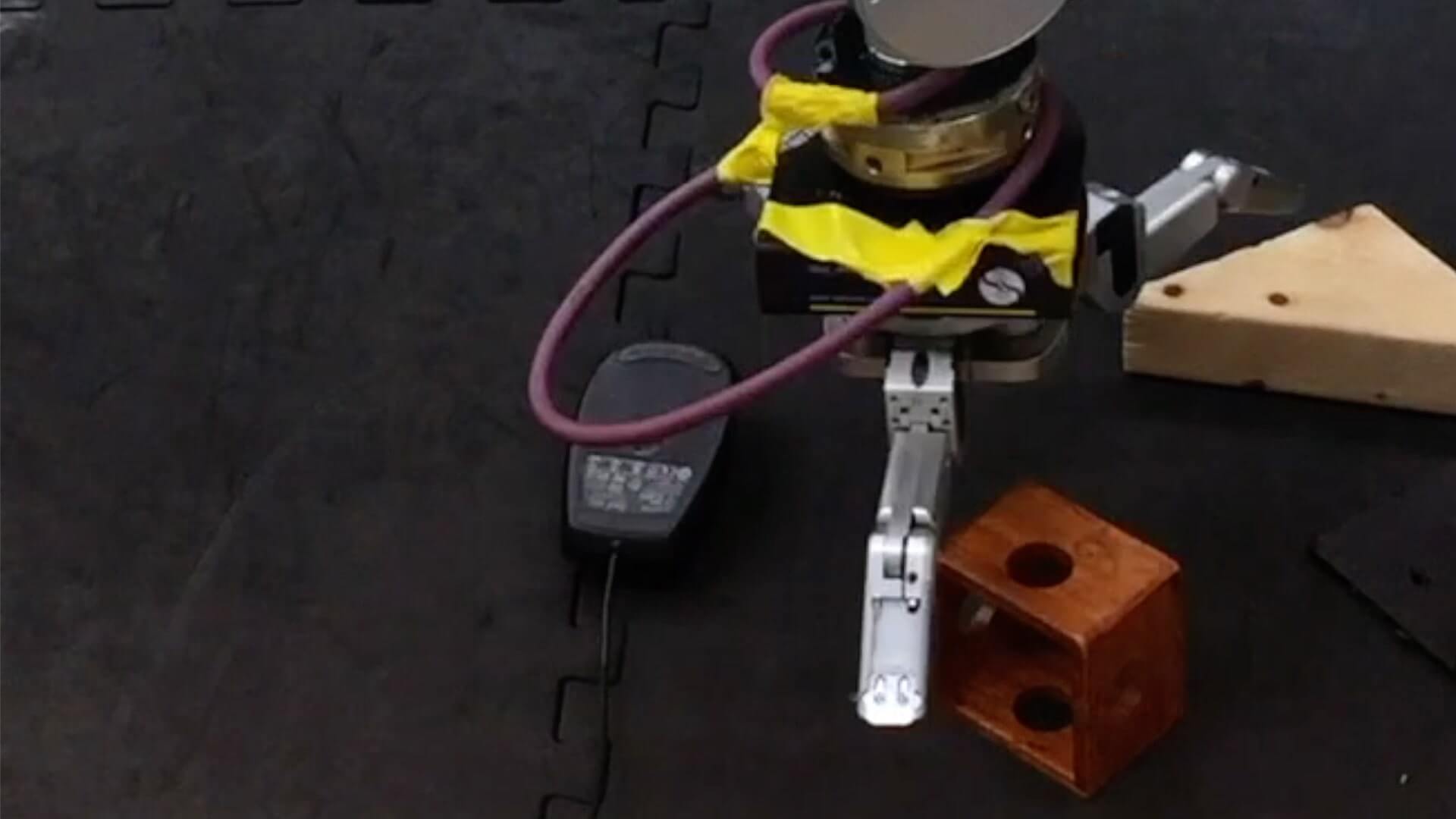}\caption{}
    \end{subfigure}
    \begin{subfigure}[t]{0.16\textwidth}
        \captionsetup{skip=0pt}\includegraphics[trim={500 0 150 0},clip,width=\linewidth]{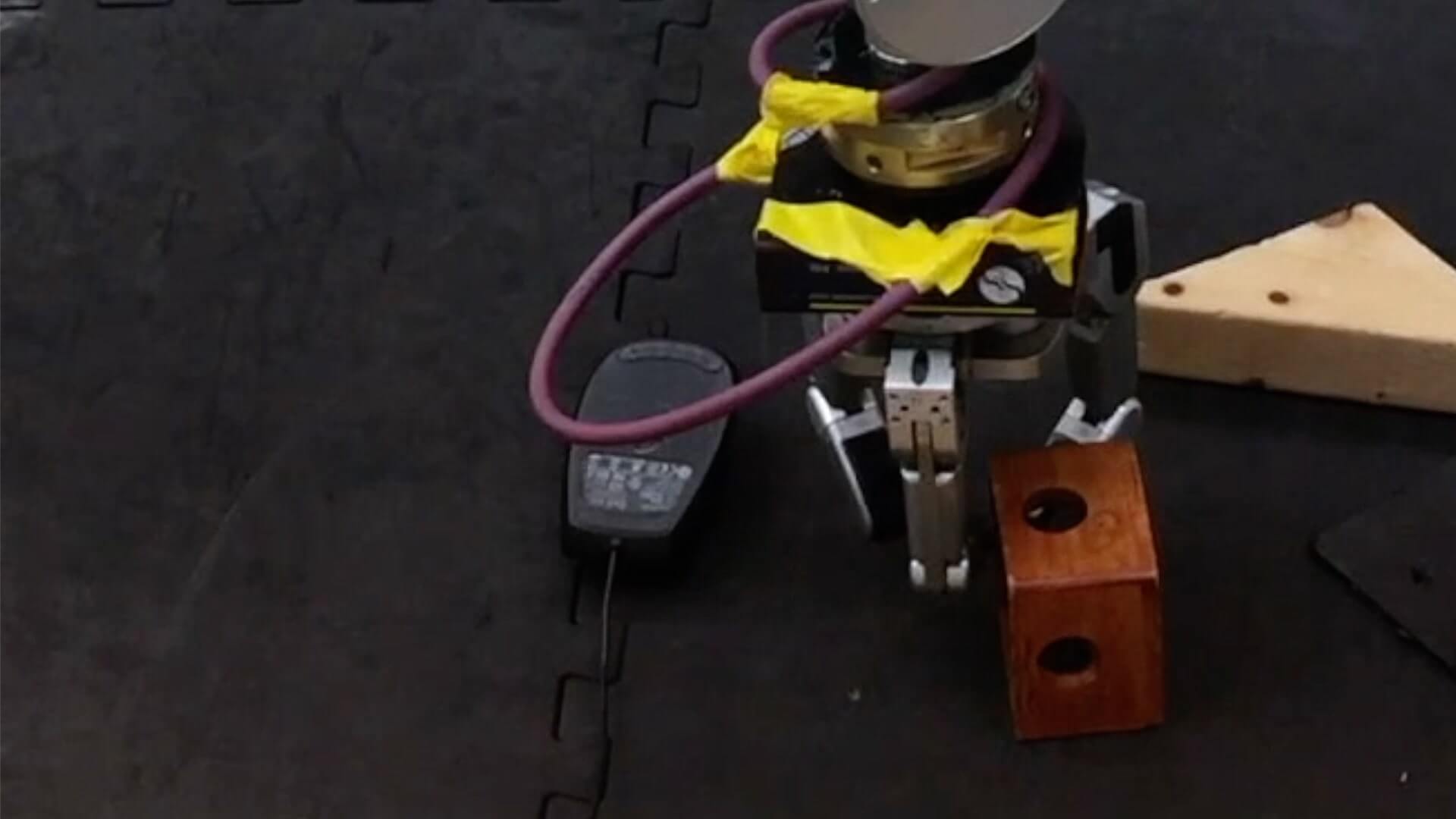}\caption{}
    \end{subfigure}
    \begin{subfigure}[t]{0.16\textwidth}
        \captionsetup{skip=0pt}\includegraphics[trim={500 0 150 0},clip,width=\linewidth]{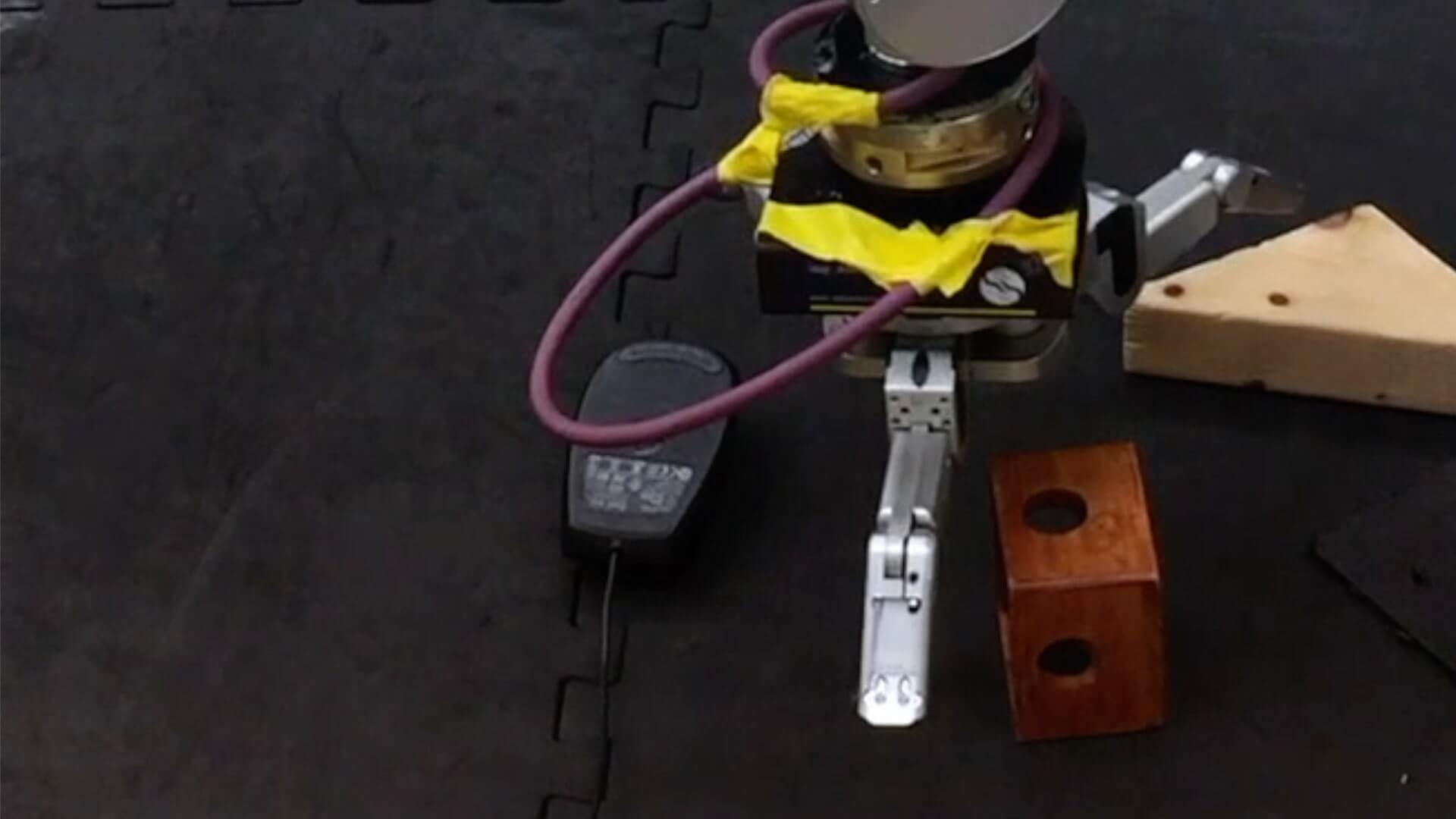}\caption{}
    \end{subfigure}
    \begin{subfigure}[t]{0.16\textwidth}
        \captionsetup{skip=0pt}\includegraphics[trim={500 0 150 0},clip,width=\linewidth]{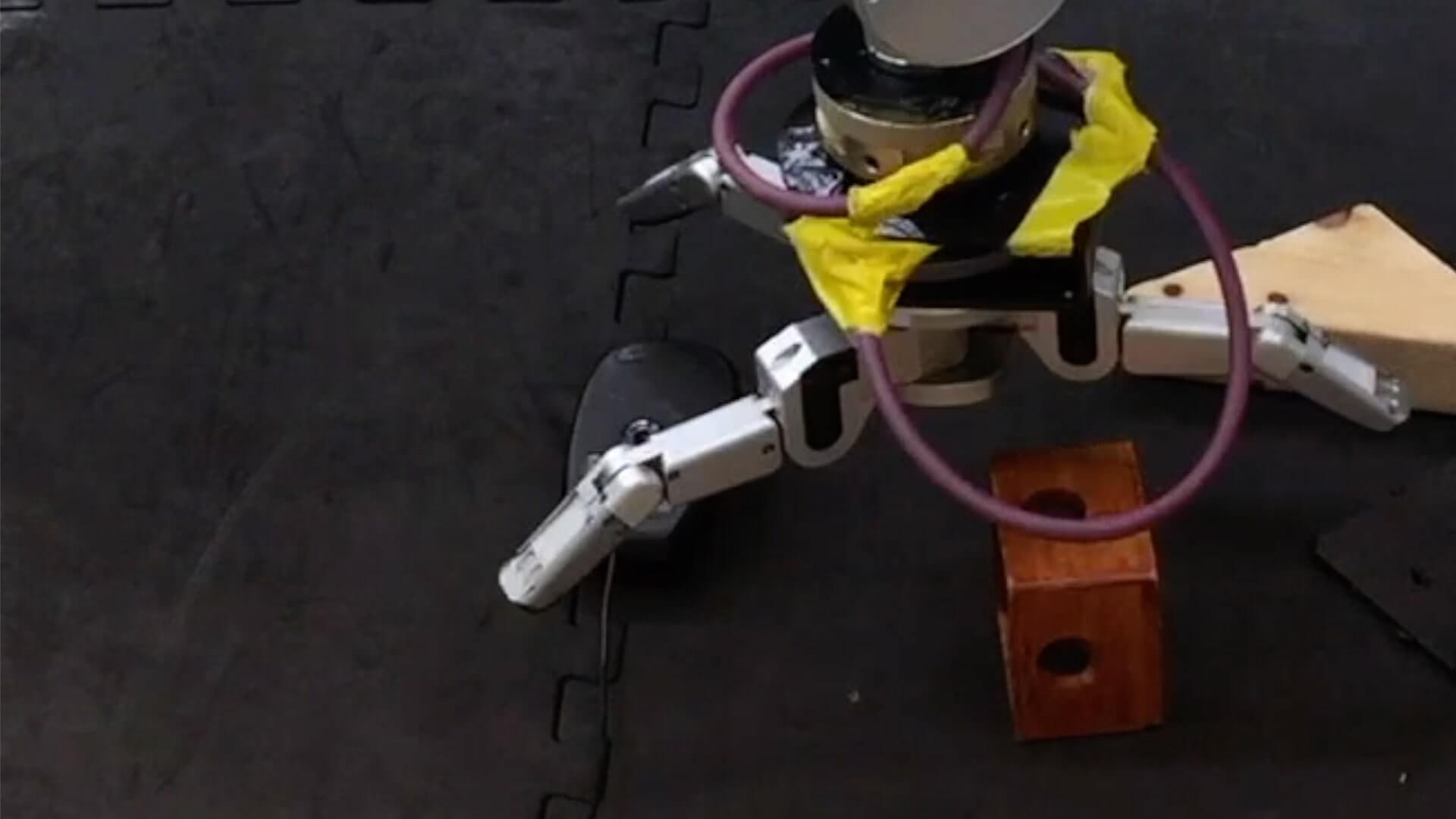}\caption{}
    \end{subfigure}
    \begin{subfigure}[t]{0.16\textwidth}
        \captionsetup{skip=0pt}\includegraphics[trim={500 0 150 0},clip,width=\linewidth]{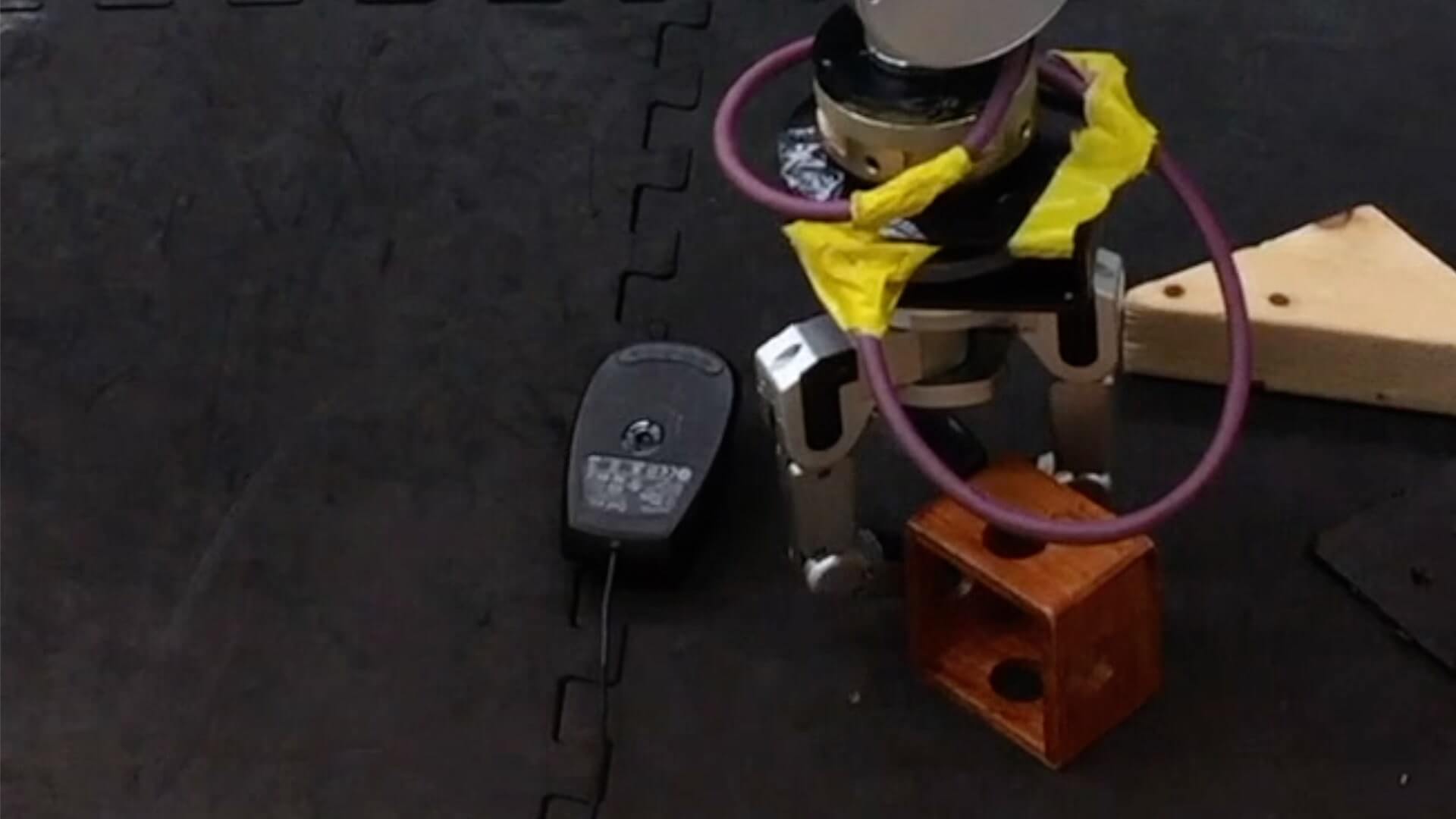}\caption{}
    \end{subfigure}
    \begin{subfigure}[t]{0.16\textwidth}
        \captionsetup{skip=0pt}\includegraphics[trim={500 0 150 0},clip,width=\linewidth]{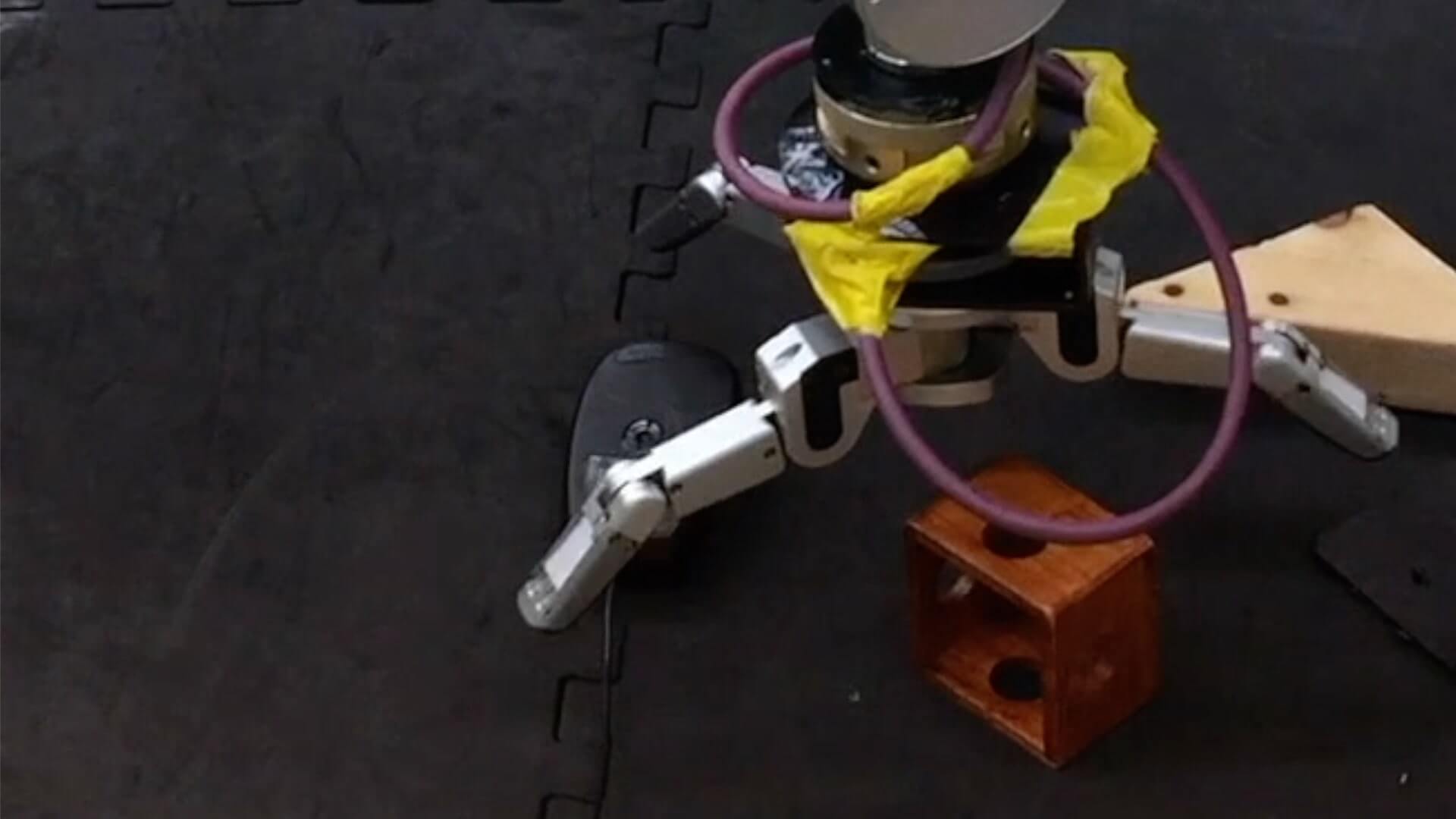}\caption{}
    \end{subfigure}
    \begin{subfigure}[t]{0.16\textwidth}
        \captionsetup{skip=0pt}\includegraphics[trim={500 0 150 0},clip,width=\linewidth]{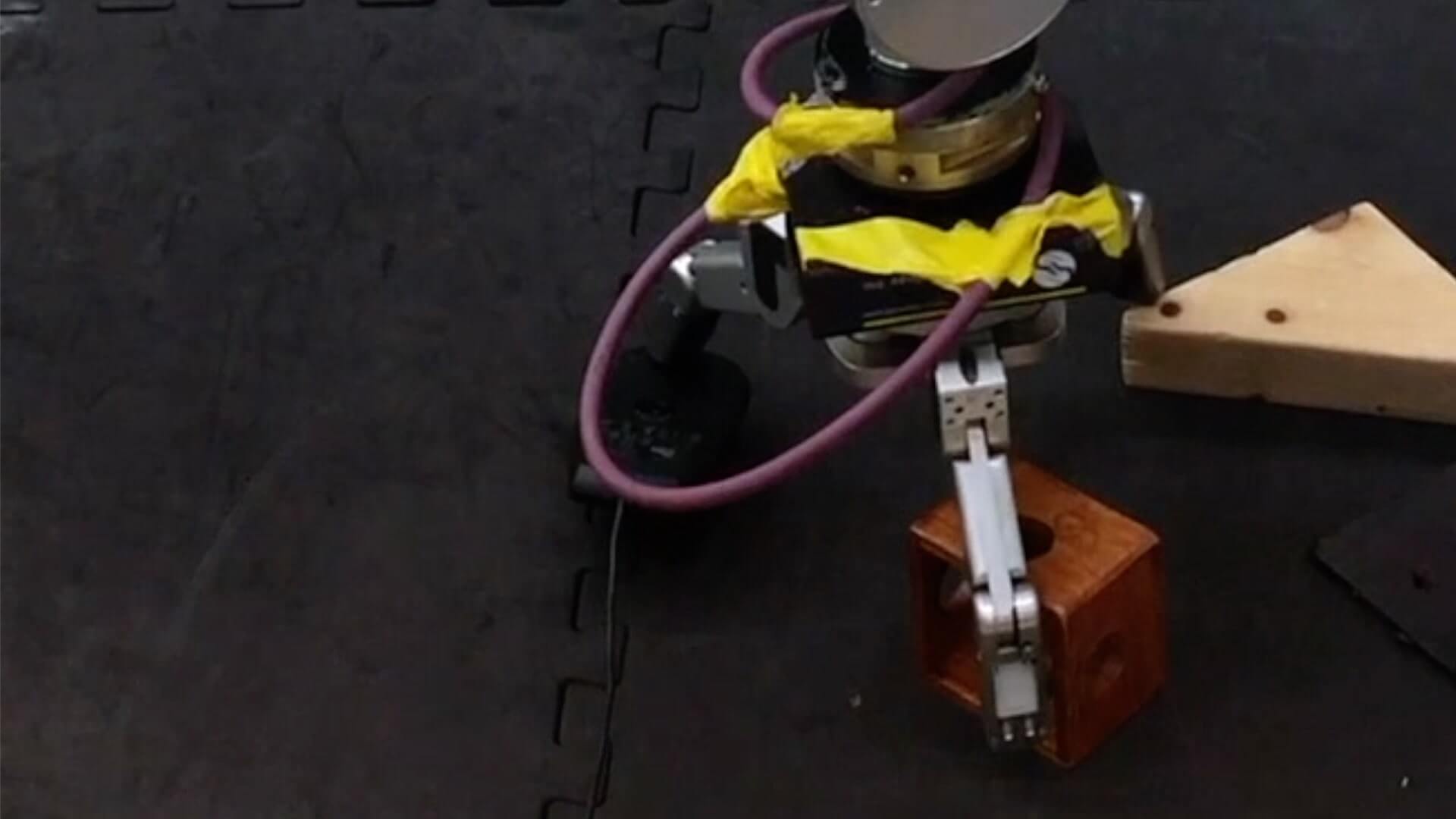}\caption{}
    \end{subfigure}
    \begin{subfigure}[t]{0.16\textwidth}
        \captionsetup{skip=0pt}\includegraphics[trim={500 0 150 0},clip,width=\linewidth]{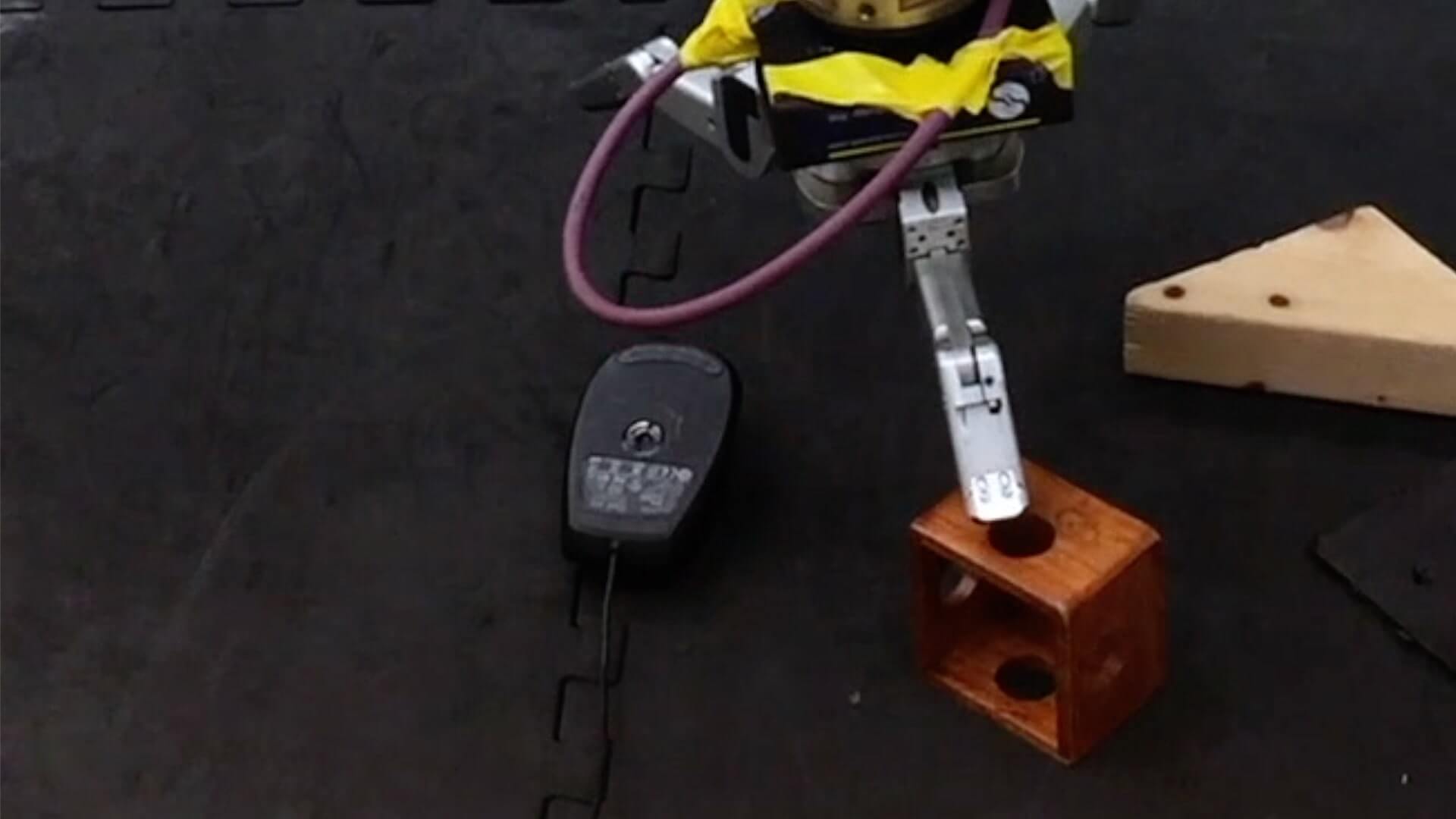}\caption{}
    \end{subfigure}
    \begin{subfigure}[t]{0.16\textwidth}
        \captionsetup{skip=0pt}\includegraphics[trim={500 0 150 0},clip,width=\linewidth]{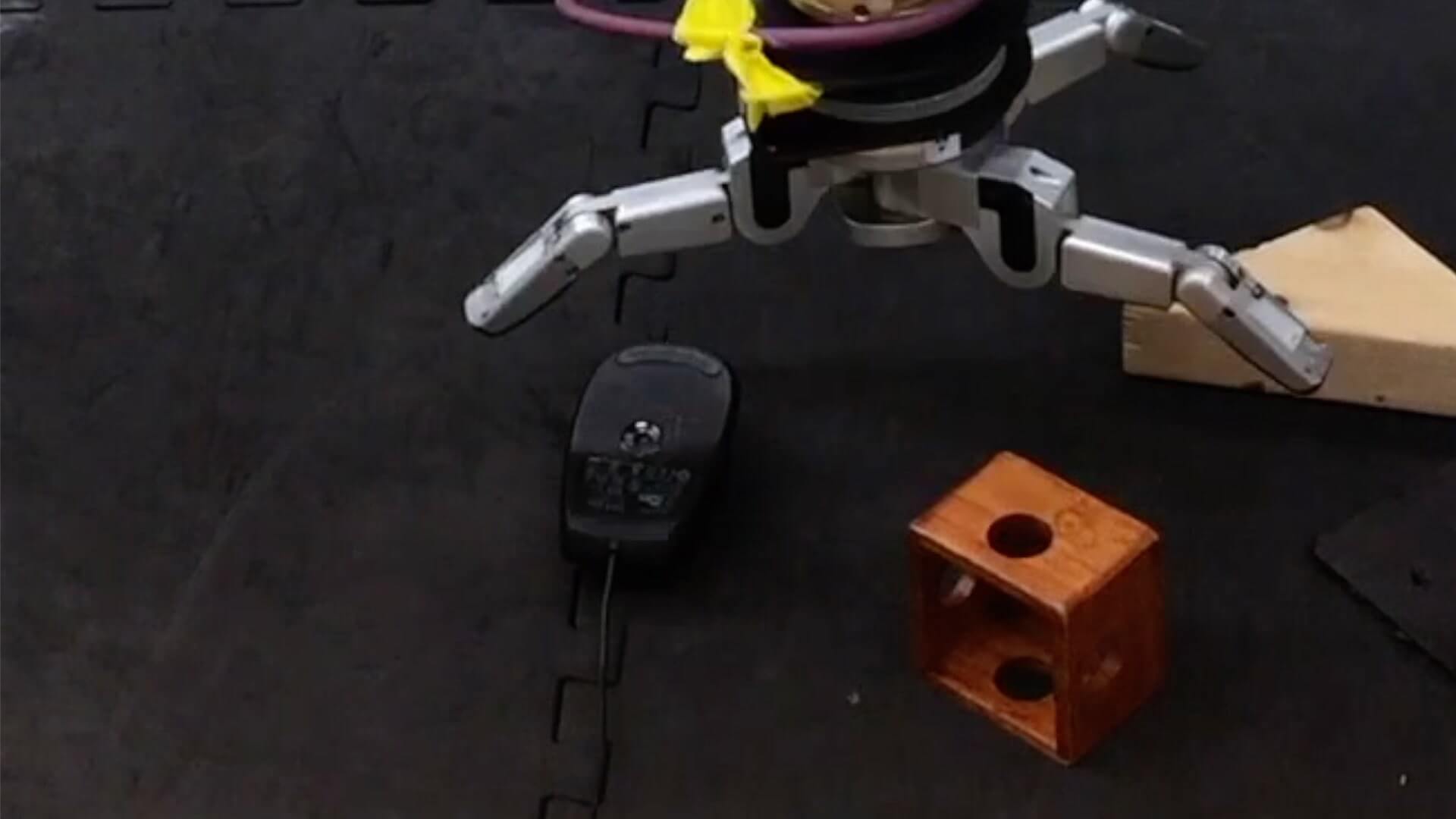}\caption{}
    \end{subfigure}
    \begin{subfigure}[t]{0.16\textwidth}
        \captionsetup{skip=0pt}\includegraphics[trim={500 0 150 0},clip,width=\linewidth]{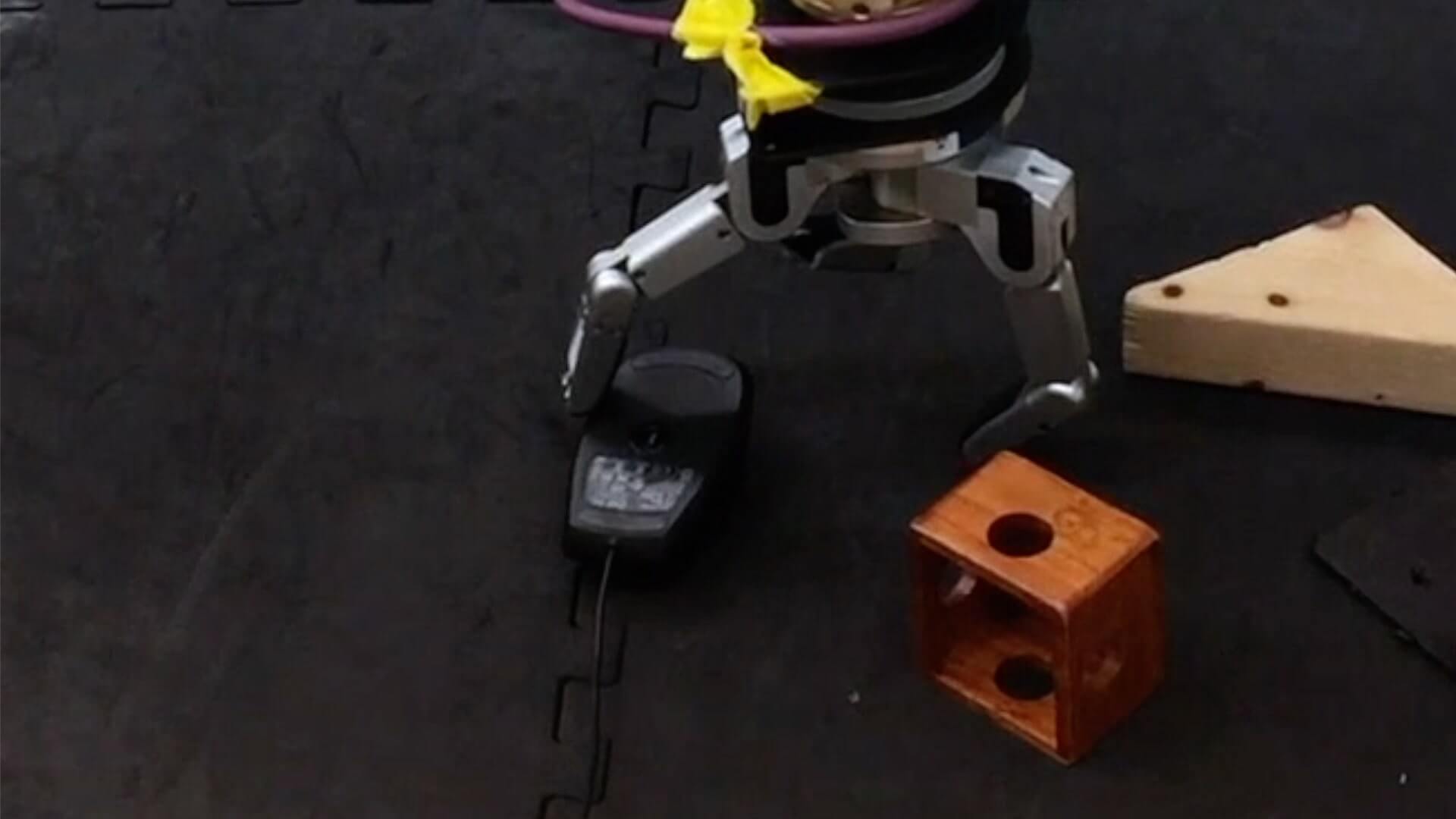}\caption{}
    \end{subfigure}
    \begin{subfigure}[t]{0.16\textwidth}
        \captionsetup{skip=0pt}\includegraphics[trim={500 0 150 0},clip,width=\linewidth]{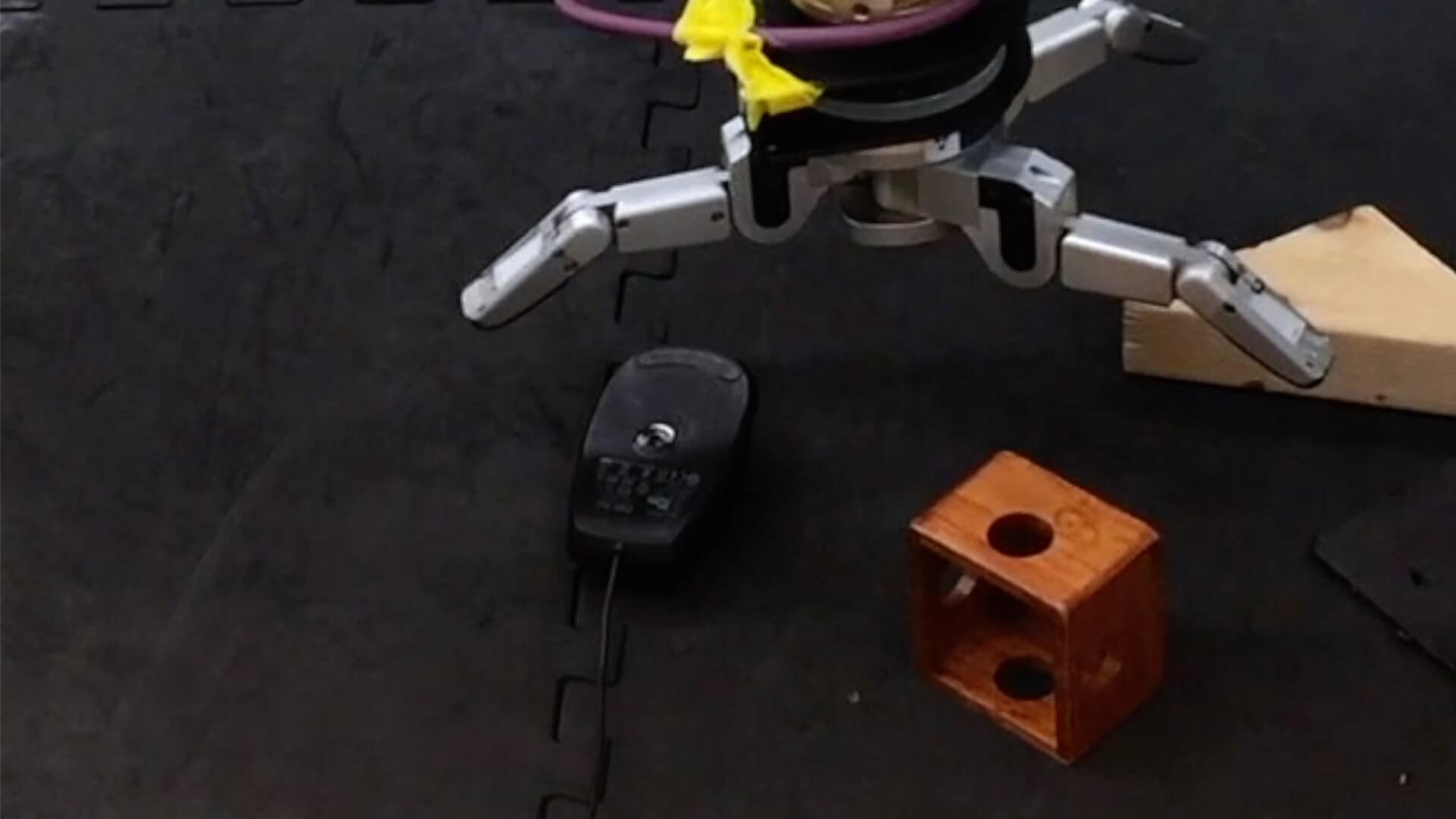}\caption{}
    \end{subfigure}
    \begin{subfigure}[t]{0.16\textwidth}
        \captionsetup{skip=0pt}\includegraphics[trim={500 0 150 0},clip,width=\linewidth]{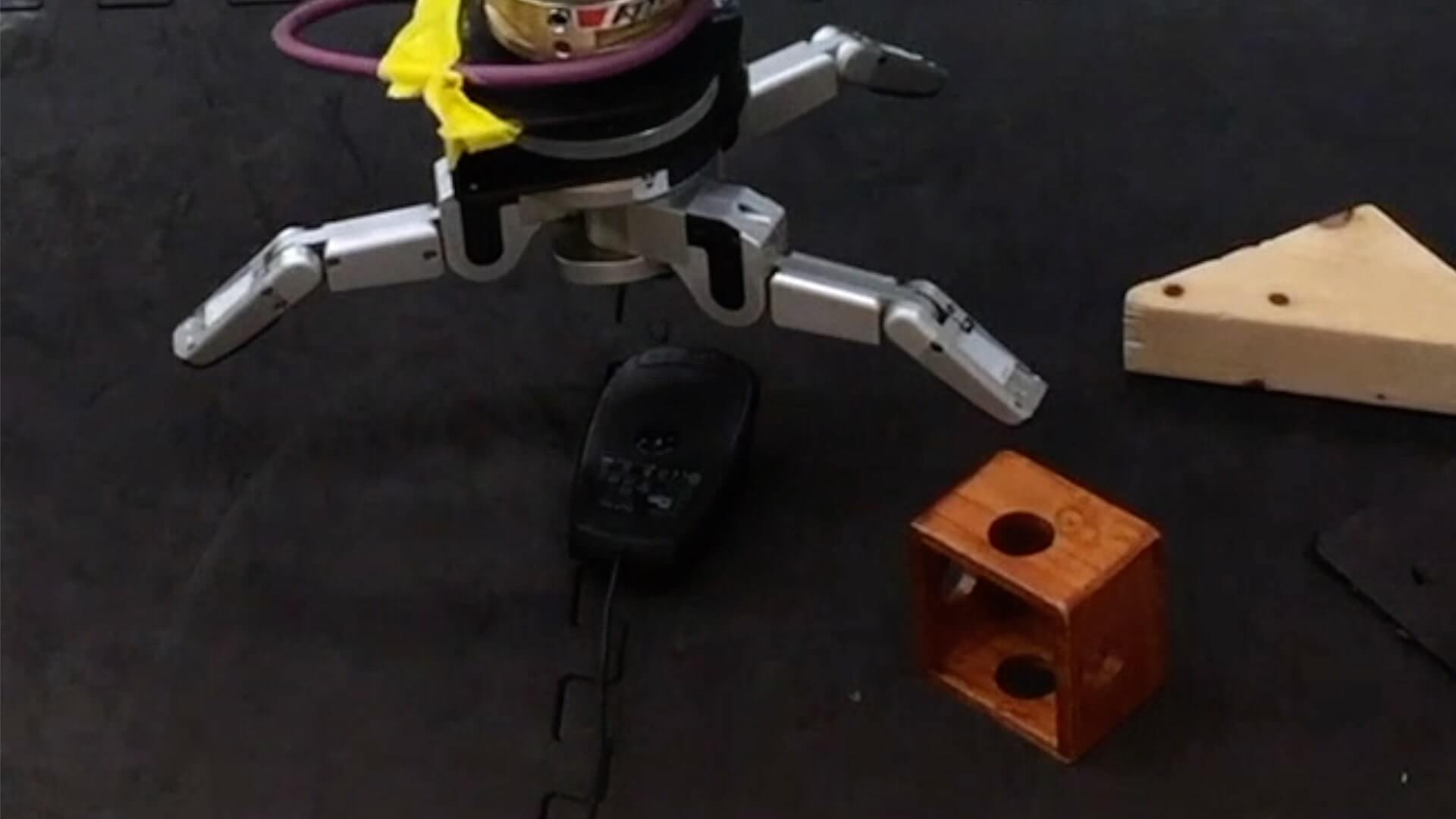}\caption{}
    \end{subfigure}
    \begin{subfigure}[t]{0.16\textwidth}
        \captionsetup{skip=0pt}\includegraphics[trim={500 0 150 0},clip,width=\linewidth]{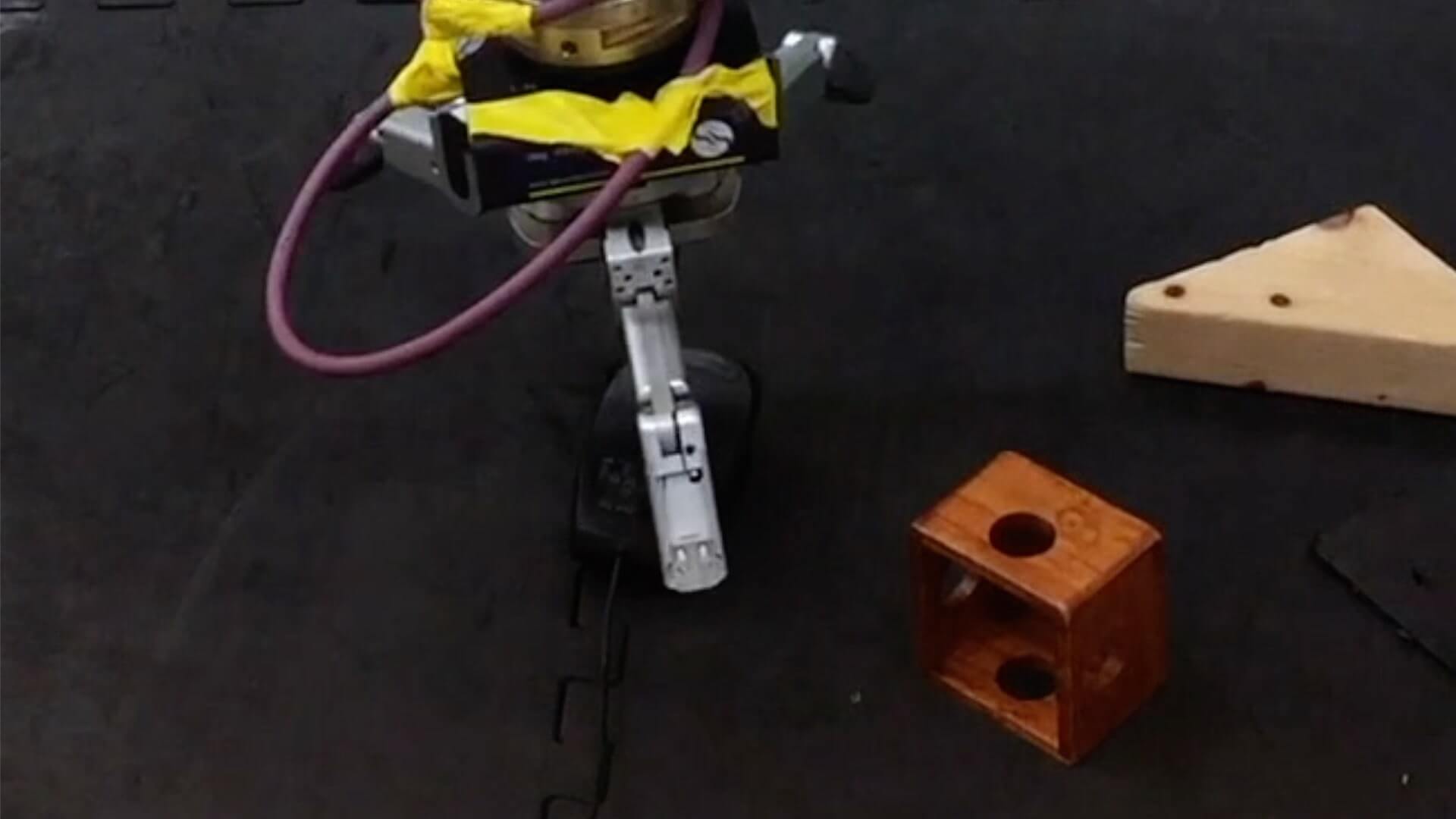}\caption{}
    \end{subfigure}
    \begin{subfigure}[t]{0.16\textwidth}
        \captionsetup{skip=0pt}\includegraphics[trim={500 0 150 0},clip,width=\linewidth]{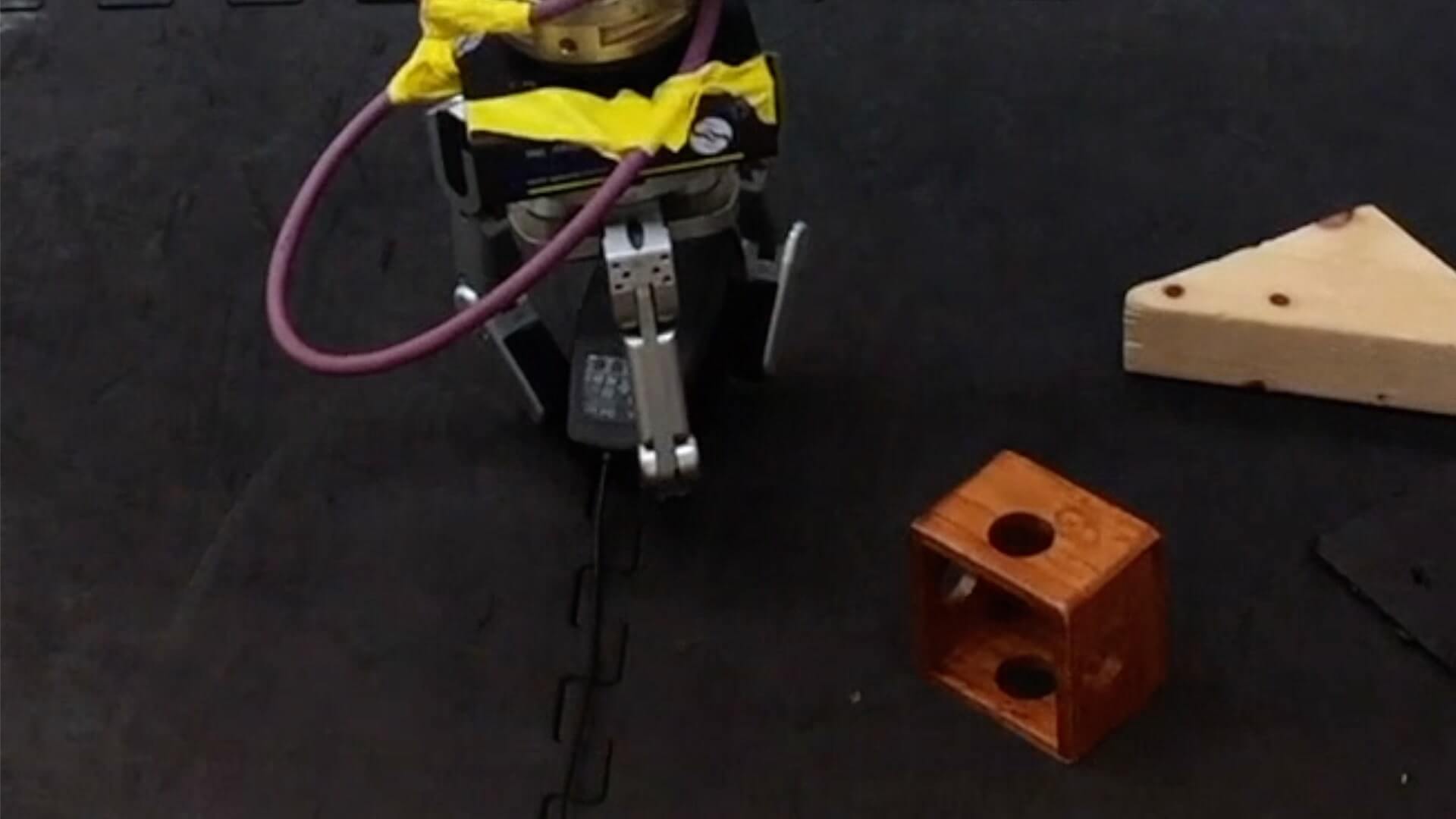}\caption{}
    \end{subfigure}
    \begin{subfigure}[t]{0.16\textwidth}
        \captionsetup{skip=0pt}\includegraphics[trim={500 0 150 0},clip,width=\linewidth]{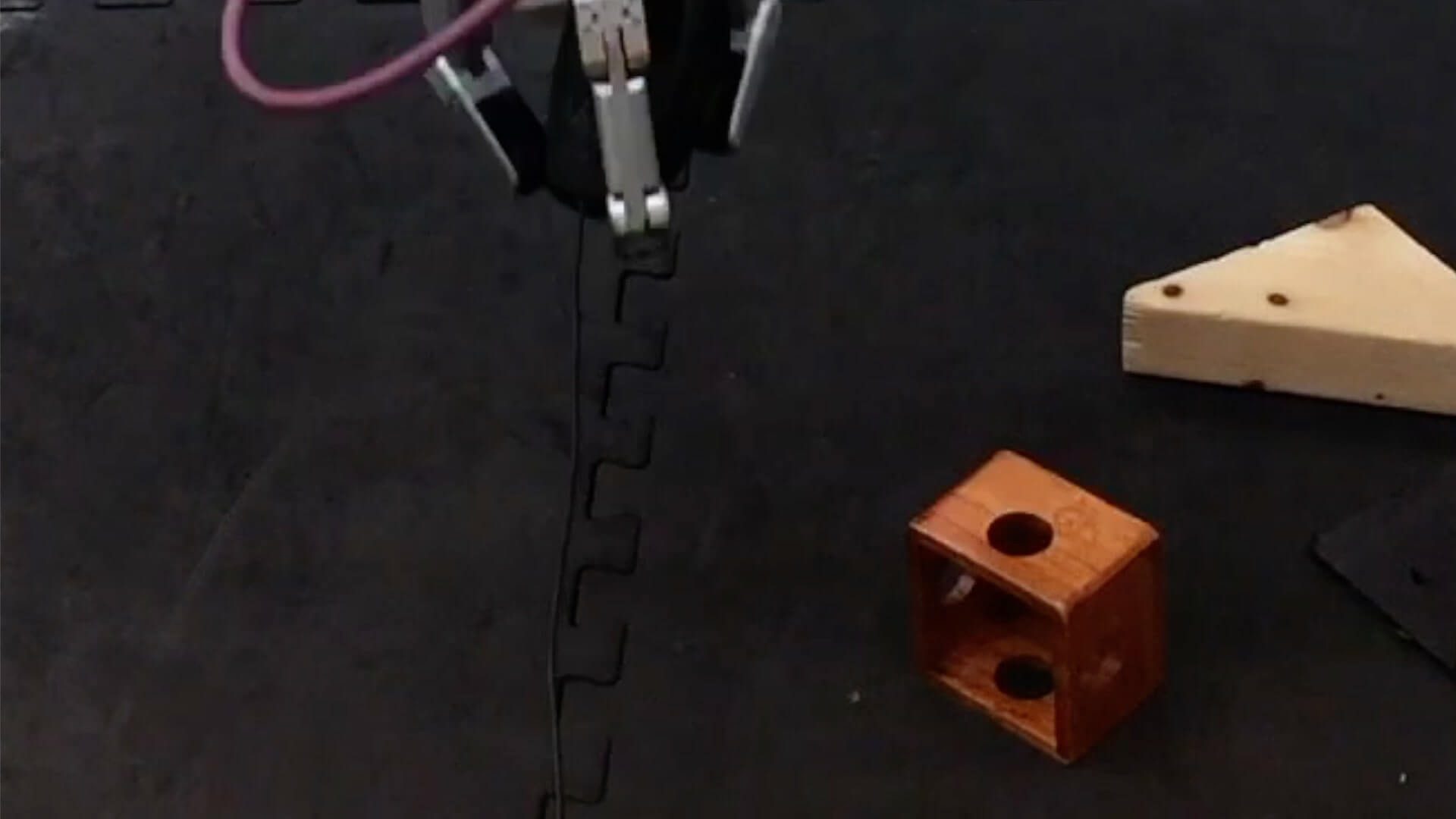}\caption{}
    \end{subfigure}
    \captionsetup{skip=3pt}
    \caption{\textbf{Learned Closed-Loop Behaviors of MAT.} Each series of images starts from \textbf{(a)}.}
    \label{fig:behaviors}
\end{figure}

Figure~\ref{fig:behaviors} shows the typical closed-loop grasping behaviors learned from MAT.

\subsection{Hyperparameters for Soft Proximal Policy Optimization}
\label{sec:hyperparams}
Table \ref{tbl:hyperparams} exhibits the hyperparameters for Soft Proximal Policy Optimization.
\begin{table}[H]
    \centering
    \caption{Hyperparameters: Soft Proximal Policy Optimization}
    \resizebox{0.6\linewidth}{!}{
    \begin{tabular}{llc}
    \toprule
    Category & Hyper-parameter & Value \\ \midrule
    \multirow{3}{*}{Baseline network} & Hidden layers & 3 \\
         & Hidden dimension & 128 \\ 
         & Base learning rate & $1\times10^{-4}$\\
         & Minibatch size & 200 \\\midrule
    \multirow{3}{*}{Subpolicy network} & Hidden layers & 3 \\
         & Hidden dimension & 128 \\ 
         & Base learning rate & $1\times10^{-4}$\\ & Minibatch size & 350 \\\midrule
    \multirow{13}{*}{Optimization} 
         & Num. actors & 10 \\
         & Num. episodes per batch / actor & 30 \\
         & Horizon & 250 \\
         & Num. epoches / batch & 10 \\
         & Discount ($\gamma$) & 0.999 \\
         & Temperature parameter ($\alpha$)& $5\times10^{-4}$ \\
         & GAE parameter ($\lambda$) & 0.95 \\
         & PPO clipping coeff. ($\epsilon$) & 0.2 \\
         & Gradient clipping & 200 \\
         & VF coeff. ($c_1$) & 1.0 \\
         & Optimizer & Adam \\
         \bottomrule
    \end{tabular}
    }
    \label{tbl:hyperparams}
\end{table}
\subsection{Simulation Training Grasping Scenes}
\begin{figure}[H]
    \centering
    \includegraphics[width=\linewidth]{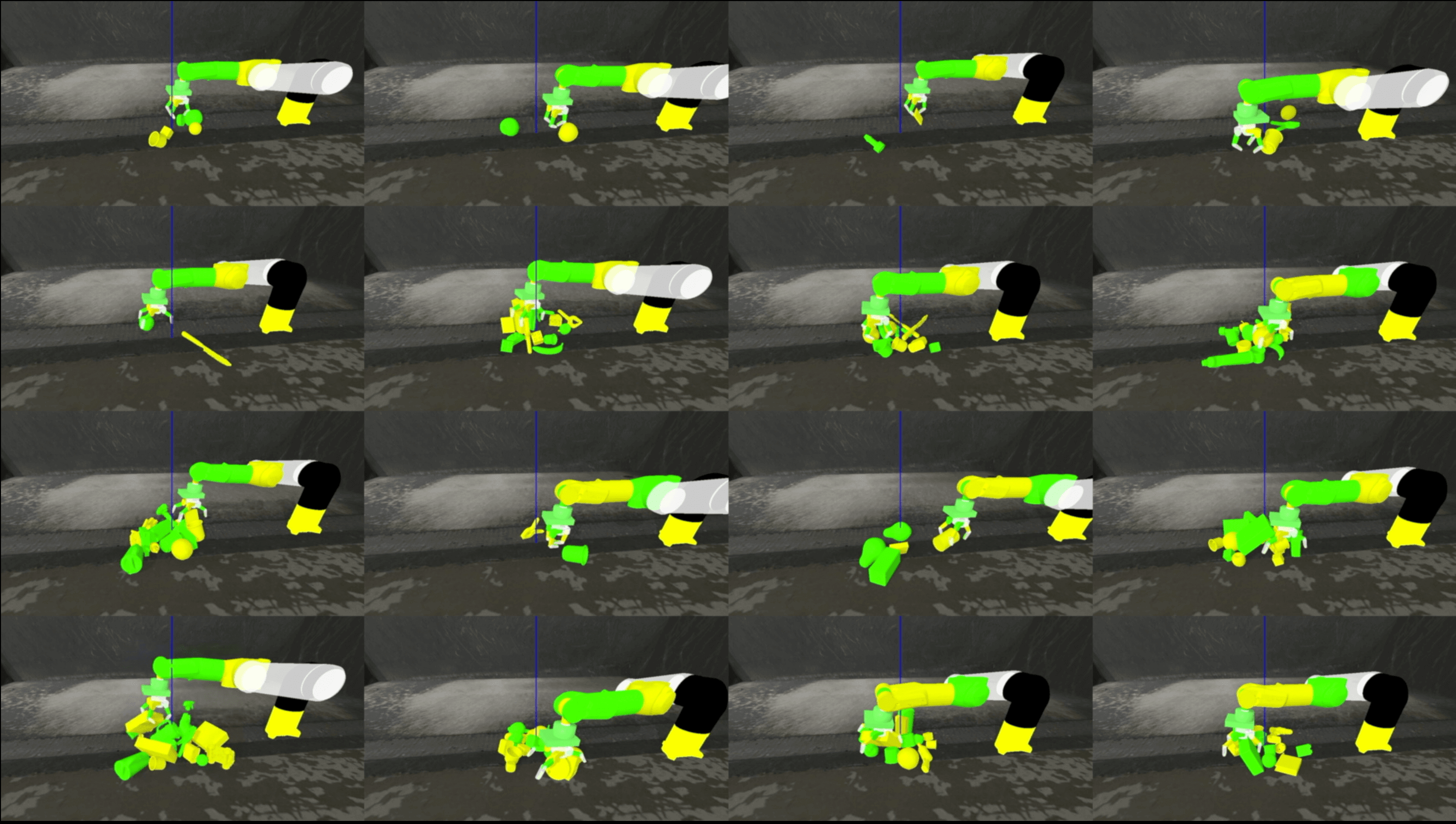}
    \caption{\textbf{Example Simulated Grasping Scenes in PyBullet for Training.} Here, 16 sample scenes are displayed, in which a random number of seen objects are dropped into the scene.}
    \label{fig:simulation-training}
\end{figure}

Figure~\ref{fig:simulation-training} exhibits some simulated cluttered grasping scenes randomly generated for training.

\subsection{Real-World Experimental Grasping Scenes}
\begin{figure}[H]
\vspace{3mm}
\centering
    \begin{subfigure}[ht]{\columnwidth}
    \includegraphics[width=\linewidth]{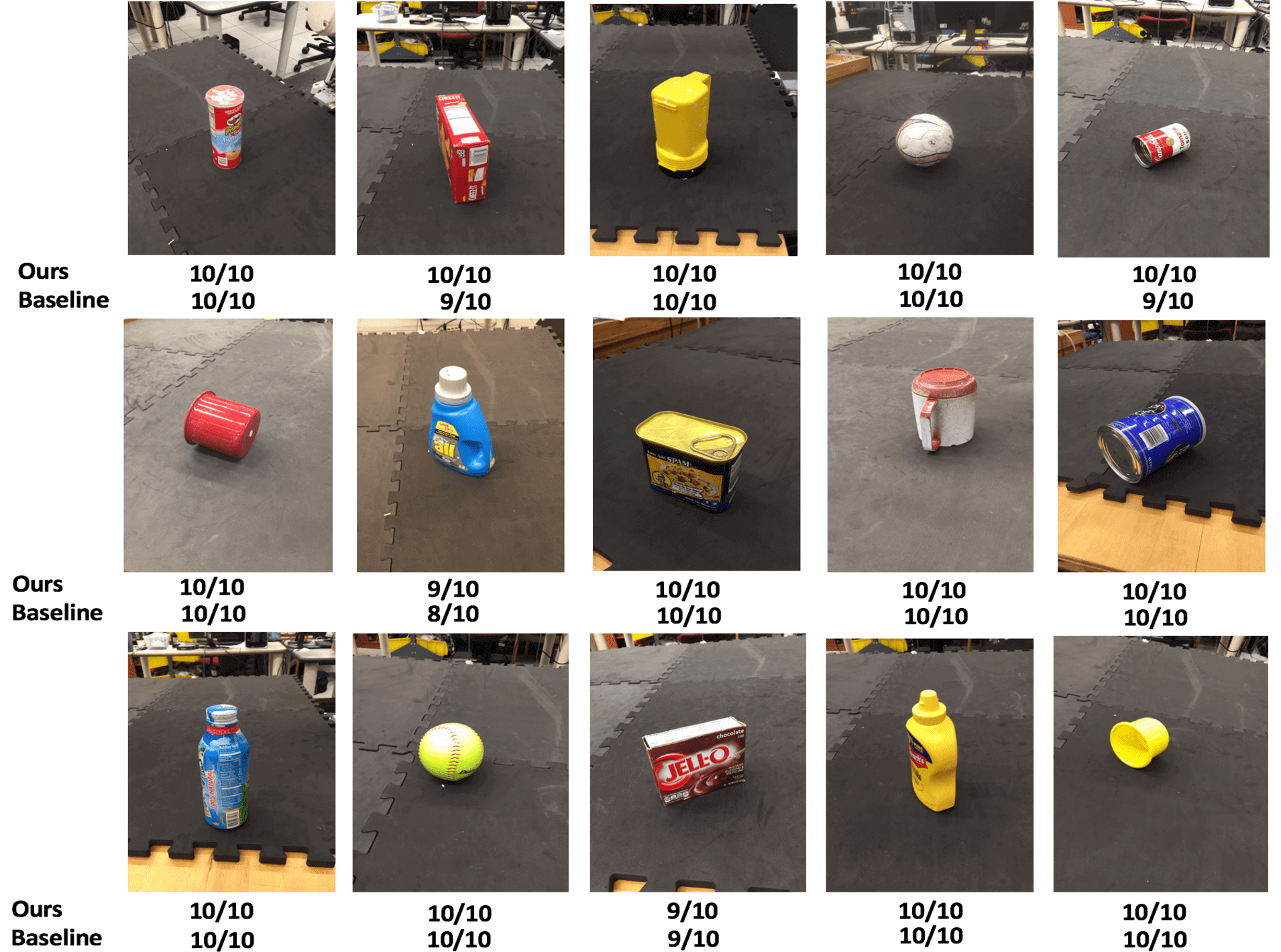}
    \caption{Seen Objects}
    \label{fig:pb}
    \end{subfigure}
    \begin{subfigure}[ht]{\columnwidth}
    \includegraphics[width=\linewidth]{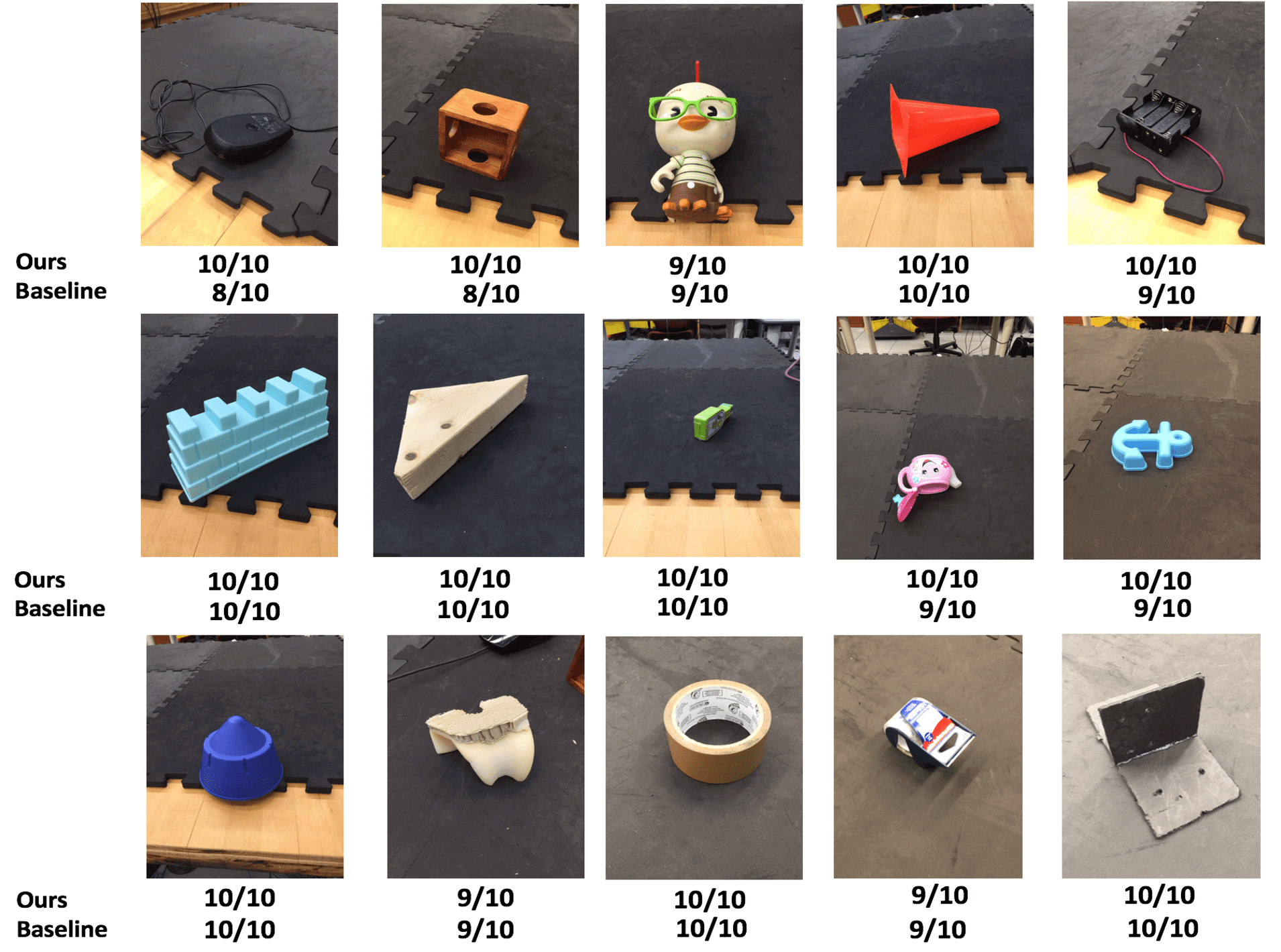}
    \caption{Novel Objects}
    \label{fig:pb}
    \end{subfigure}
    \caption{\small Single-Object Performance (MAT vs. Baseline \cite{wu2019pixeliros}) in \# Successful Grasps / \# Grasp Attempts. Calibration noise: 0cm.}
    \label{fig:singlescene}
\vspace{-5mm}
\end{figure}

\begin{figure}[H]
\vspace{3mm}
\centering
    \begin{subfigure}[ht]{\columnwidth}
    \includegraphics[width=\linewidth]{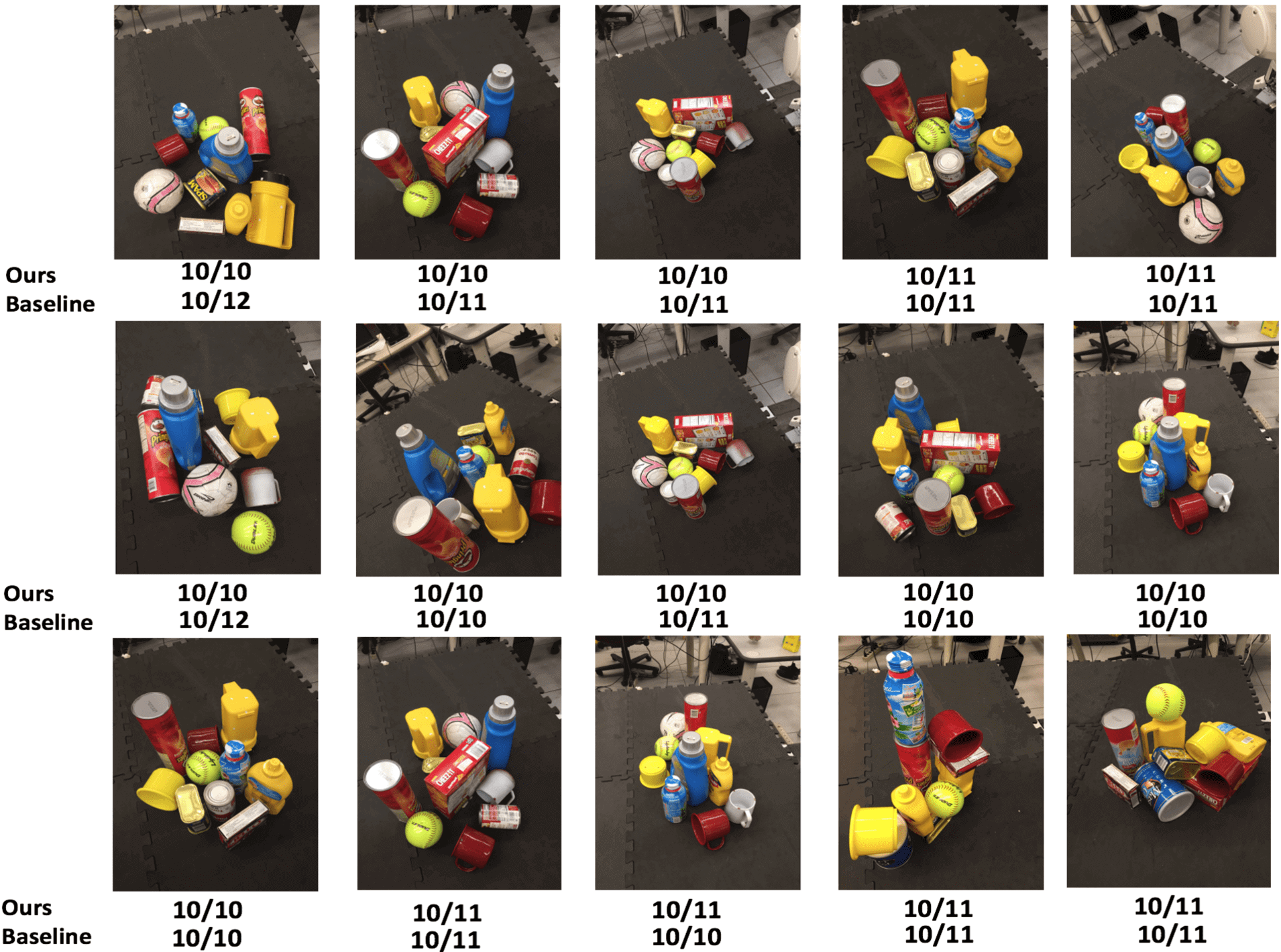}
    \caption{Seen Objects}
    \label{fig:pb}
    \end{subfigure}
    \begin{subfigure}[ht]{\columnwidth}
    \includegraphics[width=\linewidth]{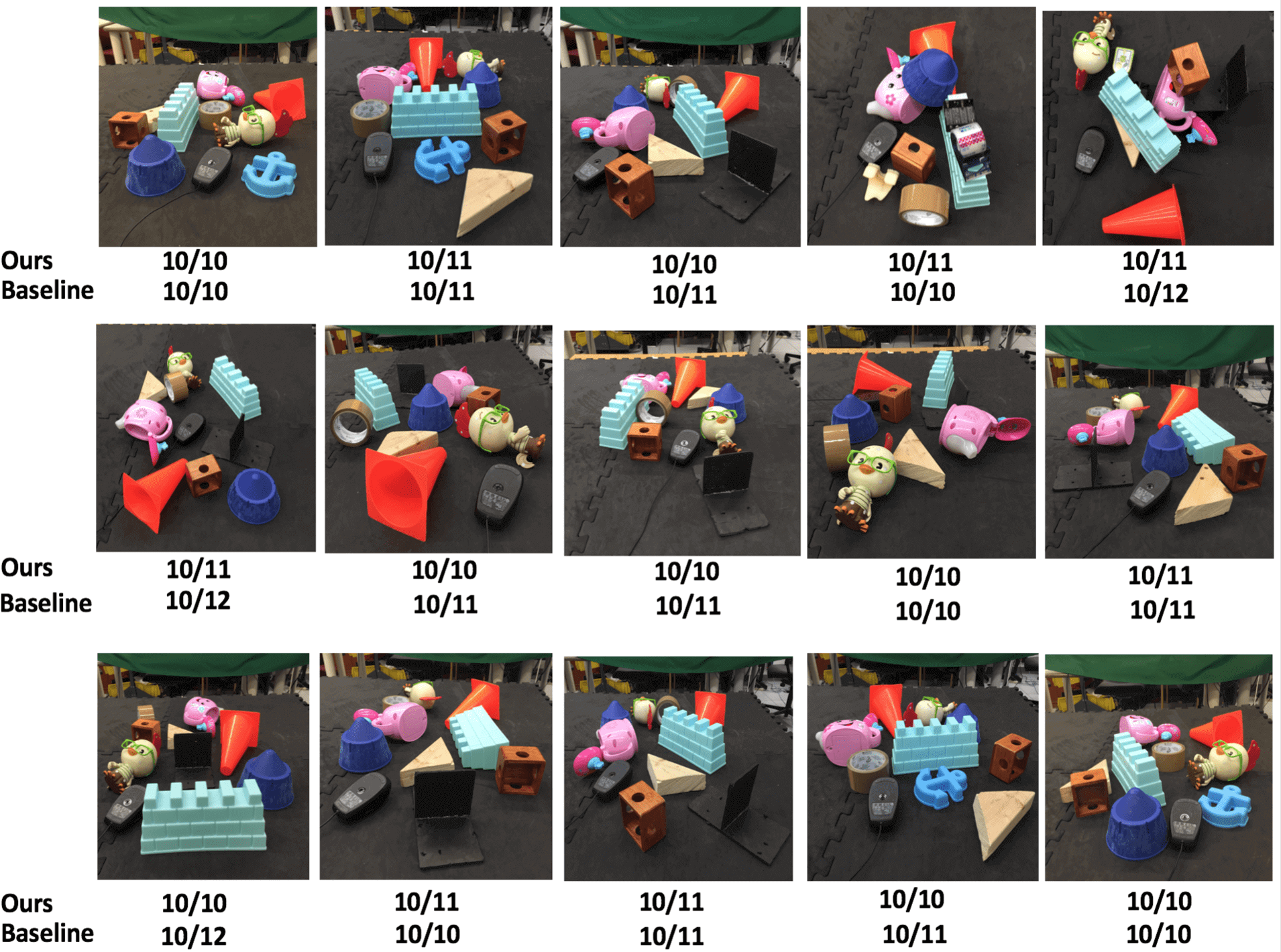}
    \caption{Novel Objects}
    \label{fig:pb}
    \end{subfigure}
    \caption{\small Cluttered-Scene Performance (MAT vs. Baseline \cite{wu2019pixeliros}) in \# Successful Grasps / \# Grasp Attempts. Calibration noise: 0cm.}
    \label{fig:multiscene}
\vspace{-5mm}
\end{figure}

In Figure~\ref{fig:singlescene} and Figure~\ref{fig:multiscene} we show some grasping scenes representative of the different real-world experiments.

\subsection{Binary Tactile Contacts Observation}
\label{Contacts}
\textbf{Stabilizing real-world raw tactile readings.} Tactile sensors on the physical Barrett BH-282 Hand stream tactile readings at a frequency of 246 Hz. These tactile readings provide the magnitude of force felt by each of the 96 tactile cells, which range from 0 to 20. To generate stable tactile readings for each cell, we calculate the running mean of the last 50 readings for each cell. Tactile readings in PyBullet simulation are stable so no averaging is required.

\textbf{Obtaining binary tactile contacts from raw tactile readings.}
To obtain binary tactile contacts ($s_t^{contacts\_binary} \in \{0, 1\}^{20 \times 96}$) from the raw running average of the tactile readings, we use a threshold of 0.8 for both the physical and the simulated hand. Values above this threshold indicate contact.

\textbf{Simulating soft tactile contacts in the real world.} 
PyBullet simulation uses soft-body contacts (allowing slight interpenetration during collision resolution) while real-world contacts are mostly hard. To reduce the sim-to-real gap, we synthesize soft tactile contacts in the real world by adding the proprioceptive effort felt by a tactile finger to the raw tactile readings of each cell on this finger.

\subsection{Position Adjustment after Reopening Action}
\label{Position}
Let $M$ be the number of links equipped with tactile cells for a multi-fingered hand. For the BH-282 Barrett Hand, $M=4$ since all three fingers and the palm have tactile cells. Let $C$ be the total number of tactile cells on each finger or the palm. For the BH-282 Barrett Hand, $C=24$. Let $T_{m, c}$ denote the binary tactile contact for the $c^{th}$ cell on the $m^{th}$ tactile link, where $m \in [1, M], c \in [1, C]$. Similarly, let $P_{m, c} = [x_{m, c}, y_{m, c}, z_{m, c}]$ be the Cartesian location of the tactile cell expressed in the world frame. Let $\Bar{M} = \{m \in [1, M]: \sum_{c=1}^C T_{m,c} > 0\}$ be the set of tactile links that have at least one active tactile cell. Let $\Bar{C}_m = \{c \in [1, C]: T_{m,c} = 1\}$ be the set of active tactile cells on an active tactile link $m$, where $m \in \Bar{M}$. Let $P_{old} = [x_{old}, y_{old}, z_{old}]$ be the Cartesian location of the end-effector palm after reopening but before position adjustment, expressed in the world frame. During position adjustment, the robot examines the most recent timestep during which at least one tactile cell of the hand was activated. The end-effector palm's new Cartesian location is then calculated as:
\begin{equation}
    P_{new} = [x_{new}, y_{new}, z_{new}] 
\end{equation}
where 
\begin{align}
    x_{new} &= \frac{1}{|\Bar{M}|}\sum_{m \in \Bar{M}}\frac{1}{|\Bar{C}_m|}\sum_{c \in \Bar{C}_m} x_{m, c} \\
    y_{new} &= \frac{1}{|\Bar{M}|}\sum_{m \in \Bar{M}}\frac{1}{|\Bar{C}_m|}\sum_{c \in \Bar{C}_m} y_{m, c}\\ z_{new} &= z_{old}
\end{align}
Intuitively, the hand's $[x, y]$ coordinates re-locate to the $[x, y]$ coordinates of the center of all active finger-palm tactile centers. Each active finger-palm tactile center is the center of all Cartesian locations of the finger or palm's active tactile cells. On the other hand, the $z$ coordinate of the hand stays unchanged. In the case where no tactile cell was activated throughout history: $P_{new} = P_{old}$.

\subsection{Position Adjustment for Initial Side Grasps}
\begin{figure}[H]
    \centering
    \begin{subfigure}[t]{0.245\textwidth}
        \captionsetup{skip=2pt}\includegraphics[trim={200 0 0 0},clip,width=\linewidth]{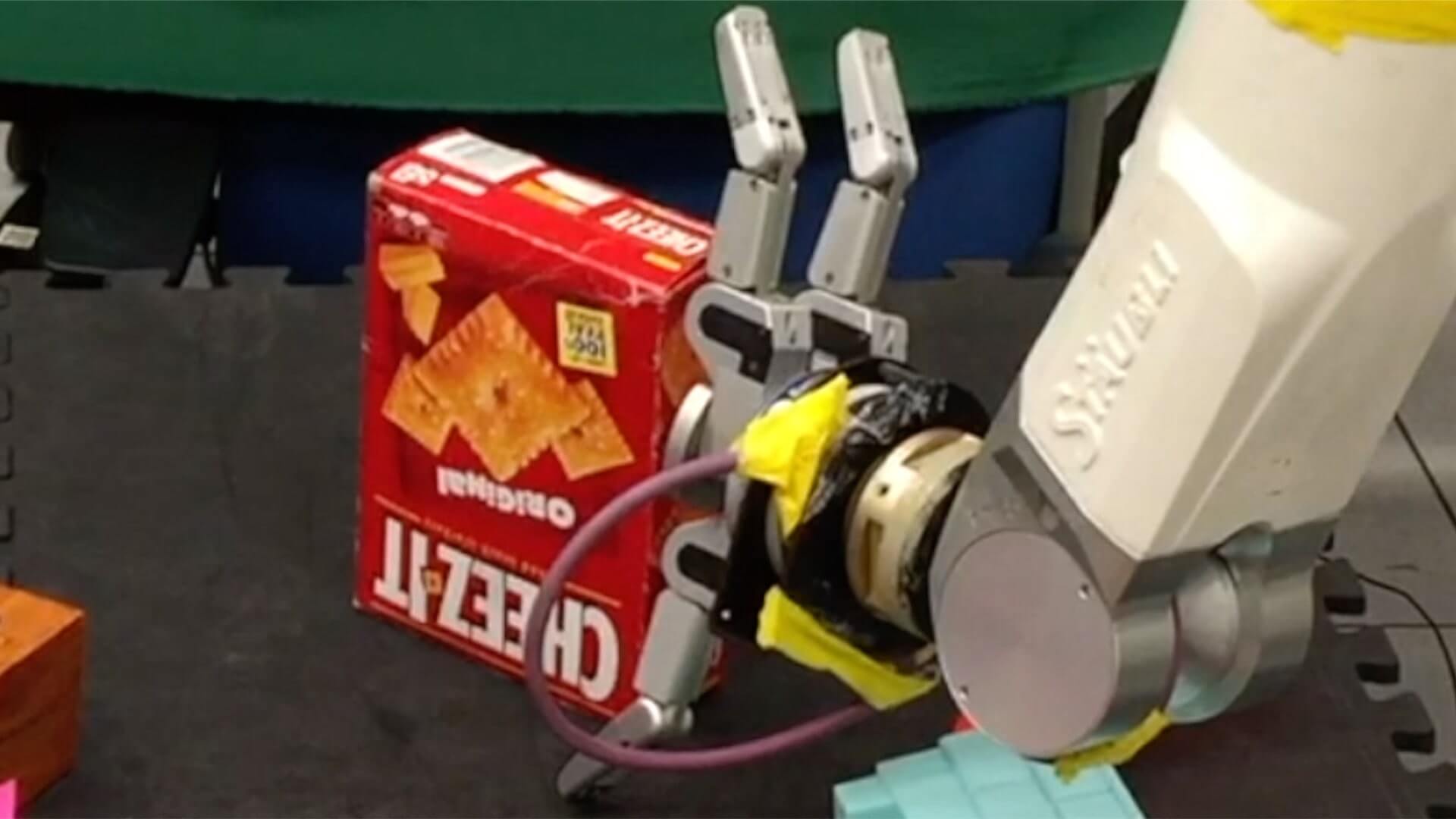}\caption{}
    \end{subfigure}
    \begin{subfigure}[t]{0.245\textwidth}
        \captionsetup{skip=2pt}\includegraphics[trim={200 0 0 0},clip,width=\linewidth]{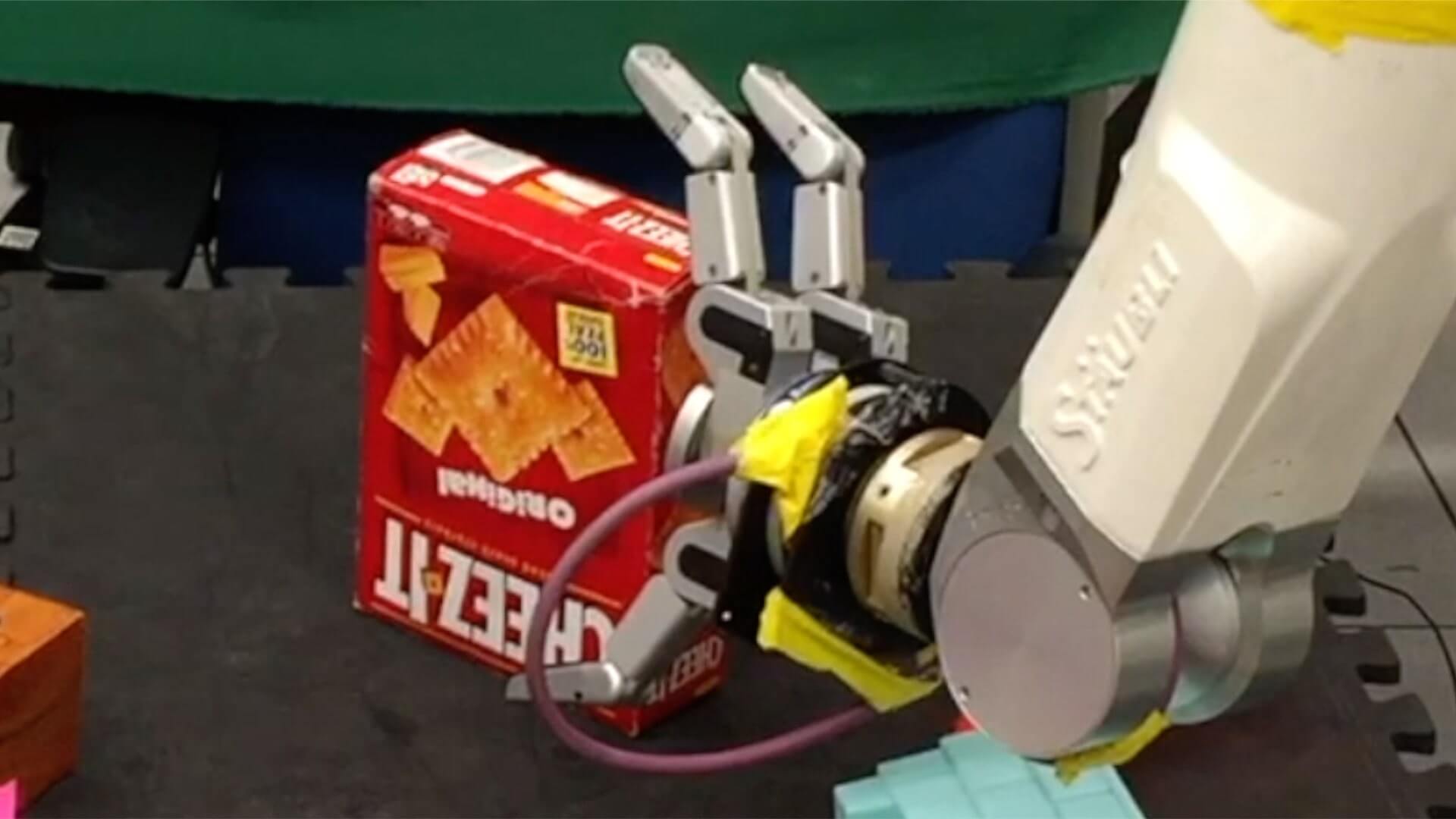}\caption{}
    \end{subfigure}
    \begin{subfigure}[t]{0.245\textwidth}
        \captionsetup{skip=2pt}\includegraphics[trim={200 0 0 0},clip,width=\linewidth]{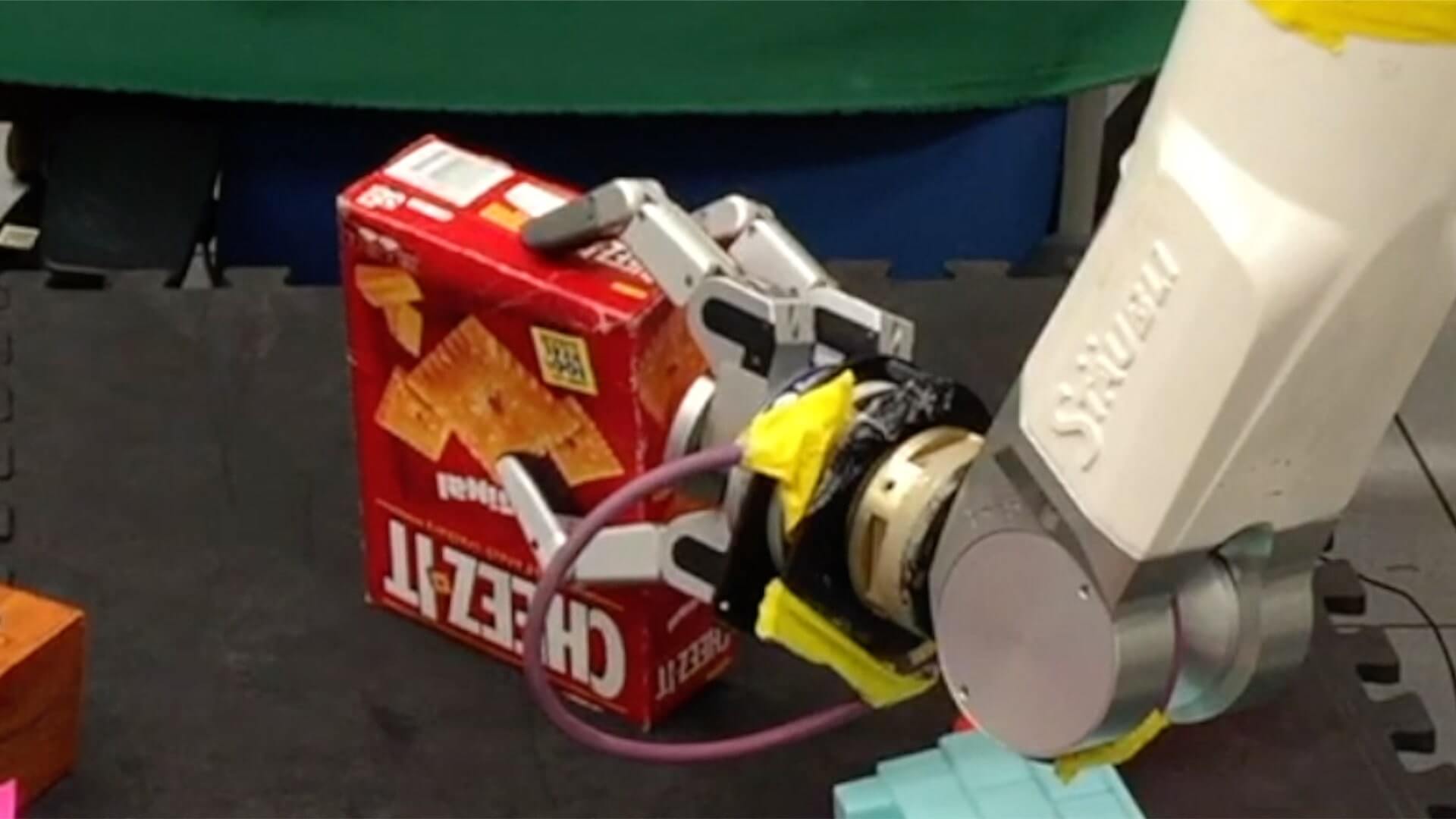}\caption{}
    \end{subfigure}
    \begin{subfigure}[t]{0.245\textwidth}
        \captionsetup{skip=2pt}\includegraphics[trim={200 0 0 0},clip,width=\linewidth]{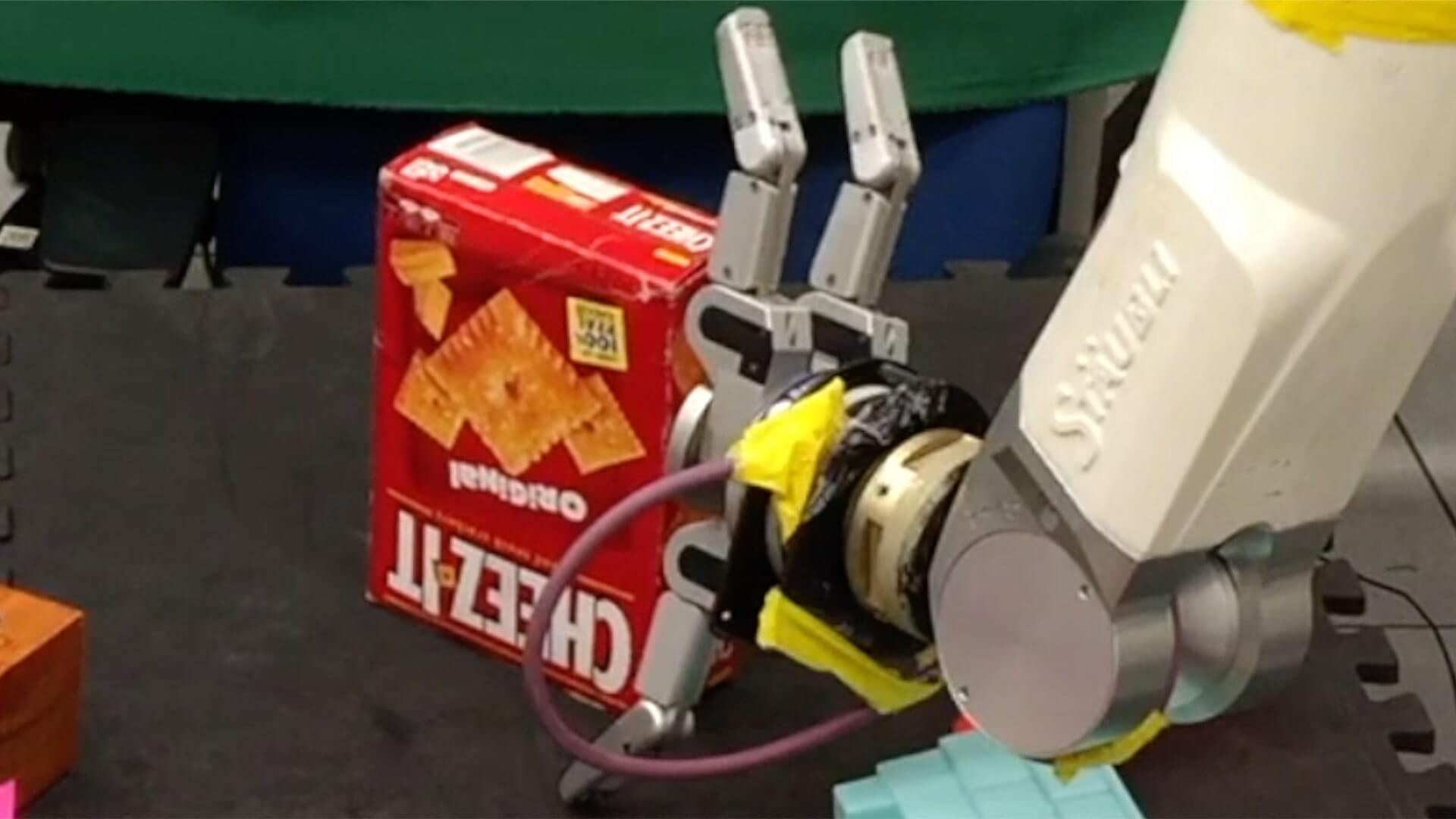}\caption{}
    \end{subfigure}
    \begin{subfigure}[t]{0.245\textwidth}
        \captionsetup{skip=2pt}\includegraphics[trim={200 0 0 0},clip,width=\linewidth]{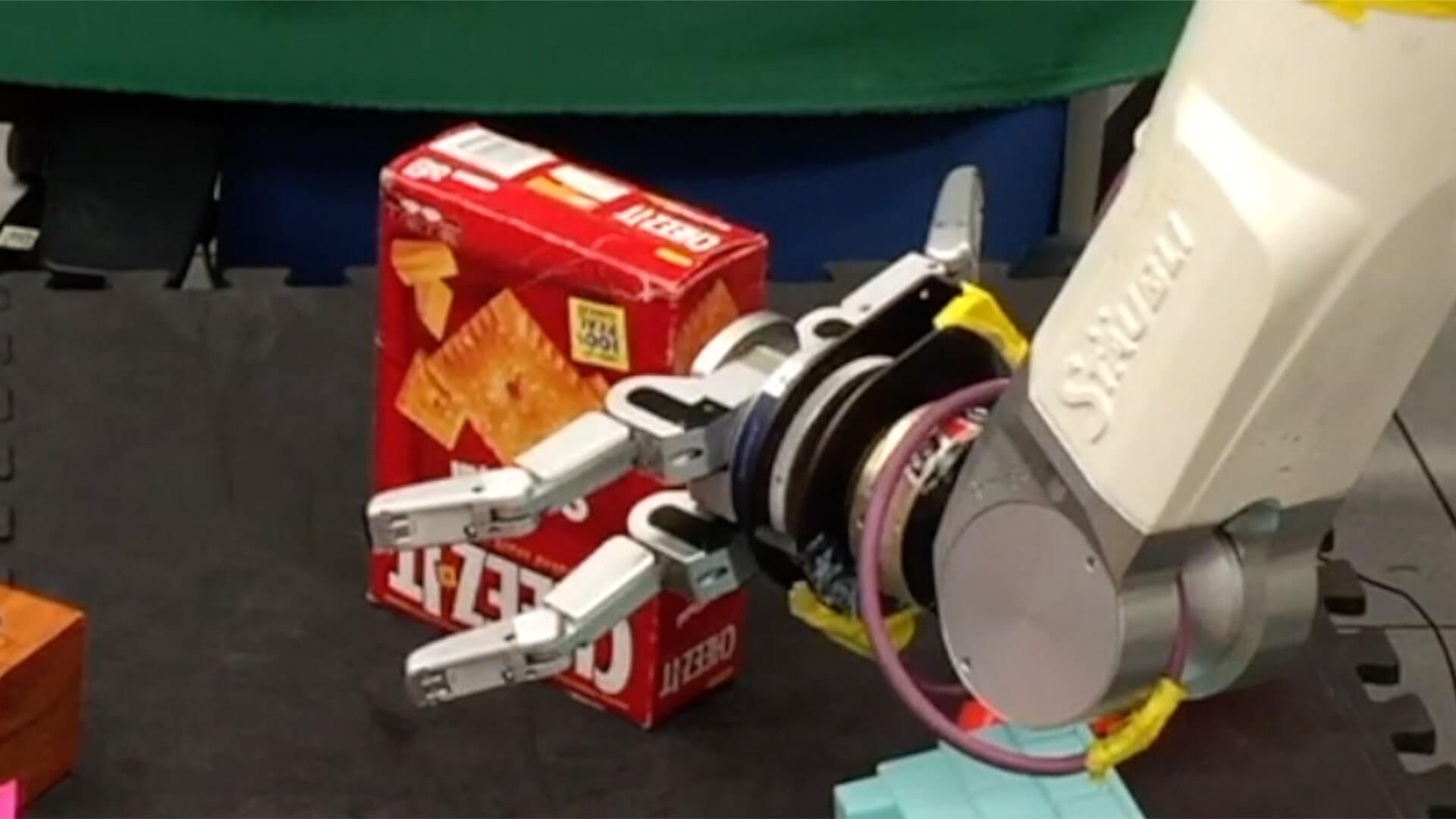}\caption{}
    \end{subfigure}
    \begin{subfigure}[t]{0.245\textwidth}
        \captionsetup{skip=2pt}\includegraphics[trim={200 0 0 0},clip,width=\linewidth]{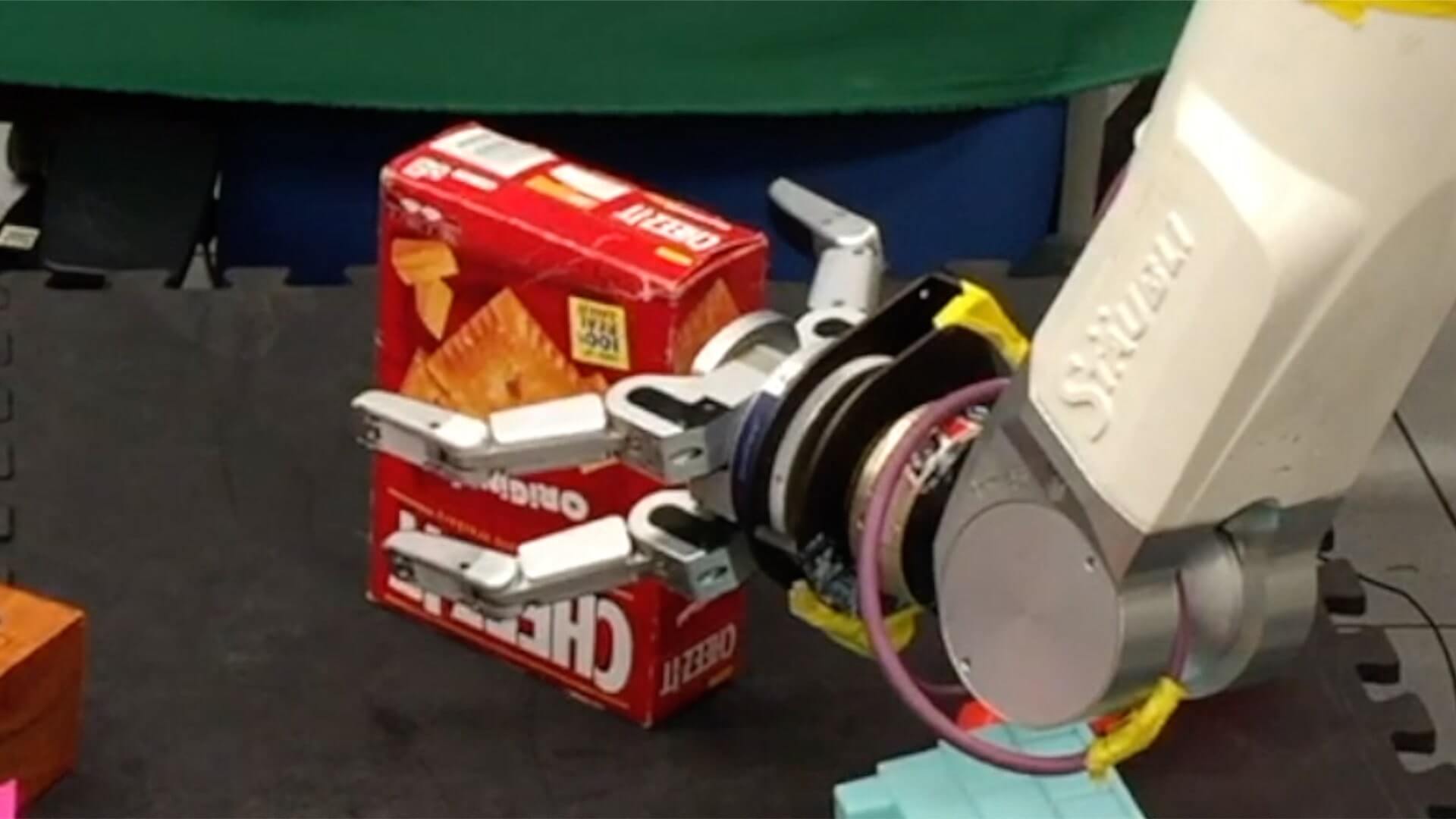}\caption{}
    \end{subfigure}
    \begin{subfigure}[t]{0.245\textwidth}
        \captionsetup{skip=2pt}\includegraphics[trim={200 0 0 0},clip,width=\linewidth]{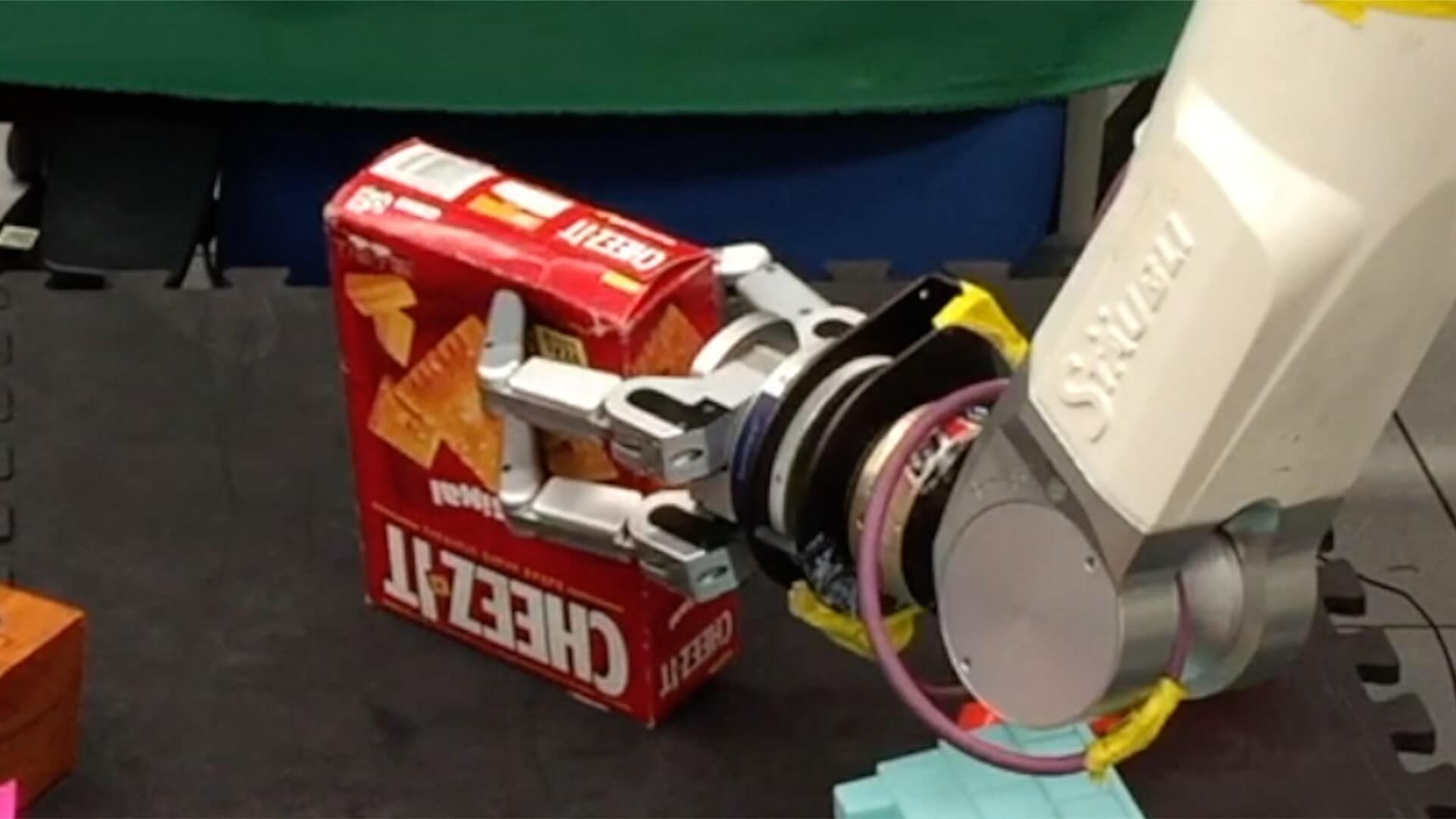}\caption{}
    \end{subfigure}
    \begin{subfigure}[t]{0.245\textwidth}
        \captionsetup{skip=2pt}\includegraphics[trim={200 0 0 0},clip,width=\linewidth]{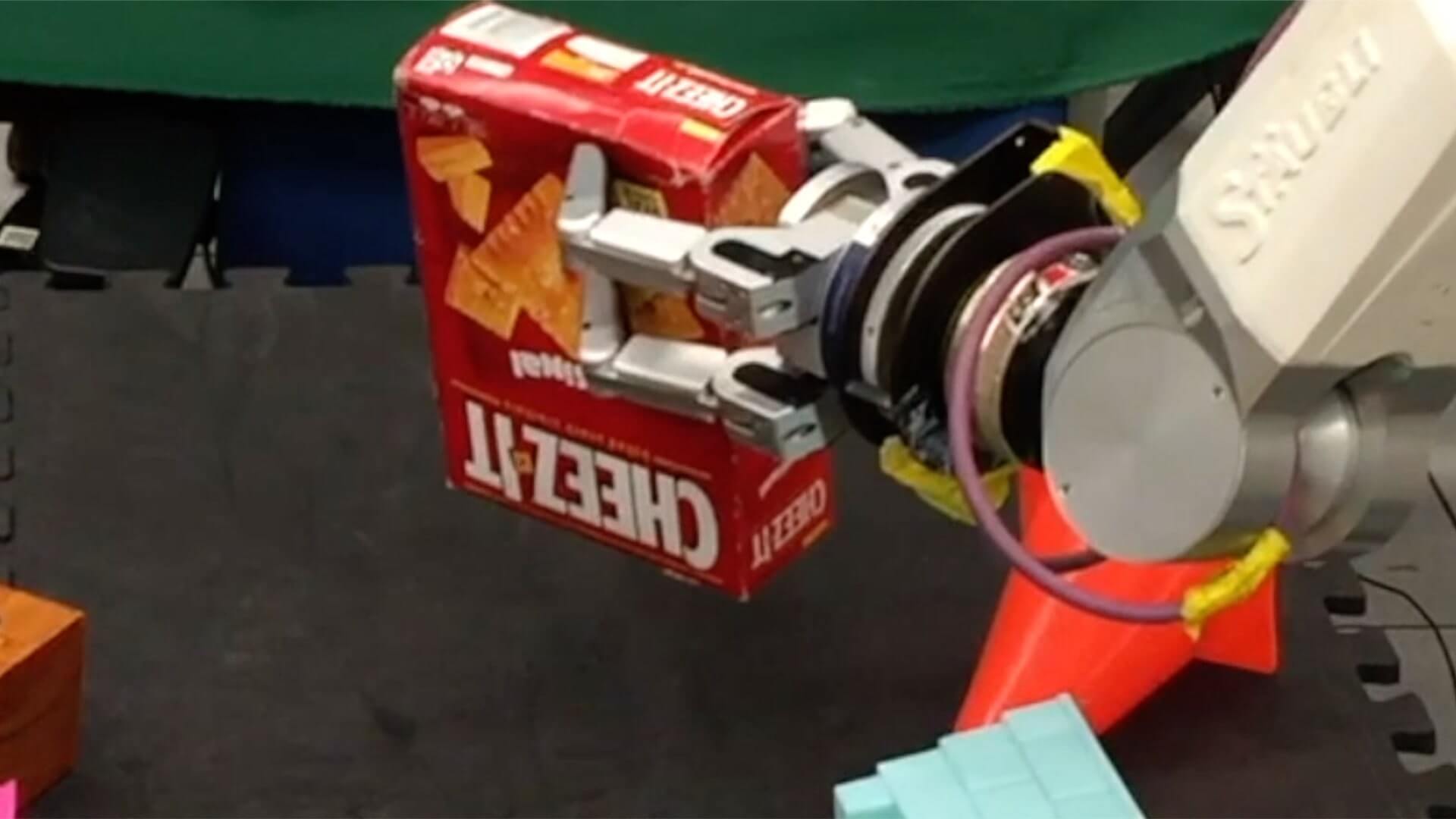}\caption{}
    \end{subfigure}
    \caption{Learned Closed-Loop Behaviors of MAT under an Initial Side Grasp}
    \label{sidegraspbehavior}
\end{figure}

For side grasps where the end-effector palm has a horizontal directional vector, we disable position adjustment for safety guarantee while still enabling orientation adjustment, because in this case the horizontal movement of the end-effector will leave the hand vulnerable to breaking. An example MAT maneuver from an initial side grasp generated by~\cite{wu2019pixeliros} is shown in Figure~\ref{sidegraspbehavior}.
\end{document}